\documentclass{article}

\usepackage{tabularx}
\usepackage{graphicx,subfigure}
\usepackage{fullpage}
\usepackage{hyperref}
\usepackage{natbib}
\usepackage{delarray}
\usepackage{times}
\usepackage{color}
\usepackage{amssymb}
\usepackage{amsmath}
\usepackage{amsthm}
\usepackage{mathrsfs}
\usepackage{tikz}
\usepackage{xspace}
\usepackage{enumerate}
\usepackage{algorithm}
\usepackage{algorithmic}
\usepackage{multirow}
\usepackage{nameref}
\usepackage{booktabs}
\usepackage{mdframed}
\usepackage{fmtcount}
\usepackage{footnote}
\usepackage{thmtools,thm-restate}
\usepackage{makecell}

\newcommand{\proj}[1]{\mbox{Proj}_{#1}}
\newcommand{\eat}[1]{}
\newcommand{\snap}{x_{s(t)}}

\newcommand{\R}{{\mathbb R}}

\newcommand{\E}{{\mathbb{E}}}

\newcommand{\inner}[2]{\langle #1, #2 \rangle}

\newcommand{\ns}[1]{\| #1 \|^2}
\newcommand{\n}[1]{\| #1 \|}

\newcommand{\tx}{\widetilde{x}}

\newcommand{\tdo}{\widetilde{O}}
\newcommand{\tdtheta}{\widetilde{\Theta}}
\newcommand{\tdc}{\widetilde{C}}
\newcommand{\tdomega}{\widetilde{\Omega}}
\newcommand{\bx}{\bar{x}}

\newcommand{\mathL}{\mathscr{L}}
\newcommand{\mathG}{\mathscr{G}}

\newcommand{\hessian}{\mathcal{H}}

        {\hspace*{\fill}$\Box$\par\vspace{4mm}}
\newenvironment{proofof}[1]{\smallskip\noindent{\bf Proof of #1.}}%
        {\hspace*{\fill}$\Box$\par}

\newtheorem{theorem}{Theorem}
\newtheorem{lemma}{Lemma}

\newtheorem{definition}{Definition}
\newtheorem{assumption}{Assumption}

\hypersetup{
    colorlinks=true,       
    linkcolor=black,    
    citecolor=black,        
    filecolor=magenta,     
    urlcolor=blue
}

\title{Stabilized SVRG: Simple Variance Reduction for Nonconvex Optimization}
\author{Rong Ge\thanks{Duke University. Email: rongge@cs.duke.edu}\\ \and  Zhize Li\thanks{Tsinghua University. Email: zz-li14@mails.tsinghua.edu.cn}\\ \and Weiyao Wang\thanks{Duke University. Email: weiyao.wang1997@gmail.com}\\ \and  Xiang Wang\thanks{Duke University. Email: xwang@cs.duke.edu}}

\usepackage{times}

\begin{document}

\maketitle

\begin{abstract}%
Variance reduction techniques like SVRG \citep{johnson2013accelerating} provide simple and fast algorithms for optimizing a convex finite-sum objective. For nonconvex objectives, these techniques can also find a first-order stationary point (with small gradient). However, in nonconvex optimization it is often crucial to find a second-order stationary point (with small gradient and almost PSD hessian). In this paper, we show that Stabilized SVRG \--- a simple variant of SVRG \--- can find an $\epsilon$-second-order stationary point using only $\tdo(n^{2/3}/\epsilon^2+n/\epsilon^{1.5})$ stochastic gradients. To our best knowledge, this is the first second-order guarantee for a simple variant of SVRG. The running time almost matches the known guarantees for finding $\epsilon$-first-order stationary points.
\end{abstract}


\section{Introduction}



Nonconvex optimization is widely used in machine learning. Recently, 
for problems like matrix sensing~\citep{bhojanapalli2016global}, matrix completion~\citep{ge2016matrix}, and certain objectives for neural networks~\citep{ge2017learning}, it was shown that all local minima are also globally optimal, therefore simple local search algorithms can be used to solve these problems.

For a convex function $f(x)$, a local and global minimum is achieved whenever the point has zero gradient: $\nabla f(x) = 0$. However, for nonconvex functions, a point with zero gradient can also be a saddle point. To avoid converging to saddle points, recent results~\citep{ge2015escaping,jin2017escape,jin2017accelerated} prove stronger results that show local search algorithms converge to $\epsilon$-approximate second-order stationary points \--- points with small gradients and almost positive semi-definite Hessians (see Definition~\ref{def:stationary}). 

In theory, \cite{xu2018first} and \cite{allen2017neon2} independently showed that finding a second-order stationary point is not much harder than finding a first-order stationary point \--- they give reduction algorithms Neon/Neon2 that can converge to second-order stationary points when combined with algorithms that find first-order stationary points. Algorithms obtained by such reductions are complicated, and they require a negative curvature search subroutine: given a point $x$, find an approximate smallest eigenvector of $\nabla^2 f(x)$. In practice, standard algorithms for convex optimization work in a nonconvex setting without a negative curvature search subroutine. 

What algorithms can be directly adapted to the nonconvex setting, and what are the simplest modifications that allow a theoretical analysis? For gradient descent, \cite{jin2017escape} showed that a simple perturbation step is enough to find a second-order stationary point, and this was later shown to be necessary \citep{du2017gradient}. For accelerated gradient, \cite{jin2017accelerated} showed a simple modification would allow the algorithm to work in the nonconvex setting, and escape from saddle points faster than gradient descent. In this paper, we show that there is also a simple modification to the Stochastic Variance Reduced Gradient (SVRG) algorithm \citep{johnson2013accelerating} that is guaranteed to find a second-order stationary point.

SVRG is designed to optimize a finite sum objective $f(x)$ of the following form: 
$$
f(x) := \frac{1}{n} \sum_{i=1}^n f_i(x),
$$
where evaluating $f$ would require evaluating every $f_i$. 
 In the original result, \cite{johnson2013accelerating} showed that when $f_i(x)$'s are $L$-smooth and $f(x)$ is $\mu$ strongly convex, SVRG finds a point with error $\epsilon$ in time $O(n\log(1/\epsilon))$ when $L/\mu=O(n)$. The same guarantees were also achieved by algorithms like SAG~\citep{roux2012stochastic}, SDCA~\citep{shalev2013stochastic} and SAGA~\citep{defazio2014saga}, but SVRG is much cleaner both in terms of implementation and analysis.

SVRG was analyzed in nonconvex regimes, \cite{reddi2016stochastic} and \cite{allen2016variance} showed that SVRG can find an $\epsilon$-first-order stationary point using $O(\frac{n^{2/3}}{\epsilon^2} +n)$ stochastic gradients. \cite{li2018simple} analyzed a batched-gradient version of SVRG and achieved the same guarantee with much simpler analysis. These results can then be combined with the reduction~\citep{allen2017neon2,xu2018first} to give complicated algorithms for finding second-order stationary points. Using more complicated optimization techniques, it is possible to design faster algorithms for finding first-order stationary points, including FastCubic~\citep{agarwal2016finding}, SNVRG~\citep{zhou2018stochastic}, SPIDER-SFO~\citep{fang2018spider}. These algorithms can also combine with procedures like Neon2 to give second-order guarantees.

In this paper, we give a variant of SVRG called Stabilized SVRG that is able to find $\epsilon$-second-order stationary points, while maintaining the simplicity of the SVRG algorithm. See Table~\ref{table:comparison} for a comparison between our algorithm and existing results. The main term $\tdo(n^{2/3}/\epsilon^2)$ in the running time of our algorithm matches the analysis with first-order guarantees. All other algorithms that achieve second-order guarantees require negative curvature search subroutines like Neon2, and many are more complicated than SVRG even without this subroutine.

\begin{table}
\centering
\begin{tabular}{|c|c|c|c|}
\hline
Algorithm & Stochastic Gradients & Guarantee & Simple \\
\hline
\makecell{SVRG~\citep{reddi2016stochastic}\\
\citep{allen2016variance}} & $O(\frac{n^{2/3}}{\epsilon^2}+n)$ & 1st-Order & \checkmark \\
\hline
Minibatch-SVRG~\citep{li2018simple} & $O(\frac{n^{2/3}}{\epsilon^2}+n)$ & 1st-Order & \checkmark \\
\hline
Neon2+SVRG~\citep{allen2017neon2} & $\tdo(\frac{n^{2/3}}{\epsilon^2} +\frac{n}{\epsilon^{1.5}}+\frac{n^{3/4}}{\epsilon^{1.75}})$ & 2nd-Order & $\times$ \\
\hline
\makecell{Neon2+FastCubic/CDHS\\
\citep{agarwal2016finding, carmon2016accelerated}} & $\tdo(\frac{n}{\epsilon^{1.5}}+\frac{n^{3/4}}{\epsilon^{1.75}} )$ & 2nd-Order & $\times$ \\
\hline
SNVRG$^+$+Neon2~\citep{zhou2018finding,zhou2018stochastic} &$\tdo(\frac{n^{1/2}}{\epsilon^2} +\frac{n}{\epsilon^{1.5}}+\frac{n^{3/4}}{\epsilon^{1.75}})$ & 2nd-Order & $\times$\\
\hline
SPIDER-SFO$^+$~\citep{fang2018spider} &$\tdo(\frac{n^{1/2}}{\epsilon^2} +\frac{1}{\epsilon^{2.5}})$ & 2nd-Order & $\times$\\
\hline
Stabilized SVRG (this paper) & $\tdo(\frac{n^{2/3}}{\epsilon^2} +\frac{n}{\epsilon^{1.5}})$ &2nd-Order & \checkmark\\ \hline
\end{tabular}
\caption{Optimization algorithms for non-convex finite-sum objective}\label{table:comparison}
\end{table}
%
%
%

\section{Preliminaries}\label{preliminaries}
\subsection{Notations}
We use $\mathbb{N},\ \R$ to denote the set of natural numbers and real numbers respectively. We use $[n]$ to denote the set $\{1,2,\cdots,n\}$. Let $I_b$ be a multi-set of size $b$ whose $i$-th element ($i=1,2,...,b$) is chosen i.i.d. from $[n]$ uniformly ($I_b$ is used to denote the samples used in a mini-batch for the algorithm). 
For vectors we use $\inner{u}{v}$ to denote their inner product, and for matrices we use $\inner{A}{B}:=\sum_{i,j}A_{ij}B_{ij}$ to denote the trace of $AB^\top.$ We use $\n{\cdot}$ to denote the Euclidean norm for a vector and spectral norm for a matrix, and $\lambda_{\max}(\cdot), \lambda_{\min}(\cdot)$ to denote the largest and the smallest eigenvalue of a real symmetric matrix. 

Throughout the paper, we use $\tdo(f(n))$ and $\tdomega(f(n))$ to hide poly log factors on relevant parameters. We did not try to optimize the poly log factors in the proof. 

\subsection{Finite-Sum Objective and Stationary Points}
Now we define the objective that we try to optimize. A finite-sum objective has the form

\begin{equation}
    \min_{x\in \R^d}\Big\{f(x):= \frac{1}{n}\sum_{i=1}^{n}f_i(x)\Big\}, \label{problem1}
\end{equation}
where $f_i$ maps a $d$-dimensional vector to a scalar and $n$ is finite. In our model, both $f_i(x)$ and $f(x)$ can be non-convex. We make standard smoothness assumptions as follows: 

\begin{assumption} \label{lipschitz}
Each individual function $f_i(x)$ has $L$-Lipschitz Gradient, that is,
$$\forall x_1, x_2 \in \R^d,\ \n{\nabla f_i(x_1)-\nabla f_i(x_2)}\leq L\n{x_1-x_2}.$$
\end{assumption}

This implies that the average function $f(x)$ also has $L$-Lipschitz gradient. We assume the average function $f(x)$ and individual functions have Lipschitz Hessian. That is,  
\begin{assumption} \label{Hessian}
The average function $f(x)$ has $\rho$-Lipschitz Hessian, which means
$$\forall x_1, x_2 \in \R^d,\ \n{\nabla^2 f(x_1)-\nabla^2 f(x_2)}\leq \rho\n{x_1-x_2};$$
each individual function $f_i(x)$  has $\rho'$-Lipschitz Hessian, which means
$$\forall x_1, x_2 \in \R^d,\ \n{\nabla^2 f_i(x_1)-\nabla^2 f_i(x_2)}\leq \rho'\n{x_1-x_2}.$$
\end{assumption}

These two assumptions are standard in the literature for finding second-order stationary points\\
\citep{ge2015escaping,jin2017escape,jin2017accelerated, allen2017neon2}. 
The goal of non-convex optimization algorithms is to converge to an approximate-second-order stationary point. 


\begin{definition}\label{def:stationary}
For a differentiable function $f$, $x$ is a first-order stationary point if $\n{\nabla f(x)}=0$; $x$ is an $\epsilon$-first-order stationary point if $\n{\nabla f(x)}\leq \epsilon$.

For twice-differentiable function $f$, $x$ is a second-order stationary point if 
$$\n{\nabla f(x)}=0\ \mbox{and }\ \lambda_{\min}(\nabla^2 f(x))\geq 0.$$
If $f$ is $\rho$-Hessian Lipschitz, $x$ is an $\epsilon$-second-order stationary point if 
$$\n{\nabla f(x)}\leq \epsilon,\ \mbox{and }\ \lambda_{\min}(\nabla^2 f(x))\geq -\sqrt{\rho\epsilon}.$$
\end{definition}

%
%
%

This definition of $\epsilon$-second-order stationary point is standard in previous literature~\citep{ge2015escaping,jin2017escape,jin2017accelerated}. 
Note that the definition of second-order stationary point uses the Hessian Lipschitzness  parameter $\rho$ of the average function $f(x)$ (instead of $\rho'$ of individual function). It is easy to check that $\rho \le \rho'$. In Appendix~\ref{sec:matrix} we show there are natural applications where $\rho' = \Theta(d) \rho$, so in general algorithms that do not depend heavily on $\rho'/\rho$ are preferred.

\subsection{SVRG Algorithm}\label{SVRG algorithm}


In this section we give a brief overview of the SVRG algorithm. In particular we follow the minibatch version in \cite{li2018simple} which is used for our analysis for simplicity.
 
SVRG algorithm has an outer loop. We call each iteration of the outer loop an {\bf epoch}. At the beginning of each epoch, define the snapshot vector $\tx$ to be the current iterate and compute its full gradient $\nabla f(\tx)$. Each epoch of SVRG consists of $m$ iterations. In each iteration, the SVRG algorithm picks $b$ random samples (with replacement) from $[n]$ and form a multi-set $I_b$, and then estimate the gradient as:
$$v_t:=\frac{1}{b}\sum_{i\in I_b}(\nabla f_i(x_t)-\nabla f_i(\tx)+\nabla f(\tx))$$
After estimating the gradient, the SVRG algorithm performs an update $x_{t+1}\leftarrow x_t-\eta v_t$, where $\eta$ is the step size. The choice of gradient estimate gives an unbiased estimate of the true gradient, and often has much smaller variance compared to stochastic gradient descent. The pseudo-code for minibatch-SVRG is given in Algorithm~\ref{alg:svrg}.

\begin{algorithm}[!htb]
	\caption{SVRG($x_0, m, b, \eta, S$)}
	\label{alg:svrg}
	\begin{algorithmic}[1]
		\REQUIRE 
		initial point $x_0$, epoch length $m$, minibatch size $b$, step size $\eta$, number of epochs $S$.
		\ENSURE
		point $x_{Sm}$.
		\FOR {$s=0,1,\cdots, S-1$}
			\STATE Compute $\nabla f(x_{sm}).$
	        \FOR{$t=1, 2, \ldots, m$}
	        	\STATE Sample $b$ i.i.d. numbers uniformly from $[n]$ and form a multi-set $I_b$.
    	    	\STATE $v_{sm+t-1}\leftarrow \frac{1}{b}\sum_{i\in I_b}\big(\nabla f_i(x_{sm+t-1})-\nabla f_i(x_{sm}) + \nabla f(x_{sm})) \big)$.
        	   	\STATE $x_{sm+t} \leftarrow x_{sm+t-1} - \eta v_{sm+t-1}$.
        	\ENDFOR
        \ENDFOR
        \RETURN $x_{Sm}$.
    \end{algorithmic}
\end{algorithm}

%

\section{Our Algorithms: Perturbed SVRG and Stabilized SVRG}\label{sec:mainresults}
In this paper we give two simple modifications to the original SVRG algorithm. First, similar to perturbed gradient descent~\citep{jin2017escape}, we add perturbations to SVRG algorithm to make it escape from saddle points efficiently. We will show that this algorithm finds an $\epsilon$-second-order stationary point in $\tdo((\frac{n^{2/3}L\Delta f}{\epsilon^2}+\frac{n\sqrt{\rho}\Delta f}{\epsilon^{1.5}})(1+(\frac{\rho'}{n^{1/3}\rho})^2))$ time, where $\Delta f:=f(x_0)-f^*$ is the difference between initial function value and the optimal function value. This algorithm is efficient as long as $\rho' \le \rho n^{1/3}$, but can be slower if $\rho'$ is much larger (see Appendix~\ref{sec:matrix} for an example where $\rho' = \Theta(d)\rho$\footnote{Existing algorithms like Neon2+SVRG try to estimate the Hessian at a single point, so they do not depend heavily on $\rho'$ (in particular, they do not depend on $\rho'$ given access to a Hessian-vector product oracle, and only depends logarithmically on $\rho'$ with a gradient oracle). However for our algorithm the iterates keep moving so it is more difficult to get the correct dependency on $\rho'$.}. 
To achieve stronger guarantees, we introduce {\em Stabilized SVRG}, which is another simple modification on top of Perturbed SVRG that improves the dependency on $\rho'$.

%

\subsection{Perturbed SVRG}

\begin{algorithm}[!htb]
	\caption{Perturbed SVRG($x_0, m, b, \eta,\delta ,\mathG$)}
	\label{alg:psvrg-simple}
	\begin{algorithmic}[1]
		\REQUIRE 
		initial point $x_0$, epoch length $m$, minibatch size $b$, step size $\eta$, perturbation radius $\delta$, threshold gradient $\mathG$
		\FOR {$s=0,1,2,\cdots$}
			\STATE Compute $\nabla f(x_{sm})$.
			\IF{not currently in a super epoch and $\n{\nabla f(x_{sm})}\leq \mathG$}
				\STATE $x_{sm}\leftarrow x_{sm}+\xi,$ where $\xi$ uniformly $\sim \mathbb{B}_0(\delta)$, start a super epoch
			\ENDIF
			\FOR {$t=1, 2, \cdots, m$}
\STATE Sample $b$ i.i.d. numbers uniformly from $[n]$ and form a multi-set $I_b$.

    	    	\STATE $v_{sm+t-1}\leftarrow \frac{1}{b}\sum_{i\in I_b}\big(\nabla f_i(x_{sm+t-1})-\nabla f_i(x_{sm}) + \nabla f(x_{sm})) \big)$.
        	   	\STATE $x_{sm+t} \leftarrow x_{sm+t-1} - \eta v_{sm+t-1}$.
        	   	\STATE \algorithmicif~Stopping condition is met
        	   		\algorithmicthen~Stop super epoch
			\ENDFOR
		\ENDFOR
    \end{algorithmic}
\end{algorithm}

Similar to gradient descent, if one starts SVRG exactly at a saddle point, it is easy to check that the algorithm will not move. To avoid this problem, we propose Perturbed SVRG. A high level description is in Algorithm~\ref{alg:psvrg-simple}. Intuitively, since at the beginning of each epoch in SVRG the gradient of the function is computed, we can add a small perturbation to the current point if the gradient turns out to be small (which means we are either near a saddle point or already at a second-order stationary point). Similar to perturbed gradient descent in \cite{jin2017escape}, we also make sure that the algorithm does not add a perturbation very often - the next perturbation can only happen either after many iterations $(T_{\max})$ or if the point travels enough distance $(\mathL)$. The full algorithm is a bit more technical and is given in Algorithm~\ref{alg:psvrg} in appendix. 

Later, we will call the steps between the beginning of perturbation and end of perturbation a {\bf super epoch}. When the algorithm is not in a super epoch, for technical reasons we also use a version of SVRG that stops at a random iteration (not reflected in Algorithm~\ref{alg:psvrg-simple} but is in Algorithm~\ref{alg:psvrg}). 

%
%

For perturbed SVRG, we have the following guarantee:
\begin{theorem}
\label{thm:maintheorem_psvrg}%
Assume the function $f(x)$ is $\rho$-Hessian Lipschitz, and each individual function $f_i(x)$ is $L$-smooth and $\rho'$-Hessian-Lipschitz. Let $\Delta f:= f(x_0)-f^*$, where $x_0$ is the initial point and $f^*$ is the optimal value of $f$. There exist mini-batch size $b=\tdo(n^{2/3})$, epoch length  $m=n/b$, step size $\eta=\tdo(1/L)$, perturbation radius $\delta=\tdo(\min(\frac{\rho^{1.5}\sqrt{\epsilon}}{\max(\rho^2,(\rho'/m)^2)}, \frac{\rho^{0.75}\epsilon^{0.75}}{\max(\rho,\rho'/m)\sqrt{L}}))$, super epoch length $T_{\max}=\tdo(\frac{L}{\sqrt{\rho\epsilon}})$, threshold gradient $\mathG=\tdo(\epsilon)$, threshold distance $\mathL=\tdo(\frac{\sqrt{\epsilon\rho}}{\max(\rho,\rho'/m)}),$ such that Perturbed SVRG (Algorithm~\ref{alg:psvrg}) will at least once get to an $\epsilon$-second-order stationary point with high probability using 
$$\tdo\Big(\big(\frac{n^{2/3}L\Delta f}{\epsilon^2}+\frac{n\sqrt{\rho}\Delta f}{\epsilon^{1.5}}\big)\big(1+(\frac{\rho'}{n^{1/3}\rho})^2\big)\Big)$$
stochastic gradients. 
\end{theorem}
    
\subsection{Stabilized SVRG}    
    
\begin{algorithm}[!htb]
	\caption{Stabilized SVRG($x_0, m, b, \eta,\delta,\mathG$)}
	\label{alg:ssvrg-simple}
	\begin{algorithmic}[1]
		\REQUIRE 
		initial point $x_0$, epoch length $m$, minibatch size $b$, step size $\eta$, perturbation radius $\delta$, threshold gradient $\mathG$
		\FOR {$s=0,1,2,\cdots$}
			\STATE Compute $\nabla f(x_{sm})$.
			\IF{not currently in a super epoch and $\n{\nabla f(x_{sm})}\leq \mathG$}
				\STATE $v_{shift} \leftarrow \nabla f(x_{sm}).$
				\STATE $x_{sm}\leftarrow x_{sm}+\xi,$ where $\xi$ uniformly $\sim \mathbb{B}_0(\delta)$, start a super epoch
			\ENDIF
			\FOR {$t=1, 2, \cdots, m$}
\STATE Sample $b$ i.i.d. numbers uniformly from $[n]$ and form a multi-set $I_b$.
    	    	\STATE $v_{sm+t-1}\leftarrow \frac{1}{b}\sum_{i\in I_b}\big(\nabla f_i(x_{sm+t-1})-\nabla f_i(x_{sm}) + \nabla f(x_{sm})) \big) -v_{shift}$.
        	   	\STATE $x_{sm+t} \leftarrow x_{sm+t-1} - \eta v_{sm+t-1}$.
        	   	\STATE \algorithmicif~Stopping condition is met
        	   		\algorithmicthen~Stop super epoch and $v_{shift}\leftarrow 0.$
			\ENDFOR
		\ENDFOR
    \end{algorithmic}
\end{algorithm}

In order to relax the dependency on $\rho'$, we further introduce stabilization in the algorithm. Basically, if we encounter a saddle point $\tx$, we will run SVRG iterations on a shifted function $\hat{f}(x):=f(x)-\inner{\nabla f(\tx)}{x-\tx}$, whose gradient at $\tx$ is {\em exactly zero}. Another minor (but important) modification is to perturb the point in a ball with much smaller radius compared to Algorithm~\ref{alg:psvrg-simple}. We will give more intuitions to show why these modifications are necessary in Section~\ref{sec:negativecurvature}.

The high level ideas of Stabilized SVRG is given in Algorithm~\ref{alg:ssvrg-simple}. In the pseudo-code, the key observation is that gradient on the shifted function is equal to the gradient of original function plus a stabilizing term. 
Detailed implementation of Stabilized SVRG is deferred to Algorithm~\ref{alg:ssvrg}. For Stabilized SVRG, the time complexity in the following theorem only has a poly-logarithmic dependency on $\rho'$, which is hidden in $\tdo(\cdot)$ notation. 
\begin{theorem}
\label{thm:maintheorem}%
Assume the function $f(x)$ is $\rho$-Hessian Lipschitz, and each individual function $f_i(x)$ is $L$-smooth and $\rho'$-Hessian Lipschitz. Let $\Delta f:= f(x_0)-f^*$, where $x_0$ is the initial point and $f^*$ is the optimal value of $f$. There exists mini-batch size $b=\tdo(n^{2/3})$, epoch length  $m=n/b$, step size $\eta=\tdo(1/L)$, perturbation radius $\delta=\tdo(\min(\frac{\sqrt{\epsilon}}{\sqrt{\rho}}, \frac{m\sqrt{\rho\epsilon}}{\rho'}))$, super epoch length $T_{\max}=\tdo(\frac{L}{\sqrt{\rho\epsilon}})$, threshold gradient $\mathG=\tdo(\epsilon)$, threshold distance $\mathL=\tdo(\frac{\sqrt{\epsilon}}{\sqrt{\rho}}),$ such that Stabilized SVRG (Algorithm~\ref{alg:ssvrg}) will at least once get to an $\epsilon$-second-order stationary point with high probability using 
$$\tdo(\frac{n^{2/3}L\Delta f}{\epsilon^2}+\frac{n\sqrt{\rho}\Delta f}{\epsilon^{1.5}})$$
stochastic gradients. 
\end{theorem}

In previous work~\citep{allen2017neon2}, it has been shown that Neon2+SVRG has similar time complexity for finding second-order stationary point, $\tdo(\frac{n^{2/3} L\Delta f}{\epsilon^2}+\frac{n\rho^2\Delta f}{\epsilon^{1.5}}+\frac{n^{3/4}\rho^2\sqrt{L}\Delta f}{\epsilon^{1.75}})$. Our result achieves a slightly better convergence rate using a much simpler variant of SVRG.
\section{Overview of Proof Techniques}\label{sec:proofskech}
In this section, we illustrate the main ideas in the proof of Theorems~\ref{thm:maintheorem_psvrg} and \ref{thm:maintheorem}. 
Similar to many existing proofs for escaping saddle points, we will show that Algorithms~\ref{alg:psvrg-simple} and \ref{alg:ssvrg-simple} can decrease the function value efficiently either when the current point $x_t$ has a large gradient ($\|\nabla f(x_t)\| \ge \epsilon$) or has a large negative curvature ($\lambda_{\min}(\nabla^2 f(x_t)) \le -\sqrt{\rho\epsilon}$). Since the function value cannot decrease below the global optimal $f^*$, the algorithms will be able to find a second-order stationary point within the desired number of iterations.

In the proof, we use similar notations as in previous paper~\citep{jin2017escape}. We use $\mathG$ to denote the threshold of the gradient norm, and show that the function value decreases if the average norm of the gradients is at least $\mathG.$ Starting from a saddle point, the super-epoch ends if the number of steps exceeds the threshold  $T_{\max}$ or the distance to the saddle point exceeds the threshold distance $\mathL$. 
In both algorithms, we choose $\mathG=\tdo(\epsilon), T_{\max}=\tdo(\frac{L}{\sqrt{\rho\epsilon}})$. For the distance threshold, we choose $\mathL=\tdo(\frac{\sqrt{\epsilon\rho}}{\max(\rho,\rho'/m)})$ for Perturbed SVRG and $\mathL=\tdo(\frac{\sqrt{\epsilon}}{\sqrt{\rho}})$ for Stabilized SVRG.


Throughout the analysis, we use $s(t)$ to denote the index of the snapshot point of iterate $x_t$. More precisely, $s(t) = m\lfloor t/m \rfloor$.

\subsection{Exploiting Large Gradients}\label{sec:largegradient}

There have already been several proofs that show SVRG can converge to a first-order stationary point, and our proof here is very similar. First, we show that the gradient estimate is accurate as long as the current point is close to the snapshot point.
\begin{lemma}
\label{lem:varaincedistance}
For any point $x_t$, let the gradient estimate be $v_t:=\frac{1}{b}\sum_{i\in I_b}(\nabla f_i(x_t)-\nabla f_i(x_{s(t)})+\nabla f(x_{s(t)}))$, where $x_{s(t)}$ is the snapshot point of the current epoch. Then, with probability at least $1-\zeta$, we have
$$\n{v_t-\nabla f(x_t)}\leq O\Big(\frac{\log(d/\zeta)L}{\sqrt{b}}\Big)\n{x_t-x_{s(t)}}.$$ 
\end{lemma}

This lemma is standard and the version for expected square error was proved in~\cite{li2018simple}. Here we only applied simple concentration inequalities to get a high probability bound.

Next, we show that the function value decrease is lower bounded by the summation of gradient norm squares. The proof of the following lemma is adopted from~\cite{li2018simple} with minor modifications.
\begin{lemma}
\label{lem:gradientfunctionvalue}
For any epoch, suppose the initial point is $x_0$, which is also the snapshot point for this epoch. Assume for any $0\leq t\leq m-1$, $\n{v_t-\nabla f(x_t)}\leq \frac{C_1 L}{\sqrt{b}}\n{x_t-x_0},$ where $C_1=\tdo(1)$ comes from Lemma~\ref{lem:varaincedistance}. Then, given $\eta\leq \frac{1}{3C_1 L}, b\geq m^2$, we have 
$$f(x_0)-f(x_t)\geq \sum_{\tau=0}^{t-1}\frac{\eta}{2}\ns{\nabla f(x_\tau)}$$
for any $1\leq t\leq m$.
\end{lemma}

Using this fact, we can now state the guarantee for exploiting large gradients.

\begin{lemma}
\label{lem:large_gradients}
For any epoch, suppose the initial point is $x_0$. Let $x_t$ be a point uniformly sampled from $\{x_\tau \}_{\tau=1}^m$. Then, given $\eta=\tdtheta(1/L), b\geq m^2$, for any value of $\mathG$ we have two cases:
\begin{enumerate}
\item if at least half of points in $\{x_\tau \}_{\tau=1}^m$ have gradient no larger than $\mathG,$ we know $\|\nabla f(x_t) \| \le \mathG$ holds with probability at least $1/2$;
\item otherwise, we know $f(x_0) - f(x_t) \ge \frac{\eta}{2}\frac{m\mathG^2}{4}$ holds with probability at least $1/5.$ 
\end{enumerate}
Further, no matter which case happens we always have $f(x_t) \le f(x_0)$ with high probability. 
\end{lemma}

As this lemma suggests, our algorithm will stop at a random iterate when it is not in a super epoch (this is reflected in the detailed Algorithms~\ref{alg:psvrg} and \ref{alg:ssvrg}). In the first case, since there are at least half points with small gradients, by uniform sampling, we know the sampled point must have small gradient with at least half probability. In the second case, the function value decreases significantly. 
Proofs for lemmas in this section are deferred to Appendix~\ref{sec:proofgradients}.

\subsection{Exploiting Negative Curvature - Perturbed SVRG}\label{sec:negativecurvatureperturb}
Section~\ref{sec:largegradient} already showed that if the algorithm is not in a super epoch, with constant probability every epoch of SVRG will either decrease the function value significantly, or end at a point with small gradient. In the latter case, if the point with small gradient also has almost positive semi-definite Hessian, then we have found an approximate-second-order stationary point. Otherwise, the algorithm will enter a super epoch, and we will show that with a reasonable probability Algorithm~\ref{alg:psvrg-simple} can decrease the function value significantly within the super epoch.


For simplicity, we will reset the indices for the iterates in the super epoch. Let the initial point be $\tx$, the point after the perturbation be $x_0$, and the iterates in this super epoch be $x_1,...,x_t$.

The proof for Perturbed SVRG is very similar to the proof of perturbed gradient descent in \cite{jin2017escape}. In particular, we perform a {\em two point analysis}. That is, we consider two coupled samples of the perturbed point $x_0,x_0'$. Let $e_1$ be the smallest eigendirection of Hessian $\hessian := \nabla^2 f(\tx)$. The two perturbed points $x_0$ and $x_0'$ only differ in the $e_1$ direction.  We couple the two trajectories from $x_0$ and $x_0'$ by choosing the same mini-batches for both of them. The iterates of the two sequences are denoted by $x_0,...,x_t$ and $x'_0,...,x'_t$ respectively.  Our goal is to show that with good probability one of these two points can escape the saddle point.

To do that, we will keep track of the difference between the two sequences $w_t = x_t - x'_t$. The key lemma in this section uses Hessian Lipschitz condition to show that the variance of $w_t$ (introduced by the random choice of mini-batch) can actually be much smaller than the variance we observe in Lemma~\ref{lem:varaincedistance}. More precisely,

\begin{lemma}\label{lem:coupled_gradient_estimator}
Let $\{x_t\}$ and $\{x_t'\}$ be two SVRG sequences running on $f$ that use the same choice of mini-batches. Let $x_{s(t)}$ be the snapshot point for iterate $t$. Let $w_t:= x_t-x_t'$ and $P_t=\max(\n{x_{s(t)}-\tx}, \n{x_{s(t)}'-\tx}, \n{x_{t}-\tx}, \n{x_{t}'-\tx})$. Then, with probability at least $1-\zeta$, we have
$$\n{\xi_t-\xi_t'}\leq O\Big(\frac{\log(d/\zeta)}{\sqrt{b}}\Big)\min\Big( L\n{w_t-w_{s(t)}} +\rho'P_t(\n{w_t}+\n{w_{s(t)}}), L(\n{w_t}+\n{w_{s(t)}}) \Big).$$  
\end{lemma}

This variance is often much smaller than before as in the extreme case, if $\rho' = 0$ (individual functions are quadratics), the variance is proportional to $\tdo(L/\sqrt{b})\|w_t - w_{s(t)}\|$. In the proof we will show that $w_t$ cannot change very quickly within a single epoch so $\|w_t - w_{s(t)}\|$ is much smaller than $\|w_t\|$ or $\|w_{s(t)}\|$. Using this new variance bound we can prove:

\begin{lemma}[informal]\label{lem:distanceproperty_psvrg_informal}
Let $\{x_t\}$ and $\{x_t'\}$ be two SVRG sequences running on $f$ that use the same choice of mini-batches. Assume $w_0=x_0-x_0'$ aligns with $e_1$ direction and $|\inner{e_1}{w_0}|\geq \frac{\delta}{4\sqrt{d}}.$ Setting the parameters appropriately we know with high probability $\max(\n{x_T-\tx},\n{x_T'-\tx})\geq \mathL$,
for some $T\leq \tdo(1/(\eta\gamma)).$
\end{lemma}

Intuitively, this lemma is true because at every iterate we expect $w_t$ to be multiplied by a factor of $(1+\eta\gamma)$ if the iterate follows exact gradient, and the variance bound from Lemma~\ref{lem:coupled_gradient_estimator} is tight enough. The precise statement of the lemma is given in Lemma~\ref{lem:distanceproperty_psvrg} in Appendix~\ref{sec:psvrg}. The lemma shows that one of the points can escape from a local neighborhood, which by the following lemma is enough to guarantee function value decrease:



\begin{lemma}\label{lem:dis_value_cross_epoch}
Let $x_0$ be the initial point, which is also the snapshot point of the current epoch. Let $\{x_t\}$ be the iterates of SVRG running on $f$ starting from $x_0$. Fix any $t\geq 1$, suppose for every $0\leq \tau \leq t-1, \n{\xi_\tau}\leq \frac{C_1 L}{\sqrt{b}}\n{x_\tau-x_{s(\tau)}},$ where $C_1$ comes from Lemma~\ref{lem:varaincedistance}. Given $\eta\leq \frac{1}{3C_1 L},b\geq m^2,$ we have 
$$\ns{x_t-x_0}\leq \frac{4t}{C_1 L}(f(x_0)-f(x_t)). $$
\end{lemma}

This lemma can be proved using the same technique as Lemma~\ref{lem:gradientfunctionvalue}. All proofs in this section are deferred to Appendix~\ref{sec:psvrg}. 

\subsection{Exploiting Negative Curvature - Stabilized SVRG}\label{sec:negativecurvature}
\label{subsec:stabilization}
The main problem in the previous analysis is that when $\rho'$ is large, the variance estimate in Lemma~\ref{lem:coupled_gradient_estimator} is no longer very strong. To solve this problem, note that the additional term $\rho'P_t(\n{w_t}+\n{w_{s(t)}})$ is proportional to $P_t$ (the maximum distance of the iterates to the initial point). If we can make sure that the iterates stay very close to the initial point for long enough we will still be able to use Lemma~\ref{lem:coupled_gradient_estimator} to get a good variance estimate. 

\begin{figure}
\centering
\begin{subfigure}
  \centering
  \includegraphics[width=0.45\textwidth]{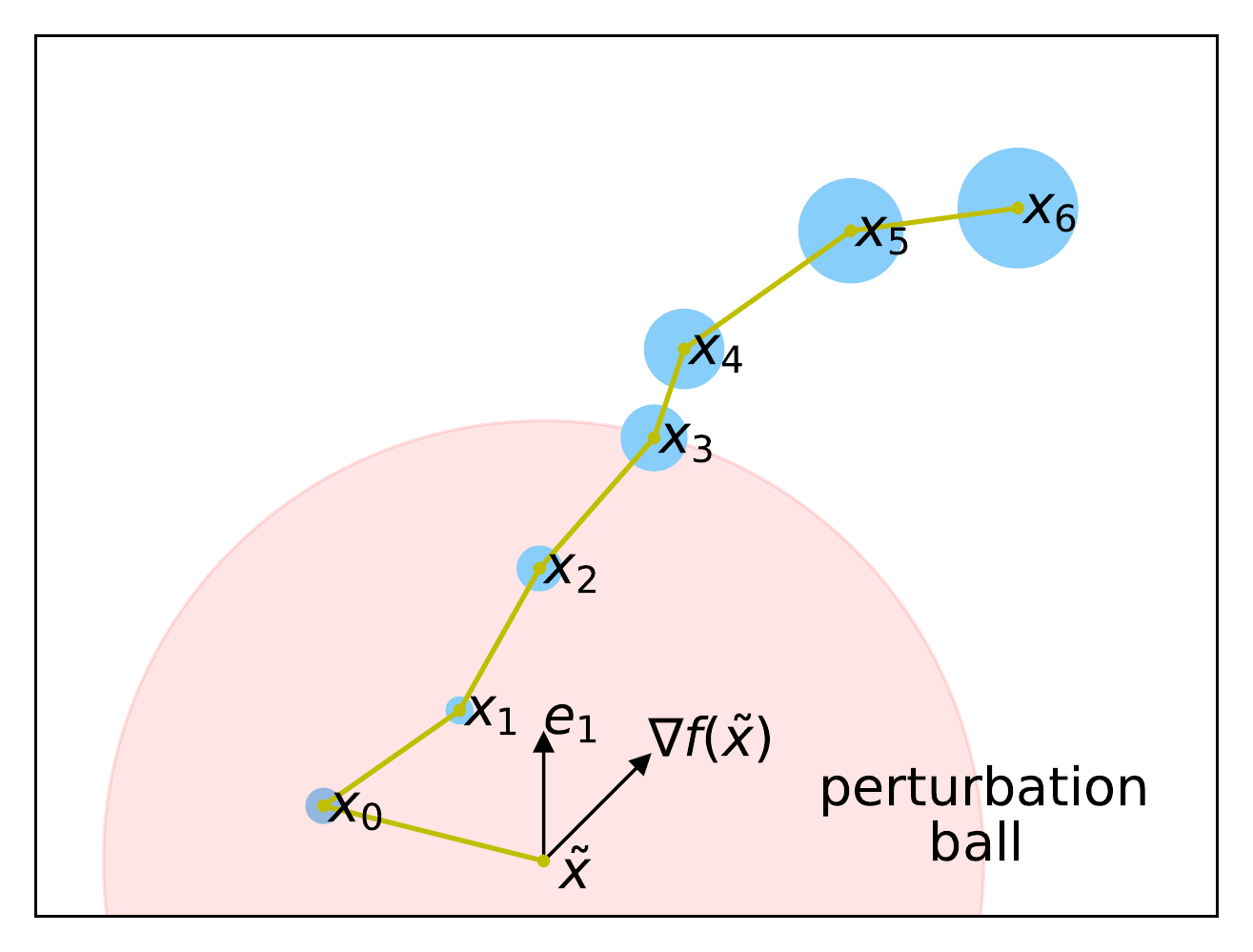}
\end{subfigure}
\begin{subfigure}
  \centering
  \includegraphics[width=0.45\textwidth]{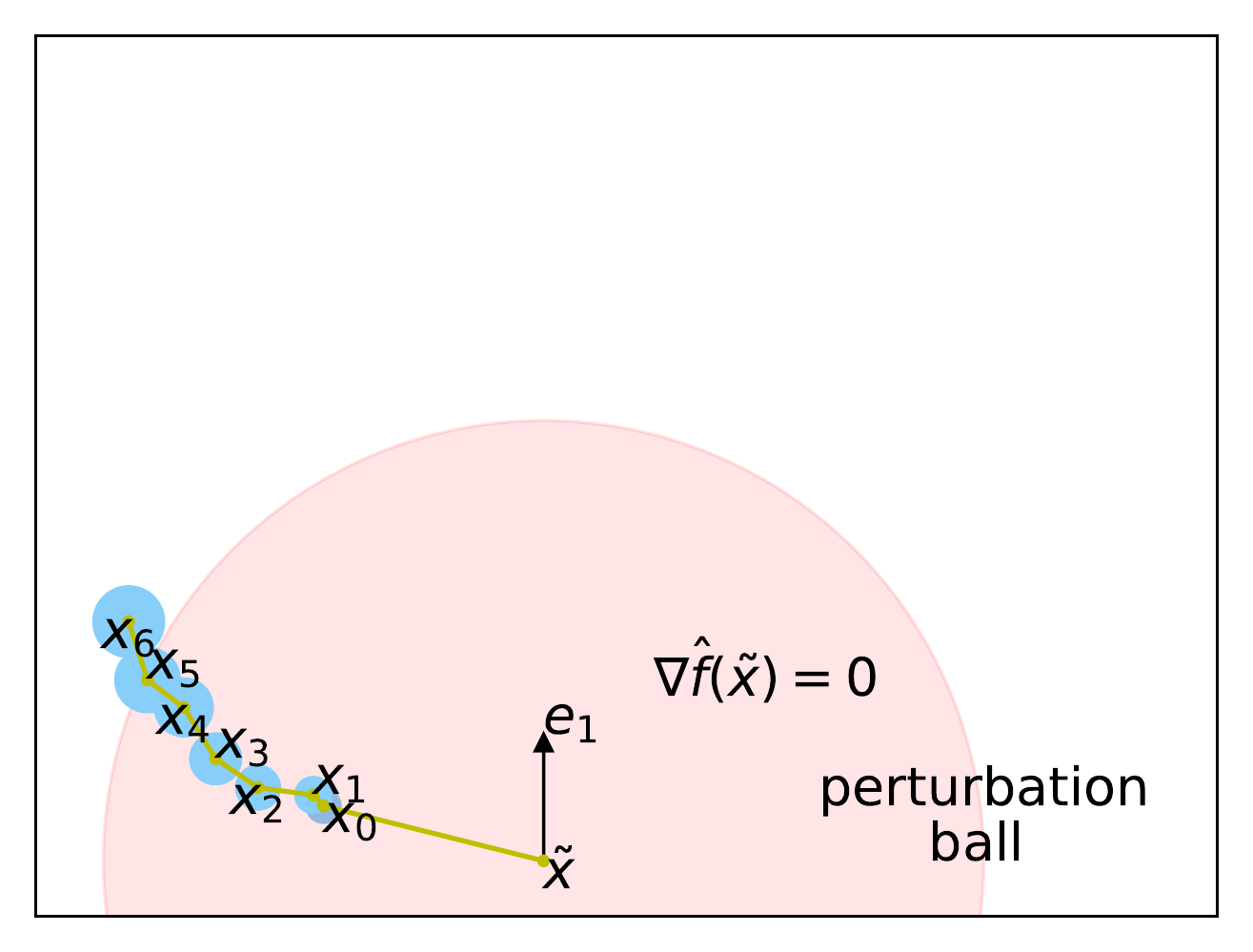}
\end{subfigure}
\caption{SVRG trajectories on the original function $f$ and the stabilized function $\hat{f}$. The size of the blue circle at each point indicates the magnitude of the variance.}
\label{fig:stabilization}
\end{figure}

However, in Perturbed SVRG, the iterates are not going to stay close to the starting point $\tx$, as the initial point $\tx$ can have a non-negligible gradient that will make the iterates travel a significant distance (see Figure~\ref{fig:stabilization} (a)). To fix this problem, we make a simple change to the function to set the gradient at $\tx$ equal to 0. More precisely, define the stabilized function $\hat{f}(x):=f(x)-\inner{\nabla f(\tx)}{x-\tx}$. After this stabilization, at least the first few iterates will not travel very far (see Figure~\ref{fig:stabilization} (b)). Our algorithm will apply SVRG on this stabilized function. 

\begin{figure}
\centering
  \includegraphics[width=0.45\textwidth]{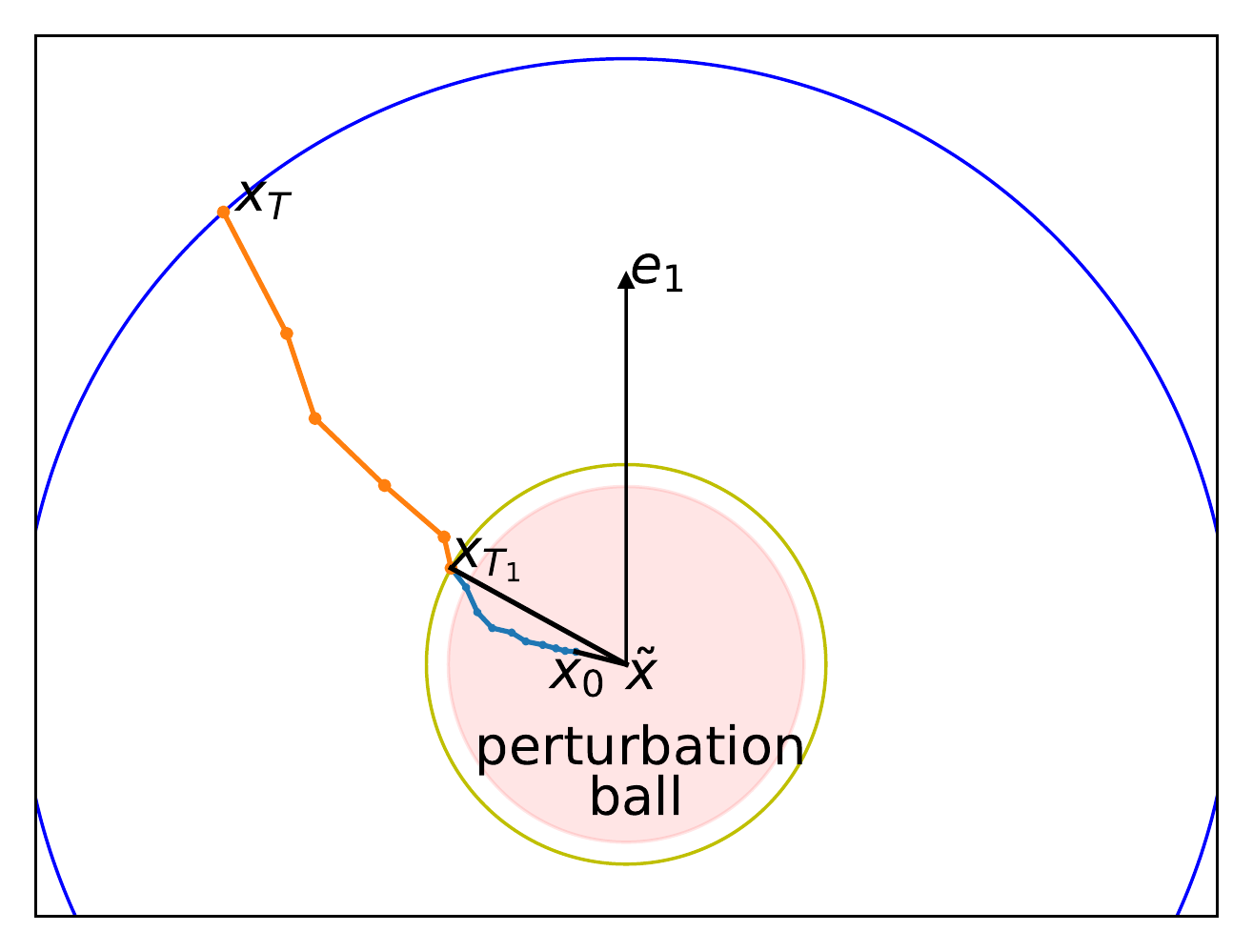}
\caption{Two phases of a super epoch in Stabilized SVRG}
\label{fig:twophase}
\end{figure}

For the stabilized function $\hat{f}(x)$, we have $\nabla \hat{f}(\tx) = 0$, so $\tx$ is an exact first-order stationary point. In this case, suppose the initial radius of perturbation $\delta$ is small, we will show that the behavior of the algorithm has two phases. In Phase 1, the iterates will remain in a ball around $\tx$ whose radius is $\tdo(\delta)$, which allows us to have very tight bounds on the variance and the potential changes in the Hessian. By the end of Phase 1, we show that the projection in the negative eigendirections of $\hessian = \nabla^2 f(\tx)$ is already at least $\tdomega(\delta)$. This means that Phase 1 has basically done a negative curvature search without a separate subroutine! 
Using the last point of Phase 1 as a good initialization, in Phase 2 we show that the point will eventually escape. See Figure~\ref{fig:twophase} for the two phases. 

The rest of the subsection will describe the two phases in more details in order to prove the following main lemma:

\begin{lemma}[informal]\label{lem:lemma2_informal}
Let $\tx$ be the initial point with gradient $\n{\nabla f(\tx)}\leq \mathG$ and $\lambda_{\min}(\hessian)=-\gamma<0$. 
Let $\{x_t\}$ be the iterates of SVRG running on $\hat{f}$ starting from $x_0$, which is the perturbed point of $\tx$. Let $T$ be the length of the current super epoch. Setting the parameters appropriately we know with probability at least $1/8$, $f(x_T)-f(\tx)\leq -C_5 \frac{\gamma^3}{\rho^2};$ and with high probability, $f(x_T)-f(\tx)\leq \frac{C_5}{20} \frac{\gamma^3}{\rho^2},$ where $T=\tdo(\frac{1}{\eta\gamma}), C_5=\tdtheta(1).$

\end{lemma}

Basically, this lemma shows that starting from a saddle point, with constant probability the function value decreases by $\tdomega( \frac{\gamma^3}{\rho^2})$ after a super epoch; with high probability, the function value does not increase by more than $\tdo( \frac{\gamma^3}{\rho^2})$. The precise statement of this lemma is given in Lemma~\ref{lem:lemma2} in Appendix~\ref{sec:proofsecondorder_stabilized}. Proofs for lemmas in this section are deferred to Appendix~\ref{sec:proofsecondorder_stabilized}.


\subsubsection{Analysis of Phase 1}
Let $S$ be the subspace spanned by all the eigenvectors of $\hessian$ with eigenvalues at most $-\frac{\gamma}{\log(d)}$. Our goal is to show that by the end of Phase 1, the projection of $x_t-\tx$ on subspace $S$ becomes large while the total movement $\n{x_t-\tx}$ is still bounded. To prove this, we use the following conditions to define Phase 1:

\paragraph{Stopping Condition:} An iterate $x_t$ is in Phase 1 if (1) $t\le 1/\eta\gamma$ or (2) $\n{\proj{S}(x_{t}-\tx)}\leq \frac{\delta}{10}$.
%
\medskip

If both conditions break, Phase 1 has ended. Intuitively, the second condition guarantees that the projection of $x_t-\tx$ on subspace $S$ is large at the end of Phase 1. The first condition makes sure that Phase 1 is long enough such that the projection of $x_t-x_{t-1}$ along positive eigendirections of $\hessian$ has shrunk significantly, which will be crucial in the analysis of Phase 2.

With the above two conditions, the length of Phase 1 can be defined as 
\begin{equation}
T_1=\sup\left\{t|\forall t'\leq t-1, \left(t'\leq \frac{1}{\eta\gamma} \right) \vee\left(\n{\proj{S}(x_{t'}-\tx)}\leq \frac{\delta}{10} \right) \right\}.\label{eq:phase1length}
\end{equation}

The main lemma for Phase 1 gives the following guarantee:

\begin{lemma}[informal]\label{lem:phase1informal}
By choosing $\eta = \tdo(1/L)$, $b = \tdo(n^{2/3})$ and $\delta=\tdo(\min(\frac{\gamma}{\rho}, \frac{m\gamma}{\rho'}))$, with constant probability, the length of the first phase $T_1$ is $\tdtheta(1/\eta\gamma)$ and 
$$\n{x_{T_1}-\tx}\leq \tdo(\delta) \ \mbox{and}\ \n{\proj{S}(x_{T_1}-\tx)}\geq \frac{1}{10}\delta.$$  
\end{lemma}

We will first show that the iterates in Phase 1 cannot go very far from the initial point:
\begin{lemma}[informal]\label{lem:phase1_informal}
Let $T_1$ be the length of Phase 1. Setting parameters appropriately we know with high probability 
$\n{x_t-x_{t-1}}\leq \tdo(\frac{1}{t})\delta$ for every $1\leq t\leq \min(T_1,\frac{\log(d)}{\eta\gamma}).$
\end{lemma}

The formal version of the above lemma is in Lemma~\ref{lem:phase1}. Taking the sum over all $t$ and note that $\sum_{t=1}^T 1/t = \Theta(\log T)$, this implies that the iterates are constrained in a ball whose radius is not much larger than $\delta$. If we choose $\delta$ to be small enough, within this ball Lemma~\ref{lem:coupled_gradient_estimator} will give very sharp bounds on the variance of the gradient estimates. This allows us to repeat the two-point analysis in Section~\ref{sec:negativecurvatureperturb} and prove that at least one sequence must have a large projection on $S$ subspace within $\frac{\log(d)}{\eta\gamma}$ steps. Recall that in the two point analysis, we consider two coupled samples of the perturbed points $x_0,x_0'$. The two perturbed points $x_0$ and $x_0'$ only differ in the $e_1$ direction. These two sequences $\{x_t\}$ and $\{x_t'\}$ share the same choice of mini-batches at each step. Basically, we prove after $\frac{\log(d)}{\eta\gamma}$ steps, the difference between two sequences along $e_1$ direction becomes large, which implies that at least one sequence must have large distance to $\tx$ on $S$ subspace. The formal version of the following lemma is in Lemma~\ref{lem:distanceproperty}.

\begin{lemma}[informal]\label{lem:distanceproperty_informal}
Let $\{x_t\}$ and $\{x_t'\}$ be two SVRG sequences running on $\hat{f}$ that use the same choice of mini-batches. Assume $w_0=x_0-x_0'$ aligns with $e_1$ direction and $|\inner{e_1}{w_0}|\geq \frac{\delta}{4\sqrt{d}}.$ Let $T_1, T_1'$ be the length of Phase 1 for $\{x_t\}$ and $\{x_t'\}$ respectively. Setting parameters appropriately with high probability we have $\min(T_1, T_1')\leq \frac{\log(d)}{\eta\gamma}.$ W.l.o.g., suppose $T_1\leq \frac{\log(d)}{\eta\gamma}$ and we further have $\n{x_{T_1}-\tx}\leq \tdo(1)\delta,\ \n{\proj{S}(x_{T_1}-\tx)}\geq \frac{1}{10}\delta.$
\end{lemma}

\paragraph{Remark:} We note that the guarantee of Lemma~\ref{lem:distanceproperty_informal} for Phase 1 is very similar to the guarantee of a negative curvature search subroutine: we find a direction $x_{T_1}-\tx$ that has a large projection in subspace $S$, which contains only the very negative eigenvectors of $\mathcal{H}$.


\subsubsection{Analysis of Phase 2}
By the guarantee of Phase 1, we know if it is successful $x_{T_1}-\tx$ has a large projection in subspace $S$ of very negative eigenvalues. 
Starting from such a point, in Phase 2 we will show that the projection of $x_t-\tx$ in $S$ grows exponentially and exceeds the threshold distance within $\tdo(\frac{1}{\eta\gamma})$ steps. 
In order to prove this, we use the following expansion,
\begin{align*}
x_t-\tx = (I-\eta\hessian)(x_{t-1}-\tx)-\eta\Delta_{t-1}(x_{t-1}-\tx)-\eta\xi_{t-1},
\end{align*}
where $\Delta_{t-1}=\int_0^1 (\nabla^2 \hat{f}(\tx +\theta(x_{t-1}-\tx))-\hessian)d\theta.$
Intuitively, if we only have the first term, it's clear that $\n{\proj{S}(x_t-\tx)}\geq (1+\frac{\eta\gamma}{\log(d)})\n{\proj{S}(x_{t-1}-\tx)}$. The norm in subspace $S$ increases exponentially and will become very far from $\tx$ in a small number of iterations. Our proof bounds the Hessian changing term $\eta\Delta_{t-1}(x_{t-1}-\tx)$ and variance term $\eta\xi_{t-1}$ separately to show that they do not influence the exponential increase.
The main lemma that we will prove for Phase 2 is:

\begin{lemma}[informal]\label{lem:phase2_informal}
Assume Phase 1 is successful in the sense that $T_1\leq \frac{\log(d)}{\eta\gamma}$ and $\n{x_{T_1}-\tx}\leq \tdo(1)\delta,\\ \n{\proj{S}(x_{T_1}-\tx)}\geq \frac{1}{10}\delta$. Setting parameters appropriately with high probability we know there exists $T=\tdo(\frac{1}{\eta\gamma})$ such that $\n{x_T-\tx}\geq \tdomega(\frac{\gamma}{\rho}).$
\end{lemma}

The precise version of the above lemma is in Lemma~\ref{lem:phase2} in Appendix~\ref{sec:proofsecondorder_stabilized}. Similar to Lemma~\ref{lem:distanceproperty_psvrg_informal}, the lemma above shows that the iterates will escape from a local neighborhood if Phase 1 was successful (which happens with at least constant probability). We can then use Lemma~\ref{lem:dis_value_cross_epoch} to bound the function value decrease.


\subsection{Proof of Main Theorems}\label{sec:proofmaintheorem}

Finally we are ready to sketch the proof for  Theorem~\ref{thm:maintheorem}. For each epoch, if the gradients are large, by Lemma~\ref{lem:large_gradients} we know with constant probability the function value decreases by at least $\tdomega(n^{1/3}\epsilon^2/L)$. For each super epoch, if the starting point has significant negative curvature, by Lemma~\ref{lem:lemma2_informal}, we know with constant probability the function value decreases by at least $\tdomega(\epsilon^{1.5}/\sqrt{\rho}).$ We also know that the number of stochastic gradient for each epoch is $\tdo(n)$ and that for each super epoch is $\tdo(n+n^{2/3}L/\sqrt{\rho\epsilon})$. Thus, we know after 
$$\tdo\left(\frac{L \Delta f}{n^{1/3}\epsilon^2}\cdot n+\frac{\sqrt{\rho}\Delta f}{\epsilon^{1.5}}\cdot (n+ \frac{n^{2/3}L}{\sqrt{\rho\epsilon}})\right)$$
stochastic gradients, the function value will decrease below the global optimal $f^*$ with high probability unless we have already met an $\epsilon$-second-order stationary point. Thus, we will at least once get to an $\epsilon$-second-order stationary point within $\tdo(\frac{n^{2/3}L\Delta f}{\epsilon^2}+\frac{n\sqrt{\rho}\Delta f}{\epsilon^{1.5}})$ stochastic gradients. The formal proof of Theorem~\ref{thm:maintheorem} is deferred to Appendix~\ref{sec:maintheoremproof}. The proof for Theorem~\ref{thm:maintheorem_psvrg} is almost the same except that it uses Lemma~\ref{lem:distanceproperty_psvrg_informal} instead of Lemma~\ref{lem:lemma2_informal} for the guarantee of the super epoch. 




\section{Conclusion}

This paper gives a new algorithm Stabilized SVRG that is able to find an $\epsilon$-second-order stationary point using $\tdo(\frac{n^{2/3}L\Delta f}{\epsilon^2}+\frac{n\sqrt{\rho}\Delta f}{\epsilon^{1.5}})$ stochastic gradients. To our best knowledge this is the first algorithm that does not rely on a separate negative curvature search subroutine, and it is much simpler than all existing algorithms with similar guarantees. In our proof, we developed the new technique of stabilization (Section~\ref{subsec:stabilization}), where we showed if the initial point has exactly 0 gradient and the initial perturbation is small, then the first phase of the algorithm can achieve the guarantee of a negative curvature search subroutine. We believe the stabilization technique can be useful for analyzing other optimization algorithms in nonconvex settings without using an explicit negative curvature search. We hope techniques like this will allow us to develop nonconvex optimization algorithms that are as simple as their convex counterparts.
\section*{Acknowledgement}
This work was supported by NSF CCF-1704656. 

\clearpage

\bibliography{ms}

\appendix

\section{Detailed Descriptions of Our Algorithm}

In this section, we give the complete descriptions of the Perturbed SVRG and Stabilized SVRG algorithms. 

\paragraph{Perturbed SVRG}

Perturbed SVRG is given in Algorithm~\ref{alg:psvrg}. The only difference of this algorithm with the high level description in Algorithm~\ref{alg:psvrg-simple} is that we have now stated the stopping condition explicitly, and when the algorithm is not running a super epoch, we choose a random iterate as the starting point of the next epoch (this is necessary because of the guarantee in Lemma~\ref{lem:gradientfunctionvalue}).

In the algorithm, the break probability in Step 16 is used to implement the random stopping. Breaking the loop with this probability is exactly equivalent to finishing the loop and sampling $x_{sm+t}$ for $t = 1,2,...,m$ uniformly at random. 

\begin{algorithm}[!htb]
	\caption{Perturbed SVRG($x_0, m, b, \eta,\delta,T_{\max} ,\mathG, \mathL$)}
	\label{alg:psvrg}
	\begin{algorithmic}[1]
		\REQUIRE 
		initial point $x_0$, epoch length $m$, minibatch size $b$, step size $\eta$, perturbation radius $\delta$, super-epoch length $T_{\max}$, threshold gradient $\mathG$, threshold length $\mathL$
		\STATE $super\_epoch \leftarrow 0.$
		\FOR {$s=0,1,2,\cdots$}
			\STATE Compute $\nabla f(x_{sm})$.
			\IF{$super\_epoch = 0\ \wedge\ \n{\nabla f(x_{sm})}\leq \mathG$}
				\STATE {$super\_epoch \leftarrow 1.$}
				\STATE {$\tx\leftarrow x_{sm}, t_{init}\leftarrow sm.$}
				\STATE $x_{sm}\leftarrow x_{sm}+\xi,$ where $\xi$ uniformly $\sim \mathbb{B}_0(\delta).$
			\ENDIF
			\FOR {$t=1, 2, \cdots, m$}
\STATE Sample $b$ i.i.d. numbers uniformly from $[n]$ and form a multi-set $I_b$.
    	    	\STATE $v_{sm+t-1}\leftarrow \frac{1}{b}\sum_{i\in I_b}\big(\nabla f_i(x_{sm+t-1})-\nabla f_i(x_{sm}) + \nabla f(x_{sm})) \big)$.
        	   	\STATE $x_{sm+t} \leftarrow x_{sm+t-1} - \eta v_{sm+t-1}$.
        	   	\IF {$super\_epoch=1\ \wedge\ \big(\n{x_{sm+t}-\tx}\geq \mathL\ \vee\ sm+t-t_{init}\geq T_{\max}\big)$}
        	   		\STATE $super\_epoch \leftarrow 0;$ Break.
				\ELSIF{$super\_epoch=0$}        	   		
        	   		\STATE {Break with probability $\frac{1}{m-(t-1)}$.}
        	   	\ENDIF 
			\ENDFOR
			\STATE {$x_{(s+1)m}\leftarrow x_{sm+t}.$}
		\ENDFOR
    \end{algorithmic}
\end{algorithm}

\paragraph{Stabilized SVRG} Stabilized SVRG is given in Algorithm~\ref{alg:ssvrg}. The only differences between Stabilized SVRG and Perturbed SVRG is that Stabilized SVRG adds an additional shift of $-\nabla f(\tilde{x})$ when it is in a super epoch ($stabilizing = 1$ in the algorithm).  

\begin{algorithm}[!htb]
	\caption{Stabilized SVRG($x_0, m, b, \eta,\delta,T_{\max} ,\mathG, \mathL$)}
	\label{alg:ssvrg}
	\begin{algorithmic}[1]
		\REQUIRE 
		initial point $x_0$, epoch length $m$, minibatch size $b$, step size $\eta$, perturbation radius $\delta$, super-epoch length $T_{\max}$, threshold gradient $\mathG$, threshold length $\mathL$
		\STATE $stabilizing\leftarrow 0.$
		\FOR {$s=0,1,2,\cdots$}
			\STATE Compute $\nabla f(x_{sm})$.
			\IF{$stabiling = 0\ \wedge\ \n{\nabla f(x_{sm})}\leq \mathG$}
				\STATE {$stabilizing \leftarrow 1.$}
				\STATE {$\tx\leftarrow x_{sm}, t_{init}\leftarrow sm.$}
				\STATE $x_{sm}\leftarrow x_{sm}+\xi,$ where $\xi$ uniformly $\sim \mathbb{B}_0(\delta).$
			\ENDIF
			\FOR {$t=1, 2, \cdots, m$}
\STATE Sample $b$ i.i.d. numbers uniformly from $[n]$ and form a multi-set $I_b$.
    	    	\STATE $v_{sm+t-1}\leftarrow \frac{1}{b}\sum_{i\in I_b}\big(\nabla f_i(x_{sm+t-1})-\nabla f_i(x_{sm}) + \nabla f(x_{sm})) \big) -stabilizing\times \nabla f(\tx)$.
    	    	
        	   	\STATE $x_{sm+t} \leftarrow x_{sm+t-1} - \eta v_{sm+t-1}$.
        	   	\IF {$stabilizing=1\ \wedge\ \big(\n{x_{sm+t}-\tx}\geq \mathL\ \vee\ sm+t-t_{init}\geq T_{\max}\big)$}
        	   		\STATE $stabilizing \leftarrow 0;$ Break.
				\ELSIF{$stabilizing=0$}        	   		
        	   		\STATE {Break with probability $\frac{1}{m-(t-1)}$.}
        	   	\ENDIF 
			\ENDFOR
			\STATE {$x_{(s+1)m}\leftarrow x_{sm+t}.$}
		\ENDFOR
    \end{algorithmic}
\end{algorithm}

\clearpage
\section{Proofs of Exploiting Large Gradients}
\label{sec:proofgradients}
In this section, we adapt the proof from \cite{li2018simple} to show that SVRG can reduce the function value when the gradient is large. First, we give guarantees on the gradient estimate (Lemma~\ref{lem:varaincedistance}). Note that previously such bounds were known in the expectation sense, here we convert the bounds to a high probability bound by applying a vector Bernstein's inequality (Lemma~\ref{vectorBernstein}).

\begingroup
\def\thetheorem{\ref{lem:varaincedistance}}
\begin{lemma}
For any point $x_t$, let the gradient estimate be $v_t:=\frac{1}{b}\sum_{i\in I_b}(\nabla f_i(x_t)-\nabla f_i(x_{s(t)})+\nabla f(x_{s(t)}))$, where $x_{s(t)}$ is the snapshot point of the current epoch. Then, with probability at least $1-\zeta$, we have
$$\n{v_t-\nabla f(x_t)}\leq O\Big(\frac{\log(d/\zeta)L}{\sqrt{b}}\Big)\n{x_t-x_{s(t)}}.$$ 
\end{lemma}
\addtocounter{theorem}{-1}
\endgroup
\begin{proofof}{Lemma~\ref{lem:varaincedistance}}
In order to apply Bernstein inequality, we first show for each $i$, the norm of $(\nabla f_i(x_t)-\nabla f_i(\snap)+\nabla f(\snap)-\nabla f(x_t))$ is bounded. 
\begin{align*}
&\n{\nabla f_i(x_t)-\nabla f_i(\snap)+\nabla f(\snap)-\nabla f(x_t)} \\
=&  \n{\nabla f(x_t)-\nabla f(\snap)-(\nabla f_i(x_t)-\nabla f_i(\snap))}\\
\leq& \n{\nabla f(x_t)-\nabla f(\snap)}+\n{(\nabla f_i(x_t)-\nabla f_i(\snap))}\\ 
\leq& 2L\n{x_t-\snap},
\end{align*}
where the last inequality is due to the smoothness of $f$ and $f_i$.

Then, we bound the summation of variance of each term as follows.
\begin{align*}
\sigma^2
&:=\sum_{i\in I_b}\E[\ns{\nabla f(x_t)-\nabla f(\snap)-(\nabla f_i(x_t)-\nabla f_i(\snap))}]\\
&\leq \sum_{i\in I_b}\E[\ns{\nabla f_i(x_t)-\nabla f_i(\snap)}]\\
&\leq \sum_{i\in I_b}L^2\ns{x_t-\snap}\\
&= bL^2\ns{x_t-\snap},
\end{align*}
where the first inequality is due to $\E[\ns{X-\E[X]}]\leq \E[X^2]$ and the second inequality holds because the gradient of $f_i$ is $L$-Lipschtiz.

Then, according to the vector version Bernstein inequality (Lemma~\ref{vectorBernstein}), we have 
$$\Pr[\n{b v_t-b\nabla f(x_t)}\geq r]\leq (d+1)\exp\Big(\frac{-r^2/2}{bL^2\ns{x_t-\snap}+\frac{2L\n{x_t-\snap}\cdot r}{3}}\Big)$$
Thus, with probability at least $1-\zeta$, we have 
$$\n{v_t-\nabla f(x_t)}\leq O\Big(\frac{\log(d/\zeta)L}{\sqrt{b}}\Big)\n{x_t-\snap},$$
where $O(\cdot)$ hides constants.
\end{proofof}


Using this upperbound on the error of gradient estimates, we can then show that the function value decreases as long as the norms of gradients are large along the path. Note that this part of the proof is also why we require $b \ge m^2$, which results in the $n^{2/3}$ term in the running time.
\begingroup
\def\thetheorem{\ref{lem:gradientfunctionvalue}}
\begin{lemma}
For any epoch, suppose the initial point is $x_0$, which is also the snapshot point for this epoch. Assume for any $0\leq t\leq m-1$, $\n{v_t-\nabla f(x_t)}\leq \frac{C_1 L}{\sqrt{b}}\n{x_t-x_0},$ where $C_1=\tdo(1)$ comes from Lemma~\ref{lem:varaincedistance}. Then, given $\eta\leq \frac{1}{3C_1 L}, b\geq m^2$, we have 
$$f(x_0)-f(x_t)\geq \sum_{\tau=0}^{t-1}\frac{\eta}{2}\ns{\nabla f(x_\tau)}$$
for any $1\leq t\leq m$.
\end{lemma}
\addtocounter{theorem}{-1}
\endgroup
\begin{proofof}{Lemma~\ref{lem:gradientfunctionvalue}}
First, we obtain the relation between $f(x_t)$ and $f(x_{t-1})$ as follows. For any $1\leq t\leq m$, 
\begin{align}
f(x_t) \leq& f(x_{t-1})
                    + \inner{\nabla f(x_{t-1})}{x_t-x_{t-1}}
                    + \frac{L}{2}\ns{x_t-x_{t-1}} \label{eq:lp} \\
         =&f(x_{t-1})
                    + \inner{\nabla f(x_{t-1})-v_{t-1}}{x_t-x_{t-1}}
                    + \inner{v_{t-1}}{x_t-x_{t-1}}
                    + \frac{L}{2}\ns{x_t-x_{t-1}} \notag \\
         =&f(x_{t-1})
                    + \inner{\nabla f(x_{t-1})-v_{t-1}}{-\eta v_{t-1}}
                    - \big(\frac{1}{\eta}- \frac{L}{2}\big)\ns{x_t-x_{t-1}} \label{eq:plug} \\
         =&f(x_{t-1})
                    + \eta\ns{\nabla f(x_{t-1})-v_{t-1}}
                    - \eta\inner{\nabla f(x_{t-1})-v_{t-1}}{\nabla f(x_{t-1})}\notag
                    \\
                    &- \big(\frac{1}{\eta}- \frac{L}{2}\big)\ns{x_t-x_{t-1}} \notag \\
         =&f(x_{t-1})
                    + \eta\ns{\nabla f(x_{t-1})-v_{t-1}}
                    - \frac{1}{\eta}\inner{x_t-\bx_t}{x_{t-1}-\bx_t}
                    - \big(\frac{1}{\eta}- \frac{L}{2}\big)\ns{x_t-x_{t-1}} \label{eq:plug2} \\
         =&f(x_{t-1})
                    + \eta\ns{\nabla f(x_{t-1})-v_{t-1}}
                    - \big(\frac{1}{\eta}- \frac{L}{2}\big)\ns{x_t-x_{t-1}} \notag\\
            &\qquad\qquad\qquad -\frac{1}{2\eta}\big(\ns{x_t-\bx_t}+\ns{x_{t-1}-\bx_t}
                            -\ns{x_t-x_{t-1}}\big) \notag \\
         =&f(x_{t-1})
                    + \frac{\eta}{2}\ns{\nabla f(x_{t-1})-v_{t-1}}
                    - \frac{\eta}{2}\ns{\nabla f(x_{t-1})}
                    - \big(\frac{1}{2\eta}- \frac{L}{2}\big)\ns{x_t-x_{t-1}},
\end{align}
where (\ref{eq:lp}) holds due to smoothness condition, and (\ref{eq:plug}) and (\ref{eq:plug2}) follow from these two definitions, 
i.e., $x_t := x_{t-1}-\eta v_{t-1}$
and $\bx_t := x_{t-1}-\eta \nabla f(x_{t-1})$.

According to the assumption, we have $\ns{\nabla f(x_{t-1})-v_{t-1}}\leq \frac{C_1^2 L^2 }{b}\ns{x_{t-1}-x_0}$. Choosing $\eta\leq \frac{1}{3C_1 L}$, we have 

\begin{align*}
f(x_t) &\leq f(x_{t-1})
                    + \frac{\eta L^2 C_1^2}{2b}\ns{x_{t-1}-x_{0}}
                    - \frac{\eta}{2}\ns{\nabla f(x_{t-1})}
                    - \big(\frac{1}{2\eta}- \frac{L}{2}\big)\ns{x_t-x_{t-1}} \notag\\
        &\leq f(x_{t-1})
                    + \frac{LC_1}{6b}\ns{x_{t-1}-x_{0}}
                    - \frac{\eta}{2}\ns{\nabla f(x_{t-1})}
                    - LC_1\ns{x_t-x_{t-1}} \\
        &\leq f(x_{t-1})
                    + \big(\frac{L}{6b}+\frac{L}{2t-1}\big)C_1\ns{x_{t-1}-x_{0}}
                    - \frac{\eta}{2}\ns{\nabla f(x_{t-1})}
                    - \frac{L}{2t}C_1\ns{x_t-x_{0}},  
\end{align*}
where the last inequality uses Young's inequality
$\ns{x_t-x_{0}}\leq \big(1+\frac{1}{\alpha}\big)\ns{x_{t-1}-x_{0}}
+(1+\alpha)\ns{x_t-x_{t-1}}$ by choosing $\alpha=2t-1$.

Now, adding the above inequalities for all iterations $1\le t\le t'$, where $t'\leq m$, 
\begin{align}
f(x_{t'}) \leq& f(x_0)
                    -\sum_{t=1}^{t'}\frac{\eta}{2}\ns{\nabla f(x_{t-1})}
                    -\sum_{t=1}^{t'}\frac{L}{2t}C_1 \ns{x_t-x_{0}} \notag\\
                    &+\sum_{t=1}^{t'}\big(\frac{L}{6b}+\frac{L}{2t-1}\big)C_1 \ns{x_{t-1}-x_{0}} \notag \\
        =& f(x_0)
                    -\sum_{t=1}^{t'}\frac{\eta}{2}\ns{\nabla f(x_{t-1})}
                    -\sum_{t=1}^{t'-1}\big(\frac{L}{2t}-\frac{L}{6b}-\frac{L}{2t+1}\big)C_1 
                        \ns{x_t-x_{0}}\notag\\
                    &-\frac{L}{2t'}C_1\ns{x_{t'}-x_{0}} \notag\\
        \leq& f(x_0)
                    -\sum_{t=1}^{t'}\frac{\eta}{2}\ns{\nabla f(x_{t-1})}
                    -\frac{L}{2t'}C_1\ns{x_{t'}-x_{0}} \label{eq:key} 
\end{align}
where (\ref{eq:key}) holds because $\frac{L}{2t}-\frac{L}{6b}-\frac{L}{2t+1}\geq 0$ for any $1\leq t\leq m$ as long as $b\geq m^2.$

Thus, for any $1\leq t'\leq m$, we have 
$$f(x_0)-f(x_{t'})\geq \sum_{\tau=0}^{t'-1}\frac{\eta}{2}\ns{\nabla f(x_\tau)}.$$
\end{proofof}

A limitation of Lemma~\ref{lem:gradientfunctionvalue} is that it only guarantees function value decrease when the {\em sum} of squared gradients is large. However, in order to connect the guarantees between first and second order steps, we want to identify a single iterate that has a small gradient. We achieve this by stopping the SVRG iterations at a uniformly random location.

\begingroup
\def\thetheorem{\ref{lem:large_gradients}}
\begin{lemma}
For any epoch, suppose the initial point is $x_0$. Let $x_t$ be a point uniformly sampled from $\{x_\tau \}_{\tau=1}^m$. Then, given $\eta=\tdtheta(1/L), b\geq m^2$, for any value of $\mathG$, we have two cases:
\begin{enumerate}
\item if at least half of points in $\{x_\tau \}_{\tau=1}^m$ have gradient no larger than $\mathG,$ we know $\|\nabla f(x_t) \| \le \mathG$ holds with probability at least $1/2$;
\item Otherwise, we know $f(x_0) - f(x_t) \ge \frac{\eta}{2}\frac{m\mathG^2}{4}$ holds with probability at least $1/5.$ 
\end{enumerate}
Further, no matter which case happens we always have $f(x_t) \le f(x_0)$ with high probability. 
\end{lemma}
\addtocounter{theorem}{-1}
\endgroup
\begin{proofof}{Lemma~\ref{lem:large_gradients}}
Let $\{x_\tau\}_{\tau=0}^m$ be the iterates of SVRG starting from $x_0$. Then, there are two cases:
\begin{itemize}
\item If at least half of points of $\{x_\tau \}_{\tau=1}^{m}$ have gradient norm at most $\mathG$, then it's clear that a uniformly sampled point $x_t$ has gradient norm $\n{\nabla f(x_t)}\leq \mathG$ with probability at least $1/2.$
\item Otherwise, we know at least half of points from $\{x_\tau\}_{\tau =1}^{m}$ has gradient norm larger than $\mathG$. Then, as long as the sampled point falls into the last quarter of $\{x_\tau \}_{\tau =1}^{m}$, we know $\sum_{\tau=0}^{t-1}\ns{\nabla f(x_\tau)}\geq \frac{m\mathG^2}{4}.$ Thus, for a uniformly sampled point $x_t$, with probability at least $1/4$, we have $\sum_{\tau=0}^{t-1}\ns{\nabla f(x_\tau)}\geq \frac{m\mathG^2}{4}$. 

According to Lemma~\ref{lem:varaincedistance} and the union bound, we know there exists $C_1=\tdo(1)$ such that with high probability, $\n{v_t-\nabla f(x_t)}\leq \frac{C_1 L}{\sqrt{b}}\n{x_t-x_0}$ holds for every $0\leq  t\leq m-1$. Combining with Lemma~\ref{lem:gradientfunctionvalue}, we know given $\eta\leq \frac{1}{3C_1 L},b\geq m^2$, we have $f(x_0)-f(x_t)\geq \sum_{\tau=0}^{t-1}\frac{\eta}{2}\ns{\nabla f(x_\tau)}$ for any $1\leq t\leq m$. By another union bound, we know with probability at least $1/5$, $f(x_0)-f(x_t)\geq \frac{\eta}{2}\frac{m\mathG^2}{4}$.
\end{itemize}

Again by Lemma~\ref{lem:varaincedistance} and Lemma~\ref{lem:gradientfunctionvalue}, we know $f(x_t)\leq f(x_0)$ holds with high probability.
\end{proofof}

\section{Proofs of Exploiting Negative Curvature - Perturbed SVRG}\label{sec:psvrg}
In this section, we show that starting from a point with negative curvature, Perturbed SVRG can decrease the function value significantly after a super epoch. 

As discussed in Section~\ref{sec:negativecurvatureperturb}, we use two point analysis to show that with good probability one of these two points can escape the saddle point. Let $\tx$ be the initial point of the super epoch. We consider two coupled samples of the perturbed point $x_0,x_0'$. 
The two perturbed points $x_0$ and $x_0'$ only differ in the $e_1$ direction, where $e_1$ is the smallest eigendirection of Hessian $\hessian := \nabla^2 f(\tx)$.  
Let the SVRG iterates running on $f$ starting from $x_0$ and $x_0'$ be $\{x_t\}$ and $\{x_t'\}$ respectively. 
We will keep track of the difference between the two sequences $w_t=x_t-x_t'$, and show that $w_t$ increases exponentially and becomes large after one super epoch, which means at least one sequence must escape the initial point $\tx$.  

In the following proof, we first show that the variance of $w_t$ can be well bounded. This is the place where we use the assumption that each individual function is $\rho'$-Hessian Lipschitz.
\begingroup
\def\thetheorem{\ref{lem:coupled_gradient_estimator}}
\begin{lemma}
Let $\{x_t\}$ and $\{x_t'\}$ be two SVRG sequences running on $f$ that use the same choice of mini-batches. Let $x_{s(t)}$ be the snapshot point for iterate $t$. Let $w_t:= x_t-x_t'$ and $P_t=\max(\n{x_{s(t)}-\tx}, \n{x_{s(t)}'-\tx}, \n{x_{t}-\tx}, \n{x_{t}'-\tx})$. Then, with probability at least $1-\zeta$, we have
$$\n{\xi_t-\xi_t'}\leq O\Big(\frac{\log(d/\zeta)}{\sqrt{b}}\Big)\min\Big( L\n{w_t-w_{s(t)}} +\rho'P_t(\n{w_t}+\n{w_{s(t)}}), L(\n{w_t}+\n{w_{s(t)}}) \Big).$$  
\end{lemma}
\addtocounter{theorem}{-1}
\endgroup

\begin{figure}
\centering
  \includegraphics[width=0.5\textwidth]{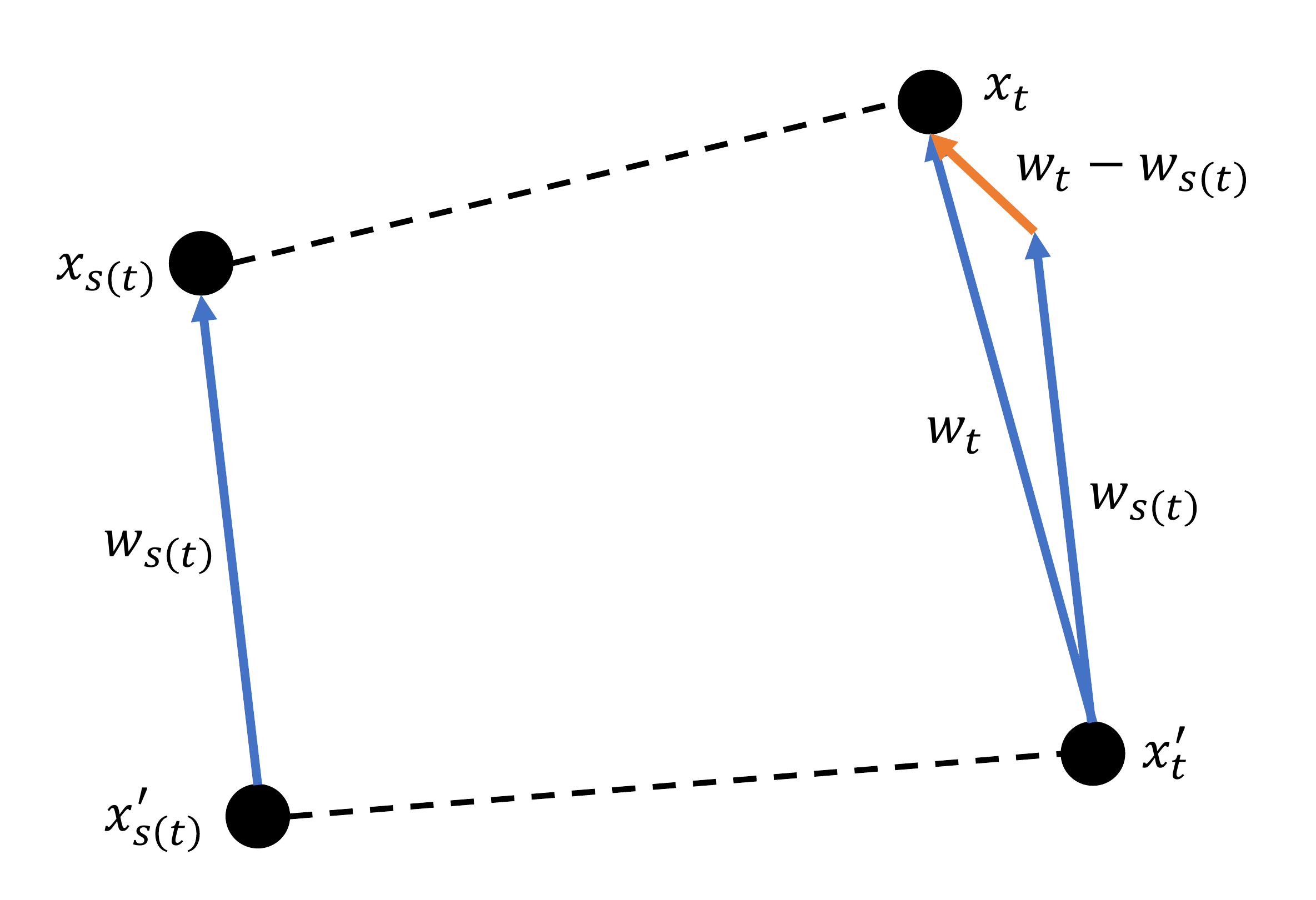}
\caption{Comparison between $\n{w_t-w_{s(t)}}$ and $\n{w_t}+\n{w_{s(t)}}$}
\label{fig:variance}
\end{figure}

In the extreme case, if each individual function $f_i$ is exactly a quadratic function, then we know $\rho' = 0$ and the variance is proportional to $\tdo(L/\sqrt{b})\|w_t - w_{s(t)}\|$. As illustrated in Figure~\ref{fig:variance},  $w_t$ cannot change very quickly within a single epoch so $\|w_t - w_{s(t)}\|$ is much smaller than $\|w_t\|$ or $\|w_{s(t)}\|$.


\begin{proofof}{Lemma~\ref{lem:coupled_gradient_estimator}}
Similar as the proof in Lemma~\ref{lem:varaincedistance}, here we use Bernstein inequality to prove that the difference between the variances of two coupled sequences is also upper bounded. 


Recall that,
\begin{align*}
\xi_t-\xi_t'=&(v_t-\nabla f(x_t))-(v_t'-\nabla f(x_t'))\\
=& \frac{1}{b}\sum_{i\in I_b}\Big( \big(\nabla f_i(x_t)-\nabla f_i(\snap)+\nabla f(\snap)-\nabla f(x_t)\big) \\
&-\big(\nabla f_i(x_t')-\nabla f_i(\snap')+\nabla f(\snap')-\nabla f(x_t')\big)\Big),
\end{align*}
where $I_b$ is a uniformly sampled multi-set of $[n]$ with size $b$. 

Let the Hessian of $f$ at $\tx$ be $\hessian$ and let the Hessian of $f_i$ at $\tx$ be $\hessian_i$ for each $i$. Let $\xi_{t,i}-\xi_{t,i}'$ be the $i$-th term in the above sum.
In order to apply Bernstein inequality, we first show for each $i$,
\begin{align*}
&\left\| \xi_{t,i}-\xi_{t,i}'\right\| \\
\leq& \left\| (\nabla f_i(x_t)-\nabla f_i(x_t') )-( \nabla f_i(\snap)- \nabla f_i(\snap'))     \right\| \\
&+ 
\left\| (\nabla f(x_t)-\nabla f(x_t') )-( \nabla f(\snap)- \nabla f(\snap'))\right\|\\
=& \left\| \int_0^1\nabla^2 f_i(x_t' + \theta(x_t-x_t'))d\theta(x_t-x_t') -\int_0^1\nabla^2 f_i(\snap' + \theta(\snap-\snap'))d\theta(\snap-\snap') \right\| \\
&+ \left\| \int_0^1\nabla^2 f(x_t' + \theta(x_t-x_t'))d\theta(x_t-x_t') -\int_0^1\nabla^2 f(\snap' + \theta(\snap-\snap'))d\theta(\snap-\snap') \right\|\\
=& \left\| \hessian_iw_t + \Delta_t^i w_t -(\hessian_iw_{s(t)} + \Delta_{s(t)}^i w_{s(t)})  \right\| 
+ \left\| \hessian w_t + \Delta_t w_t -(\hessian w_{s(t)} + \Delta_{s(t)} w_{s(t)})  \right\| \\
\leq & \n{\hessian_i}\n{w_t-w_{s(t)}} +\n{\Delta_t^i}\n{ w_t} + \n{\Delta_{s(t)}^i}\n{ w_{s(t)}} \\
&+ \n{\hessian}\n{w_t-w_{s(t)}} +\n{\Delta_t}\n{ w_t} + \n{\Delta_{s(t)}}\n{ w_{s(t)}}\\
\leq & 2L\n{w_t-w_{s(t)}}+ 2\rho' P_t (\n{w_t}+\n{w_{s(t)}})
\end{align*}
where $\Delta_t^i=\int_0^1(\nabla^2 f_i(x_t' + \theta(x_t-x_t'))-\hessian_i) d\theta(x_t-x_t')$ and $\Delta_t=\int_0^1(\nabla^2 f(x_t' + \theta(x_t-x_t'))-\hessian) d\theta(x_t-x_t')$. The last inequality holds since each individual function is $L$-smooth and $\rho'$ Hessian Lipschitz. Specifically, due to the $L$-smoothness, we have $\n{\hessian_i}, \n{\hessian}\leq L$. Because of the Hessian Lipschitz condition and the definition of $P_t$, we have 
$\n{\Delta_t^i},\n{\Delta_{s(t)}^i}, \n{\Delta_t}, \n{\Delta_{s(t)}}\leq \rho'P_t. $

Then, we bound the summation of variance of each term as follows.
\begin{align*}
&\sigma^2\\
:=&\sum_{i\in I_b}\E\left\| \xi_{t,i}-\xi_{t,i}'\right\|^2\\
\leq& \sum_{i\in I_b}\E\left[
\left\| (\nabla f_i(x_t)-\nabla f_i(x_t') )-( \nabla f_i(\snap)- \nabla f_i(\snap'))     \right\|^2
\right]\\
\leq& \sum_{i\in I_b}\left( L\n{w_t-w_{s(t)}}+ \rho' P_t (\n{w_t}+\n{w_{s(t)}})\right)^2\\
=& b\left( L\n{w_t-w_{s(t)}}+ \rho' P_t (\n{w_t}+\n{w_{s(t)}})\right)^2,
\end{align*}
where the first inequality is due to $\E[\ns{X-\E[X]}]\leq \E[X^2]$.

Then, according to the vector version Bernstein inequality (Lemma~\ref{vectorBernstein}), with probability at least $1-\zeta$, we have 
$$\n{\xi_t-\xi_t'}\leq O\Big(\frac{\log(d/\zeta)}{\sqrt{b}}\Big)\left( L\n{w_t-w_{s(t)}} +\rho'P_t(\n{w_t}+\n{w_{s(t)}}) \right),$$
where $O(\cdot)$ hides constants.

In order to prove the other bound for the variance difference, we can use smoothness condition to bound each term as follows. 
\begin{align*}
&\n{\xi_{t,i}-\xi_{t,i}'} \\
\leq& \n{\nabla f_i(x_t)-\nabla f_i(x_t')} +\n{ \nabla f_i(\snap)- \nabla f_i(\snap')} \\
&+ 
\n{(\nabla f(x_t)-\nabla f(x_t') }+\n{ \nabla f(\snap)- \nabla f(\snap')}\\
\leq& 2L(\n{w_t}+\n{w_{s(t)}}).
\end{align*}
The summation of variance of each term can be bounded as 
$$\sigma^2\leq L^2(\n{w_t}+\n{w_{s(t)}})^2 .$$
Again, using Bernstein inequality, we know with probability at least $1-\zeta$
$$\n{\xi_t-\xi_t'}\leq O\Big(\frac{\log(d/\zeta)}{\sqrt{b}}\Big) L(\n{w_t}+\n{w_{s(t)}}).$$

By union bound, we know with probability at least $1-2\zeta$, 
$$\n{\xi_t-\xi_t'}\leq O\Big(\frac{\log(d/\zeta)}{\sqrt{b}}\Big)\min\Big( L\n{w_t-w_{s(t)}} +\rho'P_t(\n{w_t}+\n{w_{s(t)}}), L(\n{w_t}+\n{w_{s(t)}}) \Big).$$ 
\end{proofof}

Suppose the initial point $\tx$ of the super epoch has a large negative curvature ($\lambda_{\min}(\hessian)=-\gamma<0$). Also assume initially the two sequences has a reasonable distance along $e_1$ direction, which is the most negative eigendirection of $\hessian$. Then, using the above bound for the variance of $w_t$, we are able to prove that the distance between two sequences increases exponentially, and becomes large after $\tdo(\frac{1}{\eta\gamma})$ steps, which means at least one sequence must escape the initial point $\tx$.

\begin{lemma}\label{lem:distanceproperty_psvrg}
Let $\{x_t\}$ and $\{x_t'\}$ be two SVRG sequences running on $f$ that use the same choice of mini-batches. Assume $w_0=x_0-x_0'$ aligns with $e_1$ direction and $|\inner{e_1}{w_0}|\geq \frac{\delta}{4\sqrt{d}}.$
Let the threshold distance $\mathL:=\frac{\gamma}{C_3\max(\rho, \rho'/m)}.$
Assume for every $0\leq t\leq \frac{2\log(\frac{d\gamma}{\rho\delta})}{\eta\gamma}-1$, $\n{\xi_t-\xi_t'}\leq \frac{C_1'}{\sqrt{b}}\min\left( L\n{w_t-w_{s(t)}} +\rho'P_t(\n{w_t}+\n{w_{s(t)}}), L(\n{w_t}+\n{w_{s(t)}}) \right),$ where $C_1'$ comes from Lemma~\ref{lem:coupled_gradient_estimator}. Then there exists large enough constant $c$ such that as long as 
\begin{align*}
\eta\leq \frac{1}{c\log(\frac{d\gamma}{\rho\delta})C_1' \cdot L}, \quad
 C_3\geq \frac{1}{\eta L}. 
\end{align*}
we have
\begin{align*}
\max(\n{x_T-\tx},\n{x_T'-\tx})\geq \mathL,
\end{align*}
for some $T\leq \frac{2\log(\frac{d\gamma}{\rho\delta})}{\eta\gamma}.$
\end{lemma}

The proof of this lemma is similar to the analysis in \cite{jin2017escape}. However, we make crucial use of Lemma~\ref{lem:coupled_gradient_estimator}. Throughout the proof, the intuition is that at every iteration, $w_t$ is close to a multiple of $e_1$. Therefore, the next $w_{t+1}$ is close to $(I-\eta H)w_t = (1+\eta \gamma) w_t$. The difference between $w_{t+1}$ and $w_t$ is therefore only $\eta \gamma w_t$ whose norm is much smaller than either $w_t$ or $w_{t+1}$. As a result, Lemma~\ref{lem:coupled_gradient_estimator} gives a much tighter bound on the variance, and allows the proof to go through. 

\begin{proofof}{Lemma~\ref{lem:distanceproperty_psvrg}}
For the sake of contradiction, assume for any $t\leq \frac{2\log(\frac{d\gamma}{\rho\delta})}{\eta\gamma},\ \max(\n{x_t-\tx}, \n{x_t'-\tx})< \mathL.$ Basically, we will show that the distance between two sequences grows exponentially and will become larger than $2\mathL$ after $\frac{2\log(\frac{d\gamma}{\rho\delta})}{\eta\gamma}$ steps, which by triangle inequality implies that at least one sequence escapes after $\frac{2\log(\frac{d\gamma}{\rho\delta})}{\eta\gamma}$ steps. 

For any $0\leq t\leq \frac{2\log(\frac{d\gamma}{\rho\delta})}{\eta\gamma},$ we will inductively prove that 
\begin{enumerate}
\item $\frac{4}{5}(1+\eta\gamma)^t\n{w_0} \leq \n{w_t}\leq \frac{6}{5} (1+\eta\gamma)^t\n{w_0};$
\item $\n{\xi_t-\xi_t'}\leq \mu\cdot \eta\gamma C_1' L(1+\eta\gamma)^t\n{w_0},$ where $\mu=\tdo(1).$
\end{enumerate}

The base case trivially holds because $\frac{4}{5}\n{w_0} \leq \n{w_0}\leq \frac{6}{5}\n{w_0}$ and $\xi_0=\xi_0'=0.$ Fix any $t\leq \frac{2\log(\frac{d\gamma}{\rho\delta})}{\eta\gamma}$, assume for every $\tau \leq t-1$, the two induction hypotheses hold, we prove they still hold for $t$. 

\paragraph{Proving Hypothesis 1.} Let's first prove $\frac{4}{5} (1+\eta\gamma)^t\n{w_0}\leq \n{w_t}\leq \frac{6}{5} (1+\eta\gamma)^t\n{w_0}$. We can expand $w_t$ as follows, 
\begin{align*}
w_t&=w_{t-1}-\eta(v_{t-1}-v_{t-1}')\\
&=(I-\eta \hessian)w_{t-1}-\eta(\Delta_{t-1}w_{t-1}+\xi_{t-1}-\xi_{t-1}')\\
&=(I-\eta \hessian)^{t}w_0-\eta\sum_{\tau=0}^{t-1}(I-\eta \hessian)^{t-\tau-1}(\Delta_\tau w_\tau+\xi_\tau-\xi_\tau')
\end{align*}
where $\Delta_{\tau}=\int_0^1(\nabla^2 f(x_\tau'+\theta(x_\tau-x_\tau'))-\hessian)d\theta$. It's clear that the first term aligns with $e$ direction and has norm $(1+\eta\gamma)^t\n{w_0}$. Thus, we only need to show $\n{\eta\sum_{\tau=0}^{t-1}(I-\eta \hessian)^{t-\tau-1}(\Delta_\tau w_\tau+\xi_\tau-\xi_\tau')}\leq \frac{1}{5} (1+\eta\gamma)^t\n{w_0}.$

We first look at the Hessian changing term. According to the assumptions, we know $\n{x_\tau -\tx}, \n{x_\tau'-\tx}\leq \mathL$ for any $\tau\leq \frac{2\log(\frac{d\gamma}{\rho\delta})}{\eta\gamma}.$ Thus,
\begin{align*}
\left\|\eta\sum_{\tau=0}^{t-1}(I-\eta \hessian)^{t-\tau-1}\Delta_\tau w_\tau\right\|
&\leq \eta\sum_{\tau=0}^{t-1} (1+\eta\gamma)^{t-\tau-1}\n{\Delta_\tau}\n{w_\tau}\\
&\leq \eta\sum_{\tau=0}^{t-1} \rho\max(\n{x_\tau-\tx}, \n{x_\tau'-\tx}) \frac{6}{5}(1+\eta\gamma)^t \n{w_0}\\ 
&\leq \eta\sum_{\tau=0}^{t-1} \frac{6}{5}\rho\frac{\gamma}{C_3\max(\rho,\rho'/m)} (1+\eta\gamma)^t \n{w_0}\\ 
&\leq \frac{1}{\gamma}\cdot \frac{12}{5}\log(\frac{d\gamma}{\rho\delta})\frac{\gamma}{C_3} (1+\eta\gamma)^t \n{w_0}\\ 
&\leq \frac{1}{10}(1+\eta\gamma)^t \n{w_0},
\end{align*}
where the second last inequality uses the assumption that $t\leq \frac{2\log(\frac{d\gamma}{\rho\delta})}{\eta\gamma}$ and the last inequality holds as long as $C_3\geq 24\log(\frac{d\gamma}{\rho\delta}).$

For the variance term, we have 
\begin{align*}
\left\| \eta\sum_{\tau=0}^{t-1}(I-\eta \hessian)^{t-\tau-1}(\xi_\tau-\xi_\tau')\right\|
&\leq \eta\sum_{\tau=0}^{t-1}(1+\eta\gamma)^{t-\tau-1}\n{\xi_\tau-\xi_\tau'}\\
&\leq \eta\sum_{\tau=0}^{t-1}(1+\eta\gamma)^{t-\tau-1}\mu\eta\gamma C_1' L(1+\eta\gamma)^\tau \n{w_0}\\
&\leq \eta\frac{2\log(\frac{d\gamma}{\rho\delta})}{\eta\gamma}\mu\eta\gamma C_1' L(1+\eta\gamma)^t \n{w_0}\\
&\leq \frac{1}{10}(1+\eta\gamma)^t \n{w_0},
\end{align*}
where the last inequality holds as long as $\eta\leq \frac{1}{20\log(\frac{d\gamma}{\rho\delta})\mu C_1'\cdot L}.$


Overall, we have $\n{\eta\sum_{\tau=0}^{t-1}(I-\eta \hessian)^{t-\tau-1}(\Delta_\tau w_\tau+\xi_\tau-\xi_\tau')}\leq \frac{1}{5} (1+\eta\gamma)^t\n{w_0}$, which implies $\frac{4}{5} (1+\eta\gamma)^t\n{w_0}\leq \n{w_t}\leq \frac{6}{5} (1+\eta\gamma)^t\n{w_0}$.

\paragraph{Proving Hypothesis 2.}Next, we show the second hypothesis also holds, $\n{\xi_t-\xi_t'}\leq \mu\cdot \eta\gamma C_1' L(1+\eta\gamma)^t\n{w_0}$. We separately consider two cases when $\frac{1}{\eta\gamma} \leq m$ and $\frac{1}{\eta\gamma} > m$. 

If $\frac{1}{\eta\gamma} \leq m$, we have 
\begin{align*}
\n{\xi_t-\xi_t'} 
\leq & \frac{C_1'}{\sqrt{b}}\left( L(\n{w_t}+\n{w_{s(t)}}) \right)\\
\leq & \frac{C_1'}{\sqrt{b}} 2L\cdot \frac{6}{5}(1+\eta\gamma)^t\n{w_0}\\
\leq & \mu \frac{C_1' L}{\sqrt{b}}(1+\eta\gamma)^t\n{w_0}\\
\leq & \mu\cdot \eta\gamma C_1' L(1+\eta\gamma)^t\n{w_0},
\end{align*}
where the third inequality holds as long as $\mu\geq 3$ and the last inequality holds because $\frac{1}{\sqrt{b}}\leq \frac{1}{m}\leq \eta\gamma.$


If $\frac{1}{\eta\gamma} >m$, we need to bound $\n{w_t-w_{s(t)}}$ more carefully. We can write $w_t-w_{s(t)}$ as follows,
$$w_t -w_{s(t)} =\left((I-\eta \hessian)^{t-s(t)}-I\right)w_{s(t)}-\eta\sum_{\tau=s(t)}^{t-1}(I-\eta \hessian)^{t-\tau-1}(\Delta_\tau w_\tau+\xi_\tau-\xi_\tau').$$

For the first term, we have 
\begin{align*}
\left\|\left((I-\eta \hessian)^{t-s(t)}-I\right)w_{s(t)} \right\| 
\leq& \n{ (I-\eta \hessian)^{t-s(t)}-I}\n{w_{s(t)}}\\
\leq& \left((1+\eta\gamma)^m -1\right)\frac{6}{5}(1+\eta\gamma)^t\n{w_0}\\
\leq& 3m\eta\gamma\cdot(1+\eta\gamma)^t\n{w_0},
\end{align*}
where the last inequality holds since $(1+\eta\gamma)^m\leq 1+2m\eta\gamma$ if $m\eta\gamma<1.$


For the hessian changing term, we have 
\begin{align*}
\n{\eta\sum_{\tau=s(t)}^{t-1}(I-\eta \hessian)^{t-\tau-1}\Delta_\tau w_\tau }
&\leq \eta\sum_{\tau=s(t)}^{t-1} 2\frac{\gamma}{C_3} (1+\eta\gamma)^t \n{w_0}\\ 
&\leq \eta m \cdot 2\frac{\gamma}{C_3} (1+\eta\gamma)^t \n{w_0}\\ 
&\leq m\eta\gamma (1+\eta\gamma)^t\n{w_0},
\end{align*}
assuming $C_3\geq 2.$ 

For the variance term, we have 
\begin{align*}
\n{ \eta\sum_{\tau=s(t)}^{t-1}(I-\eta \hessian)^{t-\tau-1}(\xi_\tau-\xi_\tau')}
\leq& \eta\sum_{\tau=s(t)}^{t-1}(1+\eta \gamma)^{t-\tau-1}\n{\xi_\tau-\xi_\tau'}\\
\leq& \eta\sum_{\tau=s(t)}^{t-1}(1+\eta \gamma)^{t-\tau-1}\mu \eta\gamma C_1' L(1+\eta\gamma)^\tau \n{w_0}\\
\leq& \mu C_1' \eta L\cdot m\eta\gamma (1+\eta\gamma)^t\n{w_0}\\
\leq& m\eta\gamma (1+\eta\gamma)^t\n{w_0},
\end{align*}
where the second inequality uses induction hypothesis and the last inequality assumes $\eta\leq \frac{1}{C_1'\mu\cdot L}.$

Overall, we have $\n{w_t-w_{s(t)}}\leq 5m\eta\gamma (1+\eta\gamma)^t\n{w_0}$. Thus, when $\frac{1}{\eta\gamma}>m$, we can bound $\n{\xi_t-\xi_{s(t)}}$ as follows, 
\begin{align*}
\n{\xi_t-\xi_t'} 
\leq & \frac{C_1'}{\sqrt{b}}\left( L\n{w_t-w_{s(t)}} +\rho'P_t(\n{w_t}+\n{w_{s(t)}}) \right)\\
\leq & \frac{C_1'}{\sqrt{b}}\left( L\cdot 5m\eta\gamma  +\rho'\frac{12\gamma}{5C_3\max(\rho,\rho'/m)} \right)(1+\eta\gamma)^t\n{w_0}\\
\leq & \frac{C_1'}{\sqrt{b}}\left( L\cdot 5m\eta\gamma  +\frac{12}{5}L\cdot m\eta\gamma\right)(1+\eta\gamma)^t\n{w_0}\\
\leq & \mu\cdot \eta\gamma C_1' L(1+\eta\gamma)^t\n{w_0},
\end{align*}
where the second last inequality assumes $C_3\geq \frac{1}{\eta L}$ and the last inequality holds as long as $\mu\geq 8$. Here, we use the fact that $P_t\leq \max(\n{x_{s(t)}-\tx}, \n{x_{s(t)}'-\tx}, \n{x_{t}-\tx}, \n{x_{t}'-\tx})\leq \mathL.$

Overall, we know there exists large enough constant $c$ such that the induction holds as long as 
\begin{align*}
&\eta\leq \frac{1}{c\log(\frac{d\gamma}{\rho\delta})C_1' \cdot L}\\
& C_3\geq \frac{1}{\eta L}. 
\end{align*}

Thus, we know $\n{w_t}\geq \frac{4}{5}(1+\eta\gamma)^t\n{w_0}$ for any $t\leq \frac{2\log(\frac{d\gamma}{\rho\delta})}{\eta\gamma}$. Specifically, when $t=\frac{2\log(\frac{d\gamma}{\rho\delta})}{\eta\gamma}$, we have 
\begin{align*}
\n{w_t}&\geq \frac{4}{5}(1+\eta\gamma)^t\n{w_0}\\
&\geq \frac{4}{5}(1+\eta\gamma)^{\frac{2\log(\frac{d\gamma}{\rho\delta})}{\eta\gamma}}\frac{\delta}{4\sqrt{d}}\\
&\geq \frac{\gamma}{5\rho},
\end{align*}
which implies $\max(\n{x_t-\tx}, \n{x_t'-\tx}) \geq \frac{\gamma}{10\rho}.$ Assuming $C_3\geq 10,$ this contradicts the assumption that $\max(\n{x_t-\tx}, \n{x_t'-\tx})< \frac{\gamma}{C_3\max(\rho,\rho'/m)}=:\mathL,$ for any $t\leq \frac{2\log(\frac{d\gamma}{\rho\delta})}{\eta\gamma}$. Thus, we know there exists $T\leq \frac{2\log(\frac{d\gamma}{\rho\delta})}{\eta\gamma}$ such that,
\begin{align*}
\max(\n{x_T-\tx},\n{x_T'-\tx})\geq \mathL.
\end{align*}
\end{proofof}

In the next lemma, we show that the function value decrease can be lower bounded by the distance to the snapshot point. Combined with the above lemma, this shows that the function value decreases significantly in the super epoch. 
The proof of this lemma is almost the same as the proof of Lemma~\ref{lem:gradientfunctionvalue}. 
\begingroup
\def\thetheorem{\ref{lem:dis_value_cross_epoch}}
\begin{lemma}
Let $x_0$ be the initial point, which is also the snapshot point of the current epoch. Let $\{x_t\}$ be the iterates of SVRG running on $f$ starting from $x_0$. Fix any $t\geq 1$, suppose for every $0\leq \tau \leq t-1, \n{\xi_\tau}\leq \frac{C_1 L}{\sqrt{b}}\n{x_\tau-x_{s(\tau)}},$ where $C_1$ comes from Lemma~\ref{lem:varaincedistance}. Given $\eta\leq \frac{1}{3C_1 L},b\geq m^2,$ we have 
$$\ns{x_t-x_0}\leq \frac{4t}{C_1 L}(f(x_0)-f(x_t)). $$
\end{lemma}
\addtocounter{theorem}{-1}
\endgroup

\begin{proofof}{Lemma~\ref{lem:dis_value_cross_epoch}}
From Equation (\ref{eq:key}) in the proof of Lemma~\ref{lem:gradientfunctionvalue}, we know for any $t'\leq t$, 
$$\ns{x_{t'}-x_{s(t')}}\leq \frac{2(t'-s(t'))}{C_1 L}(f(x_{s(t')})-f(x_{t'})),$$
where $x_{s(t')}$ is the snapshot point of $x_{t'}.$

If $t\leq m$, we know there is only one epoch from $x_0$ to $x_t$ and 
$$\ns{x_t-x_0}\leq \frac{2t}{C_1 L}(f(x_0)-f(x_t)).$$
If $t>m$, we need to divide $x_t-x_0$ into multiple epochs and bound them separately. 
We have 
\begin{align*}
\ns{x_t-x_0} &= \ns{x_m-x_0 + x_{2m}-x_m +\cdots x_t-x_{s(t)}}\\ 
&\leq \lceil \frac{t}{m}\rceil \left(\sum_{\tau=1}^{\lfloor t/m\rfloor}\ns{x_{\tau m}-x_{(\tau-1)m}} +\ns{x_t-x_{s(t)}}\right)\\
&\leq \frac{2t}{m}\cdot\frac{2m}{C_1 L}(f(x_0)-f(x_t))\\
&\leq \frac{4t}{C_1 L}(f(x_0)-f(x_t))
\end{align*}

Combining two cases, we have 
\begin{align*}
\ns{x_t-x_0}\leq \frac{4t}{C_1 L}(f(x_0)-f(x_t)).
\end{align*}
\end{proofof}

Next, we show that starting from a randomly perturbed point, with constant probability the function value decreases a lot within a super epoch. 
\begin{lemma}\label{lem:lemma2_psvrg}
Let $\tx$ be the initial point with gradient $\n{\nabla f(\tx)}\leq \mathG$ and $\lambda_{\min}(\hessian)=-\gamma<0$. 
Let $\{x_t\}$ be the iterates of SVRG running on $f$ starting from $x_0$, which is a uniformly perturbed point from $\tx$. There exist $\eta=\tdo(1/L), b=\tdo(n^{2/3}), \delta=\tdo(\min(\frac{\rho\gamma}{\max(\rho^2,(\rho'/m)^2)}, \frac{\gamma^{1.5}}{\max(\rho,\rho'/m)\sqrt{L}})), \mathG=\tdo(\frac{\gamma^2}{\rho}),\mathL=\tdo(\frac{\gamma}{\max(\rho,\rho'/m)}), T_{\max}=\tdo(\frac{1}{\eta\gamma})$ such that with probability at least $1/8,$
$$f(x_T)-f(\tx)\leq -C_5\cdot \frac{\gamma^3}{\max(\rho^2,(\rho'/m)^2)};$$
and with high probability, 
$$f(x_T)-f(\tx)\leq \frac{C_5}{20}\cdot \frac{\gamma^3}{\max(\rho^2,(\rho'/m)^2)};$$
where $C_5=\tdtheta(1)$ and $T$ is the length of the current super epoch and $T\leq T_{\max}.$
\end{lemma}

This lemma is basically a combination of Lemma~\ref{lem:distanceproperty_psvrg} and Lemma~\ref{lem:lemma2_psvrg}. Lemma~\ref{lem:distanceproperty_psvrg} shows that with reasonable probability, one of two random starting points is going to travel a large distance, while Lemma~\ref{lem:lemma2_psvrg} shows such a point would decrease the function value. The only additional thing is to prove is that the function value does not increase by too much when the point does not escape. Intuitively this is true because with high probability the function value can only increase during the initial perturbation.

\begin{proofof}{Lemma~\ref{lem:lemma2_psvrg}}
With the help of Lemma~\ref{lem:distanceproperty_psvrg}, we first prove that $\{x_t\}$ escapes the saddle point with a constant probability.
Let $\{x_t\}$ and $\{x_t'\}$ be two SVRG sequences starting from $x_0$ and $x_0'$ respectively, where $x_0$ and $x_0'$ are two perturbed points satisfying $\n{x_0-\tx},\n{x_0'-\tx}\leq \delta$. According to Lemma~\ref{lem:distanceproperty_psvrg}, we know at least one sequence escapes the saddle point if $x_0-x_0'$ aligns with $e_1$ direction and has norm as least $\frac{\delta}{4\sqrt{d}}.$ 

We first show that, for two coupled random points $x_0$ and $x_0'$, their distance is at least $\frac{\delta}{4\sqrt{d}}$ with a reasonable probability. Marginally, $x_0$ and $x_0'$ are both uniformly sampled from the ball centered at $\tx$ with radius $\delta$. They are coupled in the sense that they have the same projections onto the orthogonal subspace of $e_1$. Then, similar as the analysis in~\cite{jin2017escape},
$$\Pr\left[\n{x_0-x_0'}< \frac{\delta}{4\sqrt{d}}\right] \leq \frac{1}{2}\frac{\frac{\delta}{\sqrt{d}}\times \mbox{Vol}(\mathbb{B}_0^{(d-1)}(\delta))}{\mbox{Vol}(\mathbb{B}_0^{(d)}(\delta))}= \frac{1}{2}\frac{1}{\sqrt{\pi d}}\frac{\Gamma(d/2+1)}{\Gamma(d/2+1/2)}\leq \frac{1}{2}.$$

Thus, we know with at least half probability, we have $|\inner{x_0-x_0'}{e_1}|\geq \frac{\delta}{4\sqrt{d}}$. In order to apply Lemma~\ref{lem:distanceproperty_psvrg}, we still need to make sure $\n{\xi_t-\xi_t'}$ is well bounded for every $0\leq t\leq \frac{2\log(\frac{d\gamma}{\rho\delta})}{\eta\gamma}-1$, which happens with high probability due to Lemma~\ref{lem:coupled_gradient_estimator}. 
Thus, by the union bound and Lemma~\ref{lem:distanceproperty_psvrg}, we know with probability no less than $1/3$, at least one sequence between $\{x_t\}$ and $\{x_t'\}$ must escape the saddle point. Marginally, we know from a randomly perturbed point $x_0$,  sequence $\{x_t\}$ escapes the saddle point within a super epoch with probability at least $1/6.$ Precisely, there exists $\eta=\frac{1}{C_6\cdot L}, \mathL=\frac{\gamma}{C_3\max(\rho,\rho'/m)}, T\leq \frac{C_7}{\eta\gamma}$ such that 
$$\n{x_T-\tx}\geq \mathL$$
holds with probability at least $1/6$. Here, we have $C_3, C_6, C_7=\tdo(1).$ 

Combing Lemma~\ref{lem:varaincedistance} and Lemma~\ref{lem:dis_value_cross_epoch}, we also know with high probability
$$\ns{x_T-x_0}\leq \frac{T}{C_4 L}(f(x_0)-f(x_T))$$
where $C_4=\tdo(1)$.

By a union bound, we know with probability at least $1/8$,
we have 
\begin{align*}
f(x_0)-f(x_T) \geq& \frac{C_4 L}{T}\ns{x_T-x_0}\\
\geq& \frac{C_4 L}{T}\left(\n{x_T-\tx}-\n{x_0-\tx}\right)^2\\
\geq& \frac{C_4 L}{T}\left(\frac{\gamma}{C_3 \max(\rho,\rho'/m)}-\delta\right)^2\\
\geq& \frac{C_4 L\eta\gamma}{C_7}\frac{\gamma^2}{4C_3^2 \max(\rho^2,(\rho'/m)^2)}\\
=& \frac{C_4}{4C_7 C_3^2 C_6}\frac{\gamma^3}{\max(\rho^2,(\rho'/m)^2)},
\end{align*}
where the last inequality holds as long as $\delta \leq \frac{\gamma}{2C_3\max(\rho,\rho'/m)}.$

Let the threshold gradient $\mathG:=\frac{\gamma^2}{C_8\rho}$. Since $f$ is $L$-smooth, we have 
\begin{align*}
f(x_0)-f(\tx)\leq& \n{\nabla f(\tx)}\cdot \n{x_0 -\tx}+ \frac{L}{2}\ns{\tx-x_0}\\
\leq& \frac{\gamma^2}{C_8\rho}\delta+\frac{L}{2}\delta^2.
\end{align*}

Thus, with probability at least $1/8$, we know
\begin{align*}
f(x_T)-f(\tx) 
=& f(x_T) - f(x_0) + f(x_0) - f(\tx)\\
\leq& -\frac{C_4}{4C_7 C_3^2 C_6}\frac{\gamma^3}{\max(\rho^2,(\rho'/m)^2)}+ \frac{\gamma^2}{C_8\rho}\delta+\frac{L}{2}\delta^2.
\end{align*}

If Lemma~\ref{lem:distanceproperty_psvrg} fails, the function value is not guaranteed to decrease. On the other hand, we know that with high probability the function value does not increase, $f(x_T) - f(x_0)\leq 0.$ Thus, with high probability, we know
\begin{align*}
f(x_T)-f(\tx) 
\leq \frac{\gamma^2}{C_8\rho}\delta+\frac{L}{2}\delta^2.
\end{align*}

Assuming $\delta\leq \min(\frac{C_4C_8}{168C_7 C_3^2 C_6}\frac{\rho\gamma}{\max(\rho^2,(\rho'/m)^2)}, \sqrt{\frac{C_4}{84C_7 C_3^2 C_6}}\frac{\gamma^{1.5}}{\max(\rho,\rho'/m) \sqrt{L}}),$ we know with probability at least $1/8$,
$$f(x_T)-f(\tx)\leq  -\frac{20}{21}\cdot \frac{C_4}{4C_7 C_3^2 C_6}\frac{\gamma^3}{\max(\rho^2,(\rho'/m)^2)};$$
and with high probability, 
$$f(x_T)-f(\tx)\leq  \frac{1}{21}\cdot \frac{C_4}{4C_7 C_3^2 C_6}\frac{\gamma^3}{\max(\rho^2,(\rho'/m)^2)}.$$

We finish the proof by choosing $C_5:=\frac{20}{21}\frac{C_4}{4C_7 C_3^2 C_6}$.
\end{proofof}
\section{Proofs of Exploiting Negative Curvature - Stabilized SVRG}\label{sec:proofsecondorder_stabilized}

In this section, we analyze the behavior of Stabilized SVRG when the initial gradient is small. The proofs will depend on Lemma~\ref{lem:varaincedistance}, Lemma~\ref{lem:coupled_gradient_estimator} and Lemma~\ref{lem:dis_value_cross_epoch}, which were proved for $f$ but clearly also holds for shifted function $\hat{f}$. 

Let the initial point of the super epoch be $\tx$, whose hessian is denoted by $\hessian$. Assume the initial point has large negative curvature, $\lambda_{\min}(\hessian)=-\gamma<0.$ Let $x_0$ be the perturbed point and let $\{x_t\}$ be the SVRG iterates running on $\hat{f}$ starting from $\tx.$ As we discussed in Section~\ref{sec:negativecurvature}, there are two phases in the analysis. In the first phase, the distance between the current iterate $x_t$ and the starting point $\tx$ remains small (comparable to the random perturbation), while at the end the direction of $x_t-\tx$ aligns with the negative eigendirections. In the second phase, the distance to the initial point $\tx$ blows up exponentially and the algorithm escapes from saddle points.

To analyze the two phases of the algorithm, we make use of the following expansion for the one-step movement of the algorithm:

\begin{lemma}\label{lem:expansion}
Let $\tx$ be the initial point with Hessian $\hessian$, and $x_0$ be its perturbed point. 
Let $\{x_t\}$ be the iterates of SVRG running on $\hat{f}$ starting from $x_0$. For any $t\geq 1$, we have the following expansion,
\begin{align*}
x_t-x_{t-1}=& -\eta(I-\eta\hessian)^{t-1} \nabla \hat{f}(x_0) +\eta^2\hessian \sum_{\tau=0}^{t-2}(I-\eta\hessian)^{t-2-\tau}\xi_\tau \\
&-\eta \sum_{\tau=0}^{t-2}(I-\eta\hessian)^{t-2-\tau}\Delta_\tau (x_{\tau+1}-x_\tau) -\eta\xi_{t-1},
\end{align*}
where variance term $\xi_{\tau}= v_{\tau}-\nabla \hat{f}(x_{\tau})$ and hessian changing term $\Delta_\tau = \int_0^1(\nabla^2 \hat{f}(x_\tau+\theta (x_{\tau+1}-x_\tau))-\hessian)d\theta$. 
\end{lemma}

Intuitively, the first term $-\eta(I-\eta\hessian)^{t-1} \nabla \hat{f}(x_0)$ corresponds to what happens to the algorithm if the function is quadratic (with Hessian equal to $\hessian$ at $\tx$). The second and the fourth term measures the difference introduced by the error in the gradient updates. The third term measures the difference introduced by the fact that the Hessian is not a constant. Our analysis will bound the last three terms to show that the behavior of the algorithm is very similar to what happens if we only have the first term.

\begin{proofof}{Lemma~\ref{lem:expansion}}
According to the algorithm, we know 
\begin{align*}
x_t-x_{t-1}&=-\eta v_{t-1}\\
&= -\eta(\nabla \hat{f}(x_{t-1}) + \xi_{t-1}),
\end{align*}
where $\xi_{t-1}= v_{t-1}-\nabla \hat{f}(x_{t-1}).$ We can further expand $\nabla \hat{f}(x_t)$ as follows. 
\begin{align*}
\nabla \hat{f}(x_t)&= \nabla \hat{f}(x_{t-1}) +\int_0^1\Big(\nabla^2 \hat{f}\big(x_{t-1}+\theta (x_t-x_{t-1})\big)\Big)d\theta(x_t-x_{t-1})\\
&= \nabla \hat{f}(x_{t-1})+\hessian(x_t-x_{t-1}) +\Delta_{t-1}(x_t-x_{t-1})\\
&=\nabla \hat{f}(x_{t-1}) -\eta\hessian(\nabla \hat{f}(x_{t-1})+\xi_{t-1})+\Delta_{t-1}(x_t-x_{t-1})\\
&=(I-\eta\hessian) \nabla \hat{f}(x_{t-1}) -\eta\hessian \xi_{t-1} +\Delta_{t-1}(x_t-x_{t-1})\\
&=(I-\eta\hessian)^t \nabla \hat{f}(x_0) -\eta\hessian \sum_{\tau=0}^{t-1}(I-\eta\hessian)^{t-1-\tau}\xi_\tau + \sum_{\tau=0}^{t-1}(I-\eta\hessian)^{t-1-\tau}\Delta_\tau (x_{\tau+1}-x_\tau),
\end{align*}
where $\Delta_\tau = \int_0^1(\nabla^2 \hat{f}(x_\tau+\theta (x_{\tau+1}-x_\tau))-\hessian)d\theta$. 
Thus, we know 
\begin{align*}
x_t-x_{t-1}=&-\eta(\nabla \hat{f}(x_{t-1}) + \xi_{t-1})\notag\\
=& -\eta(I-\eta\hessian)^{t-1} \nabla \hat{f}(x_0) +\eta^2\hessian \sum_{\tau=0}^{t-2}(I-\eta\hessian)^{t-2-\tau}\xi_\tau \\
&-\eta \sum_{\tau=0}^{t-2}(I-\eta\hessian)^{t-2-\tau}\Delta_\tau (x_{\tau+1}-x_\tau) -\eta\xi_{t-1}
\end{align*}
\end{proofof}

\subsection{Proofs of Phase 1}

In Phase 1, the goal of the algorithm is to stay close to the original point $\tx$, while making $x_t-\tx$ aligned with the negative eigendirections of $\hessian$ (Hessian at $\tx$).

Recall the definition of the length of Phase 1 as follows,
$$
T_1=\sup\left\{t|\forall t'\leq t-1, \left(t'\leq \frac{1}{\eta\gamma} \right) \vee\left(\n{\proj{S}(x_{t'}-\tx)}\leq \frac{\delta}{10} \right) \right\}.
$$
We will first show that $x_t-x_{t-1}$ is bounded by $\tdo(1/t)\delta$ for every $1\leq t\leq \min(T_1, \frac{\log(d)}{\eta\gamma})$. This lemma is very technical, and the main idea is to use the expansion in Lemma~\ref{lem:expansion} and bound the terms by considering their projections in different subspaces. Intuitively, the behavior can be separated into several cases based on the eigenvalues of $\hessian$ in the corresponding subspace:

\begin{enumerate}
\item eigenvalue smaller than $-\gamma/\log d$. These directions will grow exponentially, and we will stop the first phase when the projection in this subspace is large.
\item eigenvalue between $-\gamma/\log d$ and $0$. These directions will also grow, but they do not grow by more than a constant factor.
\item small positive eigenvalue (smaller than $+\gamma$). These directions don't move much throughout the iterates.
\item large positive eigenvalue (much larger than $\gamma$). These directions move very fast at the beginning, but converges very quickly and will not move much later on.
\end{enumerate}

In the proof we will consider the behavior of these separate subspaces (where cases 3 and 4 will be combined). The detailed proof is deferred to Section~\ref{sec:lemma19}. 

\begin{lemma}\label{lem:phase1}
Let $T_1$ be the length of Phase 1. Assume for any $0\leq t\leq \min(T_1,\frac{\log(d)}{\eta\gamma})-1$, $\n{\xi_t}\leq \frac{C_1L}{\sqrt{b}}\n{x_t-x_{s(t)}},$ where $C_1$ comes from Lemma~\ref{lem:varaincedistance}. Then, there exists large enough constant $c$ such that as long as
\begin{align*}
\eta\leq \frac{1}{cC_1\log(nd)\log(n\frac{\log(d)}{\eta\gamma})\cdot L},
\quad \mu\geq c\log(d)\log^2(\frac{\log(d)}{\eta\gamma}),
\quad \delta\leq \frac{\gamma}{\rho \mu^2},
\end{align*}
we have for every $1\leq t\leq \min(T_1,\frac{\log(d)}{\eta\gamma})$,
$$\n{x_t-x_{t-1}}\leq \frac{\mu}{t}\delta.$$
\end{lemma}


Now we want to prove that Phase 1 is successful with a reasonable probability. That is, at the end of Phase 1, with reasonable probability the distance $x_{T_1} - \tx$ is order $\tdo(\delta)$, while $\proj{S}(x_{T_1} - \tx)$ is at least $\delta/10$, where $\delta$ is the perturbation radius. By the above lemma, actually we only need to show that the length of Phase 1 is bounded by $\frac{\log(d)}{\eta\gamma}$. In the following proof, we show that between a pair of coupled sequences, at least one of them must end the Phase 1 within $\frac{\log(d)}{\eta\gamma}$ steps. 
Similar as in Lemma~\ref{lem:distanceproperty_psvrg}, we use two point analysis to show the difference between two sequences along $e_1$ direction increases exponentially and will become very large after $\frac{\log(d)}{\eta\gamma}$ steps, which implies that at least one sequence must have a large projection on $S$ subspace.
\begin{lemma}\label{lem:distanceproperty}
Let $\{x_t\}$ and $\{x_t'\}$ be two SVRG sequences running on $\hat{f}$ that use the same choice of mini-batches. Assume $w_0=x_0-x_0'$ aligns with $e_1$ direction and $|\inner{e_1}{w_0}|\geq \frac{\delta}{4\sqrt{d}}.$ Let $T_1, T_1'$ be the length of Phase 1 for $\{x_t\}$ and $\{x_t'\}$ respectively. Assume for every $1\leq t\leq \min(T_1,\frac{\log(d)}{\eta\gamma})$, $\n{x_t-x_{t-1}}\leq \frac{C_2}{t}\delta$ and for every $1\leq t\leq \min(T_1',\frac{\log(d)}{\eta\gamma})$, $\n{x_t'-x_{t-1}'}\leq \frac{C_2}{t}\delta$, where $C_2$ comes from Lemma~\ref{lem:phase1}. Assume for every $0\leq t\leq \frac{\log(d)}{\eta\gamma}-1$, $\n{\xi_t-\xi_t'}\leq \frac{C_1'}{\sqrt{b}}\min\left( L\n{w_t-w_{s(t)}} +\rho'P_t(\n{w_t}+\n{w_{s(t)}}), \\ L(\n{w_t}+\n{w_{s(t)}}) \right),$ where $C_1'$ comes from Lemma~\ref{lem:coupled_gradient_estimator}.  Then there exists large enough constant $c$ such that as long as 
\begin{align*}
 \delta \leq \min\left( \frac{\gamma}{c\log(d)\log(\frac{\log(d)}{\eta\gamma})C_2\rho }, \frac{m\eta L\gamma}{\rho'}\right),\quad
\eta\leq \frac{1}{c\log(d)\log(\frac{\log(d)}{\eta\gamma})C_1' C_2\cdot L},
\end{align*}
we have $\min(T_1, T_1')\leq \frac{\log(d)}{\eta\gamma}.$ W.l.o.g., suppose $T_1\leq \frac{\log(d)}{\eta\gamma}$ and we further have
\begin{align*}
&\forall 0\leq t\leq T_1,\ \n{x_{t}-\tx}\leq 3\log(\frac{\log(d)}{\eta\gamma})C_2 \delta,\\
&\n{\proj{S}(x_{T_1}-\tx)}\geq \frac{1}{10}\delta.
\end{align*}
\end{lemma}

\begin{proofof}{Lemma~\ref{lem:distanceproperty}}
For the sake of contradiction, assume the length of Phase 1 for both sequences are larger than $\frac{\log(d)}{\eta\gamma}.$ Basically, we will show that the distance between two sequences along $e_1$ direction grows exponentially and will become very large after $\frac{\log(d)}{\eta\gamma}$ steps, which implies that at least one sequence has a large projection along $e_1$ direction after $\frac{\log(d)}{\eta\gamma}$ steps. 

For any $0\leq t\leq \frac{\log(d)}{\eta\gamma},$ we will inductively prove that 
\begin{enumerate}
\item $\n{\proj{e_1} w_t}\geq \frac{4}{5} (1+\eta\gamma)^t\n{w_0}\ \mbox{and}\ \n{w_t}\leq \frac{6}{5} (1+\eta\gamma)^t\n{w_0};$
\item $\n{\xi_t-\xi_t'}\leq \mu\cdot \eta\gamma C_1' L(1+\eta\gamma)^t\n{w_0},$ where $\mu=\tdo(1).$
\end{enumerate}

The base case trivially holds. Fix any $t\leq \frac{\log(d)}{\eta\gamma}$, assume for every $\tau \leq t-1$, the two induction hypotheses hold, we prove they still hold for $t$. 

\paragraph{Proving Hypothesis 1.} Let's first prove $\n{\proj{e_1} w_t}\geq \frac{4}{5} (1+\eta\gamma)^t\n{w_0}\ \mbox{and}\ \n{w_t}\leq \frac{6}{5} (1+\eta\gamma)^t\n{w_0}$. We can expand $w_t$ as follows, 
\begin{align*}
w_t&=w_{t-1}-\eta(v_{t-1}-v_{t-1}')\\
&=(I-\eta \hessian)w_{t-1}-\eta(\Delta_{t-1}w_{t-1}+\xi_{t-1}-\xi_{t-1}')\\
&=(I-\eta \hessian)^{t}w_0-\eta\sum_{\tau=0}^{t-1}(I-\eta \hessian)^{t-\tau-1}(\Delta_\tau w_\tau+\xi_\tau-\xi_\tau')
\end{align*}
where $\Delta_{\tau}=\int_0^1(\nabla^2 \hat{f}(x_\tau'+\theta(x_\tau-x_\tau'))-\hessian)d\theta$. It's clear that the first term aligns with $e$ direction and has norm $(1+\eta\gamma)^t\n{w_0}$. Thus, we only need to show $\n{\eta\sum_{\tau=0}^{t-1}(I-\eta \hessian)^{t-\tau-1}(\Delta_\tau w_\tau+\xi_\tau-\xi_\tau')}\leq \frac{1}{5} (1+\eta\gamma)^t\n{w_0}.$

We first look at the Hessian changing term. According to the assumptions, we know $\n{x_\tau -\tx}, \n{x_\tau'-\tx}\leq 3\log(\frac{\log(d)}{\eta\gamma})C_2 \delta$ for any $\tau\leq \frac{\log(d)}{\eta\gamma}.$ Thus,
\begin{align*}
\left\|\eta\sum_{\tau=0}^{t-1}(I-\eta \hessian)^{t-\tau-1}\Delta_\tau w_\tau\right\|
&\leq \eta\sum_{\tau=0}^{t-1} (1+\eta\gamma)^{t-\tau-1}\n{\Delta_\tau}\n{w_\tau}\\
&\leq \eta\sum_{\tau=0}^{t-1} \rho\max(\n{x_\tau-\tx}, \n{x_\tau'-\tx}) \frac{6}{5}(1+\eta\gamma)^t \n{w_0}\\ 
&\leq \eta\sum_{\tau=0}^{t-1} \frac{18}{5}\log(\frac{\log(d)}{\eta\gamma})C_2 \rho \delta (1+\eta\gamma)^t \n{w_0}\\ 
&\leq \frac{1}{\gamma}\cdot 4\log(d)\log(\frac{\log(d)}{\eta\gamma})C_2 \rho \delta (1+\eta\gamma)^t \n{w_0}\\ 
&\leq \frac{1}{10}(1+\eta\gamma)^t \n{w_0},
\end{align*}
where the last inequality holds as long as $\delta\leq \frac{\gamma}{40\log(d)\log(\frac{\log(d)}{\eta\gamma})C_2\rho}.$

By the analysis in Lemma~\ref{lem:distanceproperty_psvrg}, we can bound the variance term as follows,
\begin{align*}
\left\| \eta\sum_{\tau=0}^{t-1}(I-\eta \hessian)^{t-\tau-1}(\xi_\tau-\xi_\tau')\right\| \leq \frac{1}{10}(1+\eta\gamma)^t \n{w_0},
\end{align*}
assuming $\eta\leq \frac{1}{10\log(d)\mu C_1'\cdot L}.$


Overall, we have $\n{\eta\sum_{\tau=0}^{t-1}(I-\eta \hessian)^{t-\tau-1}(\Delta_\tau w_\tau+\xi_\tau-\xi_\tau')}\leq \frac{1}{5} (1+\eta\gamma)^t\n{w_0}$, which implies $\n{\proj{e} w_t}\geq \frac{4}{5} (1+\eta\gamma)^t\n{w_0}\ \mbox{and}\ \n{w_t}\leq \frac{6}{5} (1+\eta\gamma)^t\n{w_0}$.

\paragraph{Proving Hypothesis 2.}Next, we show the second hypothesis also holds, $\n{\xi_t-\xi_t'}\leq \mu\cdot \eta\gamma C_1' L(1+\eta\gamma)^t\n{w_0}$. We separately consider two cases when $\frac{1}{\eta\gamma} \leq m$ and $\frac{1}{\eta\gamma} > m$. If $\frac{1}{\eta\gamma} \leq m$, the analysis is same as in Lemma~\ref{lem:distanceproperty_psvrg}. We have 
$
\n{\xi_t-\xi_t'} 
\leq  \mu\cdot \eta\gamma C_1' L(1+\eta\gamma)^t\n{w_0},
$
as long as $\mu\geq 3$.


If $\frac{1}{\eta\gamma} >m$, we need to bound $\n{w_t-w_{s(t)}}$ more carefully. We can write $w_t-w_{s(t)}$ as follows,
$$w_t -w_{s(t)} =\left((I-\eta \hessian)^{t-s(t)}-I\right)w_{s(t)}-\eta\sum_{\tau=s(t)}^{t-1}(I-\eta \hessian)^{t-\tau-1}(\Delta_\tau w_\tau+\xi_\tau-\xi_\tau').$$

The analysis for the first term and the variance term is again same as in Lemma~\ref{lem:distanceproperty_psvrg}. We have 
\begin{align*}
\left\|\left((I-\eta \hessian)^{t-s(t)}-I\right)w_{s(t)} \right\| +\left\| \eta\sum_{\tau=s(t)}^{t-1}(I-\eta \hessian)^{t-\tau-1}(\xi_\tau-\xi_\tau')\right\|
\leq 4m\eta\gamma\cdot(1+\eta\gamma)^t\n{w_0},
\end{align*}
assuming $\eta\leq \frac{1}{C_1'\mu\cdot L}$.


For the Hessian changing term, we have 
\begin{align*}
\left\| \eta\sum_{\tau=s(t)}^{t-1}(I-\eta \hessian)^{t-\tau-1}\Delta_\tau w_\tau \right\|
&\leq \eta\sum_{\tau=s(t)}^{t-1} 3\log(\frac{\log(d)}{\eta\gamma})C_2 \rho \delta\frac{6}{5} (1+\eta\gamma)^t \n{w_0}\\ 
&\leq \eta m \cdot 4\log(\frac{\log(d)}{\eta\gamma})C_2 \rho \delta (1+\eta\gamma)^t \n{w_0}\\ 
&\leq m\eta\gamma (1+\eta\gamma)^t\n{w_0},
\end{align*}
assuming $\delta\leq \frac{\gamma}{4\log(\frac{\log(d)}{\eta\gamma})C_2\rho}.$ 


Overall, we have $\n{w_t-w_{s(t)}}\leq 5m\eta\gamma (1+\eta\gamma)^t\n{w_0}$. Thus, when $\frac{1}{\eta\gamma}>m$, we can bound $\n{\xi_t-\xi_t'}$ as follows, 
\begin{align*}
\n{\xi_t-\xi_t'} 
\leq & \frac{C_1'}{\sqrt{b}}\left( L\n{w_t-w_{s(t)}} +\rho'P_t(\n{w_t}+\n{w_{s(t)}}) \right)\\
\leq & \frac{C_1'}{\sqrt{b}}\left( L\cdot 5m\eta\gamma  +8\log(\frac{\log(d)}{\eta\gamma})C_2 \rho'\delta \right)(1+\eta\gamma)^t\n{w_0}\\
\leq & \frac{C_1'}{\sqrt{b}}\left( L\cdot 5m\eta\gamma  +L\cdot 8\log(\frac{\log(d)}{\eta\gamma})C_2 m\eta\gamma\right)(1+\eta\gamma)^t\n{w_0}\\
\leq & \mu\cdot \eta\gamma C_1' L(1+\eta\gamma)^t\n{w_0},
\end{align*}
where the second last inequality assumes $\delta \leq \frac{m\eta L\gamma}{\rho'}$ and the last inequality holds as long as $\mu\geq 5+8\log(\frac{\log(d)}{\eta\gamma})C_2 $. Here, we also use the fact that $P_t\leq \max(\n{x_{s(t)}-\tx}, \n{x_{s(t)}'-\tx}, \n{x_{t}-\tx}, \n{x_{t}'-\tx})\leq 3\log(\frac{\log(d)}{\eta\gamma})C_2 \delta.$

Overall, we know there exists large enough constant $c$ such that the induction holds given 
\begin{align*}
& \delta \leq \min\left( \frac{\gamma}{c\log(d)\log(\frac{\log(d)}{\eta\gamma})C_2\rho }, \frac{m\eta L\gamma}{\rho'}\right)\\
&\eta\leq \frac{1}{c\log(d)\log(\frac{\log(d)}{\eta\gamma})C_1' C_2\cdot L}. 
\end{align*}

Thus, we know $\n{\proj{e_1}w_t}\geq \frac{4}{5}(1+\eta\gamma)^t\n{w_0}$ for any $t\leq \frac{\log(d)}{\eta\gamma}$. Specifically, when $t=\frac{\log(d)}{\eta\gamma}$, we have 
\begin{align*}
\n{\proj{e_1}w_t}&\geq \frac{4}{5}(1+\eta\gamma)^t\n{w_0}\\
&\geq \frac{4}{5}(1+\eta\gamma)^{\frac{\log(d)}{\eta\gamma}}\frac{\delta}{4\sqrt{d}}\\
&> \frac{\delta}{5},
\end{align*}
which implies $\max(\n{\proj{e_1} x_t-\tx}, \n{\proj{e_1} x_t'-\tx}) > \frac{\delta}{10}.$ This contradicts the assumption that neither sequence stops within $\frac{\log(d)}{\eta\gamma}$ steps. Thus, we know $\min(T_1, T_1')\leq \frac{\log(d)}{\eta\gamma}.$ Without loss of generality, suppose $T_1\leq \frac{\log(d)}{\eta\gamma}$, we have 
\begin{align*}
&\forall 0\leq t\leq T_1,\ \n{x_t-\tx}\leq 3\log(\frac{\log(d)}{\eta\gamma})C_2 \delta,\\
&\n{\proj{S}(x_{T_1}-\tx)}\geq \frac{1}{10}\delta.
\end{align*}
\end{proofof}


\subsection{Proof of Lemma~\ref{lem:phase1}}\label{sec:lemma19}
In this section, we show that in Phase $1$ the total movement is bounded by $\tdo(\delta)$ within $\frac{\log(d)}{\eta\gamma}$ steps. 
We recall Lemma~\ref{lem:phase1} as follows.
\begingroup
\def\thetheorem{\ref{lem:phase1}}
\begin{lemma}
Let $T_1$ be the length of Phase 1. Assume for any $0\leq t\leq \min(T_1,\frac{\log(d)}{\eta\gamma})-1$, $\n{\xi_t}\leq \frac{C_1L}{\sqrt{b}}\n{x_t-x_{s(t)}},$ where $C_1$ comes from Lemma~\ref{lem:varaincedistance}. Then, there exists large enough constant $c$ such that as long as
\begin{align*}
\eta\leq \frac{1}{cC_1\log(nd)\log(n\frac{\log(d)}{\eta\gamma})\cdot L},
\quad \mu\geq c\log(d)\log^2(\frac{\log(d)}{\eta\gamma}),
\quad \delta\leq \frac{\gamma}{\rho \mu^2},
\end{align*}
we have for every $1\leq t\leq \min(T_1,\frac{\log(d)}{\eta\gamma})$,
$$\n{x_t-x_{t-1}}\leq \frac{\mu}{t}\delta.$$
\end{lemma}
\addtocounter{theorem}{-1}
\endgroup

\begin{proofof}{Lemma~\ref{lem:phase1}}

We prove for every $1\leq t\leq \min(T_1, \frac{\log(d)}{\eta\gamma}),\ \n{x_t-x_{t-1}}\leq \frac{\mu}{t}\delta$ by induction. For the base case, we have $x_1-x_0=-\eta \nabla \hat{f}(x_0)$. Since the gradient at $\tx$ is zero, we have 
\begin{align*}
\n{\nabla \hat{f}(x_0)}&=\n{\nabla \hat{f}(x_0)-\nabla \hat{f}(\tx)}\\
 &\leq L \n{x_0-\tx}\\
 &\leq L\delta,
\end{align*}
where the first inequality holds since $f$ ($\hat{f}$) is $L$-smooth. As long as $\mu \geq \eta L,$ we have $\n{x_1-x_0}\leq \mu \delta.$ 

Fix any $t\leq \min(T_1, \frac{\log(d)}{\eta\gamma}),$ suppose for any $t'\leq t-1, \n{x_{t'}-x_{t'-1}}\leq \frac{\mu}{t'}\delta$, we will prove $\n{x_{t}-x_{t-1}}\leq \frac{\mu}{t}\delta.$ In order to prove $\n{x_{t}-x_{t-1}}\leq \frac{\mu}{t}\delta,$ we will separately bound its projections onto three orthogonal subspaces. Specifically, we consider the following three subspaces:
\begin{itemize}
\item $S$: subspace spanned by the eigenvectors of $\hessian$ with eigenvalues within $[-\gamma, -\frac{\gamma}{\log (d)}]$.
\item $S^\perp_-$: subspace spanned by the eigenvectors of $\hessian$ with eigenvalues within $(-\frac{\gamma}{\log (d)}, 0]$.
\item $S^\perp_+$: subspace spanned by the eigenvectors of $\hessian$ with eigenvalues within $(0, L]$.
\end{itemize}

Regarding the projections onto $S^\perp_-$ and $S^\perp_+$, we will use the following expansion of $x_t-x_{t-1}$, 
\begin{align}
&x_t-x_{t-1}\notag \\
=& -\eta(I-\eta\hessian)^{t-1} \nabla \hat{f}(x_0) +\eta^2\hessian \sum_{\tau=0}^{t-2}(I-\eta\hessian)^{t-2-\tau}\xi_\tau \notag\\
&-\eta \sum_{\tau=0}^{t-2}(I-\eta\hessian)^{t-2-\tau}\Delta_\tau (x_{\tau+1}-x_\tau) -\eta\xi_{t-1} \label{eq:expansion1}
\end{align}
and bound its four terms one by one. In the expansion, we denote $\Delta_\tau := \int_0^1(\nabla^2 \hat{f}(x_\tau+\theta (x_{\tau+1}-x_\tau))-\hessian)d\theta$.

For the projection in subspace $S$, after $\frac{1}{\eta\gamma}$ steps, we cannot bound it using the above expansion since the exponential factor can be very large. Instead, we bound the projection in subspace $S$ by the stopping condition $\n{\proj{S}(x_{t-1}-\tx)}\leq \frac{\delta}{10}$ using an alternative expansion,
\begin{align*}
x_t-x_{t-1} =& -\eta(\nabla \hat{f}(x_{t-1}) +\xi_{t-1})\\
=& -\eta\hessian(x_{t-1}-\tx)-\eta\Delta'_{t-1}(x_{t-1}-\tx)-\eta\xi_{t-1},
\end{align*}
where $\Delta'_{t-1}=\int_0^1(\nabla^2 \hat{f}(\tx+\theta(x_{t-1}-\tx))-\hessian)d\theta.$

We will first bound the projections of $x_t-x_{t-1}$ on $S^\perp_+$ and $S^\perp_-$ by considering the four terms in Eqn.~\ref{eq:expansion1}. For the first term, the projection in subspace $S^\perp_-$ can increase but will not increase by more than a constant factor; the projection in $S^\perp_+$ might start large but will decrease as the number of iterations increases.

\paragraph{Bounding $\n{\proj{S^\perp_-}\eta(I-\eta\hessian)^{t-1}\nabla \hat{f}(x_0)}:$} For this term we will show that its projection on $S^\perp_-$ is small to begin with and cannot be amplified by more than a constant.
Recall that $\nabla \hat{f}(x_0)=\hessian (x_0-\tx)+\Delta(x_0-\tx)$, where $\Delta=\int_0^1(\nabla^2 \hat{f}(\tx+\theta (x_0-\tx))-\hessian)d\theta$. Due to the Hessian lipschitzness of $f$, we have $\n{\Delta}\leq \rho\delta.$ Then, we can bound $\eta \n{\proj{S^\perp_-}(I-\eta\hessian)^{t-1}\nabla \hat{f}(x_0)}$ as follows.
\begin{align*}
\eta \n{\proj{S^\perp_-}(I-\eta\hessian)^{t-1}\nabla \hat{f}(x_0)} =&\eta \left\|\proj{S^\perp_-}(I-\eta\hessian)^{t-1}\left(\hessian (x_0-\tx)+\Delta(x_0-\tx)\right)\right\|\\
\leq& \eta \n{\proj{S^\perp_-}(I-\eta\hessian)^{t-1}\hessian (x_0-\tx)}\\
&+\eta\n{\proj{S^\perp_-}(I-\eta\hessian)^{t-1}\Delta (x_0-\tx)}\\
\leq&\eta(1+\frac{\eta\gamma}{\log(d)})^{\frac{\log (d)}{\eta \gamma}}\frac{\gamma}{\log(d)}\delta +  \eta(1+\frac{\eta\gamma}{\log(d)})^{\frac{\log (d)}{\eta \gamma}} \rho\delta^2\\
\leq&  \frac{e}{\log(d)}\eta \gamma \delta + e\eta\rho \delta^2\\
\leq&  2e\eta \gamma \delta,
\end{align*}
where the last inequality holds as long as $\delta\leq \frac{\gamma}{\rho}$. Since $t\leq \frac{\log(d)}{\eta\gamma}$, we have 
$$\eta \n{\proj{S^\perp_-}(I-\eta\hessian)^{t-1}\nabla \hat{f}(x_0)}\leq \frac{2e\log(d)}{t}\delta.$$

\paragraph{Bounding $\n{\proj{S^\perp_+}\eta(I-\eta\hessian)^{t-1}\nabla \hat{f}(x_0)}:$} 
The key observation here is that $\nabla \hat{f}(x_0)$ can only be large along an eigendirection if the corresponding eigenvalue $\lambda$ is large; however in this case the $(I-\eta\hessian)$ term will also be significantly smaller than 1 in such a direction so the contribution from this direction decreases quickly. More precisely, we have
\begin{align*}
\eta \n{\proj{S^\perp_+}(I-\eta\hessian)^{t-1}\nabla \hat{f}(x_0)} =&\eta \n{\proj{S^\perp_+}(I-\eta\hessian)^{t-1}(\hessian (x_0-\tx)+\Delta(x_0-\tx))}\\
\leq& \eta \n{\proj{S^\perp_+}(I-\eta\hessian)^{t-1}\hessian (x_0-\tx)}\\
&+\eta\n{\proj{S^\perp_+}(I-\eta\hessian)^{t-1}\Delta (x_0-\tx)}\\
\leq& \n{\proj{S^\perp_+}(I-\eta\hessian)^{t-1}\eta\hessian}\delta+\eta\rho\delta^2\\
\leq& \frac{1}{t}\delta + \eta\rho\delta^2,
\end{align*}
where the last inequality holds since $(1-\lambda)^{t-1}\lambda\leq 1/t$ for $0<\lambda\leq 1$. Assuming $\delta \leq \frac{ \gamma}{\rho}$, we can further show 
$$\eta\rho\delta^2\leq  \eta\gamma\delta\leq \frac{\log(d)}{t}\delta .$$ 
Thus, we have 
$$\eta \n{\proj{S^\perp_+}(I-\eta\hessian)^{t-1}\nabla \hat{f}(x_0)}\leq \frac{2\log(d)}{t}\delta.$$

Next we will bound the norm of the variance term.
The main observation here is that based on induction hypothesis, we can have a good upperbound on $\|\xi_\tau\|$. Now, for subspaces $S^\perp_+$ and $S^\perp_-$, we will show that the additional matrices in front of $\xi_\tau$ will not amplify its norm by too much.

\paragraph{Bounding $\n{\proj{S^\perp_+}\eta^2\hessian \sum_{\tau=0}^{t-2}(I-\eta\hessian)^{t-2-\tau}\xi_\tau}$ :}
For each $\tau\leq t-2$, we bound variance term $\n{\xi_\tau}$ as follows, 
\begin{align*}
\n{\xi_\tau} &= \n{v_\tau - \nabla \hat{f}(x_\tau)}\\
&\leq \frac{C_1L}{\sqrt{b}}\n{x_\tau-x_{s(\tau)}}\\
&\leq \frac{C_1L}{m}\n{x_\tau-x_{s(\tau)}}\\
&\leq \frac{C_1L}{m}\sum_{\tau'=s(\tau)+1}^\tau \n{x_{\tau'}-x_{\tau'-1}}\\
&\leq \frac{C_1L}{m}\sum_{\tau'=s(\tau)+1}^\tau \frac{\mu}{\tau'}\delta,
\end{align*}
where the second inequality assumes $b\geq m^2$ and the last inequality is due to the induction hypothesis. If $t\leq 2m$, we bound $\n{\proj{S^\perp_+}\eta^2\hessian \sum_{\tau=0}^{t-2}(I-\eta\hessian)^{t-2-\tau}\xi_\tau}$
as follows.
\begin{align*}
\left\| \proj{S^\perp_+}\eta^2\hessian \sum_{\tau=0}^{t-2}(I-\eta\hessian)^{t-2-\tau}\xi_\tau\right\|&\leq
\eta \sum_{\tau=0}^{t-2}\n{\proj{S^\perp_+} \eta \hessian(I-\eta\hessian)^{t-2-\tau}}\n{\xi_\tau}\\
&\leq \eta \sum_{\tau=0}^{t-2} \frac{1}{t-1-\tau} \left( \frac{C_1L}{m}\sum_{\tau'=s(\tau)+1}^\tau \frac{\mu}{\tau'}\delta\right)\\
&\leq \eta \sum_{\tau=0}^{t-2} \frac{1}{t-1-\tau} \left(\frac{2C_1\log(2m)L}{m}\mu\delta\right)\\
&\leq \frac{4C_1\log^2(2m)}{m}\eta L\mu\delta\\
&\leq \frac{8C_1\log^2(2m)}{t}\eta L\mu\delta,
\end{align*}
where the third inequality holds since $\sum_{\tau'=s(\tau)+1}^\tau \frac{1}{\tau'}\leq \log(\tau)+1\leq \log(2m)+1\leq 2\log(2m).$

If $t>2m$, we bound $\n{\proj{S^\perp_+}\eta^2\hessian \sum_{\tau=0}^{t-2}(I-\eta\hessian)^{t-2-\tau}\xi_\tau}$
as follows.
\begin{align*}
\left\| \proj{S^\perp_+}\eta^2\hessian \sum_{\tau=0}^{t-2}(I-\eta\hessian)^{t-2-\tau}\xi_\tau\right\|&\leq
\eta \sum_{\tau=0}^{t-2}\n{\proj{S^\perp_+} \eta \hessian(I-\eta\hessian)^{t-2-\tau}}\n{\xi_\tau}\\
&\leq \eta\left(\sum_{\tau=0}^{m-1}\frac{1}{t-1-\tau}\n{\xi_\tau} + \eta\sum_{\tau=m}^{t-2}\frac{1}{t-1-\tau}\n{\xi_\tau}\right).
\end{align*} 
We bound these two terms in slightly different ways. For the first term, we have,
\begin{align*}
\eta\sum_{\tau=0}^{m-1}\frac{1}{t-1-\tau}\n{\xi_\tau}
&\leq \eta\sum_{\tau=0}^{m-1}\frac{1}{t-1-\tau}\left(\frac{2C_1\log(m)L}{m}\mu\delta\right)\\
&\leq \eta\sum_{\tau=0}^{m-1}\frac{2}{t}\left(\frac{2C_1\log(m)L}{m}\mu\delta\right)\\
&\leq \frac{4C_1\log(m)}{t}\eta L\mu\delta,
\end{align*}
where the second inequality holds since $t-m>t/2.$ For the second term, we bound it as follows.
\begin{align*}
\eta\sum_{\tau=m}^{t-2}\frac{1}{t-1-\tau}\n{\xi_\tau}
&\leq \eta\sum_{\tau=m}^{t-2}\frac{1}{t-1-\tau}\left( \frac{C_1L}{m}\sum_{\tau'=s(\tau)+1}^\tau \frac{\mu}{\tau'}\delta\right)\\
&\leq \eta\sum_{\tau=m}^{t-2}\frac{1}{t-1-\tau}\left(C_1L \frac{\mu}{s(\tau)+1}\delta\right)\\
&\leq C_1\eta L\mu \delta  \sum_{\tau=m}^{t-2}\frac{1}{t-1-\tau} \cdot \frac{1}{\tau-m+1}\\
&= C_1\eta L\mu \delta  \sum_{\tau=m}^{t-2}\left(\frac{1}{t-1-\tau} + \frac{1}{\tau-m+1}\right)\frac{1}{t-m}\\
&\leq \frac{8C_1\log(\frac{\log(d)}{\eta\gamma})}{t}\eta L \mu \delta 
\end{align*}
where the third inequality holds because $\tau-s(\tau)\leq m.$ Thus, if $t>2m,$ we have 
$$\left\| \proj{S^\perp_+}\eta^2\hessian \sum_{\tau=0}^{t-2}(I-\eta\hessian)^{t-2-\tau}\xi_\tau\right\| \leq \frac{8C_1\log(m\frac{\log(d)}{\eta\gamma})}{t}\eta L\mu\delta.$$

Thus, combining two cases when $t\leq 2m$ and $t>2m$, we know
\begin{align*}
\left\| \proj{S^\perp_+}\eta^2\hessian \sum_{\tau=0}^{t-2}(I-\eta\hessian)^{t-2-\tau}\xi_\tau\right\|
\leq \max\left(8C_1\log^2(2m),8C_1\log(m\frac{\log(d)}{\eta\gamma})\right)\frac{1}{t}\eta L\mu\delta.
\end{align*}

\paragraph{Bounding $\n{\proj{S^\perp_-}\eta^2\hessian \sum_{\tau=0}^{t-2}(I-\eta\hessian)^{t-2-\tau}\xi_\tau}$ :}
Let's now consider the projection on the $S^\perp_-$ subspace. 
\begin{align*}
&\left\|\proj{S^\perp_-}\eta^2\hessian \sum_{\tau=0}^{t-2}(I-\eta\hessian)^{t-2-\tau}\xi_\tau\right\|\\
\leq& \eta^2 \sum_{\tau=0}^{t-2}\n{\proj{S^\perp_-}\hessian}\n{\proj{S^\perp_-}(I-\eta \hessian)^{t-2-\tau}}\n{\xi_\tau}\\
\leq& \eta^2 \frac{\gamma}{\log(d)} (1+\frac{\eta\gamma}{\log(d)})^{\frac{\log(d)}{\eta\gamma}} \sum_{\tau=0}^{t-2}\n{\xi_\tau}\\
\leq& \eta^2 \frac{e\gamma}{\log(d)} \left(\sum_{\tau=0}^{m-1} \frac{2C_1\log(m)L}{m}\mu\delta +\sum_{\tau=m}^{t-2} C_1 L\frac{1}{\tau-m+1}\mu\delta\right)\\
\leq& 2\log(m\frac{\log(d)}{\eta\gamma})e C_1 \eta L \mu\delta\frac{\eta\gamma}{\log(d)}\\
\leq& 2\log(m\frac{\log(d)}{\eta\gamma})e C_1 \eta L \mu\delta\frac{\log(d)}{t\log(d)}\\
=& \frac{2e C_1\log(m\frac{\log(d)}{\eta\gamma})}{t} \eta L\mu\delta.
\end{align*}

Next we bound the Hessian changing term. This is easy because this term is actually of order $\delta^2$ where $\delta$ is the radius of the initial perturbation. Therefore we can bound it as long as we make $\delta$ small.

\paragraph{Bounding $\n{\proj{S^\perp_+\cap S^\perp_-}\eta \sum_{\tau=0}^{t-2}(I-\eta\hessian)^{t-2-\tau}\Delta_\tau (x_{\tau+1}-x_\tau)}$ :} First, we bound $\n{\Delta_\tau}$ for each $\tau\leq t-2.$
\begin{align*}
\n{\Delta_\tau}\leq & \rho\max(\n{x_{\tau+1}-\tx}, \n{x_\tau -\tx})\\
\leq &\rho(\sum_{\tau'=1}^{\tau+1}\n{x_{\tau'}-x_{\tau'-1}} +\n{x_0-\tx})\\
\leq &\rho(\sum_{\tau'=1}^{\tau+1} \frac{1}{\tau'}\mu\delta+\delta)\\
\leq &3\log(\frac{\log(d)}{\eta\gamma})\rho\mu\delta,
\end{align*}
where the third inequality uses the induction hypothesis. Then, for the Hessian changing term, we have
\begin{align*}
&\left\|\proj{S^\perp_+\cap S^\perp_-} \eta \sum_{\tau=0}^{t-2}(I-\eta\hessian)^{t-2-\tau}\Delta_\tau (x_{\tau+1}-x_\tau)\right\|\\
\leq& \eta\sum_{\tau=0}^{t-2}\n{\proj{S^\perp_+\cap S^\perp_-}(I-\eta\hessian)^{t-2-\tau}}\n{\Delta_\tau}\n{(x_{\tau+1}-x_\tau)}\\
\leq& \eta\sum_{\tau=0}^{t-2}(1+\frac{\eta\gamma}{\log(d)})^{\frac{\log(d)}{\eta\gamma}} \cdot 3\log(\frac{\log(d)}{\eta\gamma})\rho\mu\delta \frac{1}{\tau+1}\mu\delta\\
\leq& \eta\sum_{\tau=0}^{t-2}e \cdot 3\log(\frac{\log(d)}{\eta\gamma})\rho\mu\delta \frac{1}{\tau+1}\mu\delta\\
\leq& 6e \log^2(\frac{\log(d)}{\eta\gamma})\eta\rho\mu^2\delta^2\\
\leq& 6e\log^2(\frac{\log(d)}{\eta\gamma})\eta\gamma\delta\\
\leq& 6e\log^2(\frac{\log(d)}{\eta\gamma})\log(d)\frac{1}{t}\delta,
\end{align*}
where the second last inequality holds as long as $\delta\leq \frac{\gamma}{\rho \mu^2}.$

Next, we bound the norm of the error in the last gradient estimate. This follows immediately from induction hypothesis.

\paragraph{Bounding $\n{\eta\xi_{t-1}}$: }
For the last term $\eta \xi_{t-1}.$ If $t\leq 2m$, we have
\begin{align*}
\n{\eta\xi_{t-1}}&\leq \eta \frac{2C_1\log(2m)L}{m}\mu\delta\\
&\leq 4C_1\log(2m)\frac{1}{t}\eta L\mu \delta.
\end{align*}
If $t>2m$, we have 
\begin{align*}
\n{\eta\xi_{t-1}}&\leq \eta \frac{C_1 L }{s(t-1)+1}\mu\delta\\
&\leq \eta \frac{C_1 L }{t-m}\mu\delta\\
&\leq \frac{2}{t}\eta C_1 L\mu\delta.
\end{align*}

Overall, we have 
\begin{align*}
\n{\eta\xi_{t-1}}\leq 4C_1\log(2m)\frac{1}{t}\eta L\mu\delta.
\end{align*}

Until now, we have already bounded the projection of $x_t-x_{t-1}$ in subspace $S^\perp_+$ and $S^\perp_-$. Finally, we bound the projection of $x_t-x_{t-1}$ on the $S$ subspace. If $t-1\leq \frac{1}{\eta\gamma}$, we bound it using the expansion in Eqn.~\ref{eq:expansion1} similar as above. If $t-1> \frac{1}{\eta\gamma}$, we use the stopping condition to bound the projection on $S$.

\paragraph{Bounding $\n{\proj{S} (x_t-x_{t-1})}$}
If $t-1\leq \frac{1}{\eta\gamma},$ the exponential factor $(1+\eta\gamma)^{t-1}$ is still a constant. Similar as the analysis for the projection on subspace $S^\perp_-$, we have the following bound,
\begin{align*}
&\left\|\proj{S}\eta(I-\eta\hessian)^{t-1}\nabla \hat{f}(x_0)\right\|\leq \frac{2e\log(d)}{t}\delta,\\
&\left\|\proj{S}\eta^2\hessian \sum_{\tau=0}^{t-2}(I-\eta\hessian)^{t-2-\tau}\xi_\tau\right\|
\leq \frac{2e C_1\log(m\frac{\log(d)}{\eta\gamma})\log(d)}{t} \eta L\mu\delta,\\
&\left\|\proj{S}\eta \sum_{\tau=0}^{t-2}(I-\eta\hessian)^{t-2-\tau}\Delta_\tau (x_{\tau+1}-x_\tau)\right\|\leq 6e\log^2(\frac{\log(d)}{\eta\gamma})\log(d)\frac{1}{t}\delta.
\end{align*}

If $t-1> \frac{1}{\eta\gamma},$ according to the stopping condition of Phase 1, we know $\n{\proj{S} (x_{t-1}-\tx)}\leq \frac{\delta}{10}.$ In order to better exploit this property, we express $x_t-x_{t-1}$ in the following way,
\begin{align*}
x_t-x_{t-1} =& -\eta(\nabla \hat{f}(x_{t-1}) +\xi_{t-1})\\
=& -\eta\hessian(x_{t-1}-\tx)-\eta\Delta_{t-1}(x_{t-1}-\tx)-\eta\xi_{t-1},
\end{align*}
where $\Delta_{t-1}=\int_0^1(\nabla^2 \hat{f}(\tx+\theta(x_{t-1}-\tx))-\hessian)d\theta.$ For the first term, we have 
\begin{align*}
\n{\proj{S} \eta\hessian(x_{t-1}-\tx)}\leq \eta\gamma\n{\proj{S} (x_{t-1}-\tx)}\leq \eta\gamma\frac{\delta}{10}\leq \frac{\log(d)}{10t}\delta.
\end{align*}
For the hessian changing term, we have
\begin{align*}
\n{\proj{S} \eta\Delta_{t-1}(x_{t-1}-\tx)} &\leq \n{ \eta\Delta_{t-1}(x_{t-1}-\tx)}\\
&\leq \eta \rho\ns{x_{t-1}-\tx} \\
&\leq \eta \rho( 3\log(\frac{\log(d)}{\eta\gamma})\mu\delta)^2\\
&\leq 9\log^2(\frac{\log(d)}{\eta\gamma})\eta\gamma\delta\\
&\leq 9\log^2(\frac{\log(d)}{\eta\gamma})\log(d)\frac{1}{t}\delta
\end{align*}
where the second last inequality assumes $\delta\leq \frac{\gamma}{\rho \mu^2}.$

Combining the bound for the projections onto all three subspaces, we know there exists absolute constant $c$, such that
\begin{align*}
\n{x_t-x_{t-1}}\leq \frac{c}{2}\log(d)\log^2(\frac{\log(d)}{\eta\gamma})\frac{1}{t}\delta+ \frac{c}{2} C_1\log(nd)\log(n\frac{\log(d)}{\eta\gamma})\frac{1}{t}\eta L\mu\delta,
\end{align*}
assuming $\delta\leq \min(\frac{\gamma}{\rho}, \frac{\gamma}{\rho \mu^2})$. Now, we know $\n{x_t-x_{t-1}}\leq \frac{1}{t}\mu\delta$, as long as 
\begin{align*}
&\eta\leq \frac{1}{cC_1\log(nd)\log(n\frac{\log(d)}{\eta\gamma})\cdot L},\\
&\mu\geq c\log(d)\log^2(\frac{\log(d)}{\eta\gamma}),\\
&\delta\leq \frac{\gamma}{\rho \mu^2}.
\end{align*}
\end{proofof}

\subsection{Proofs of Phase 2}
We have shown that at the end of Phase 1, $x_{T_1}-\tx$ becomes aligned with the negative directions. Based on this property, we show the projection of $x_t-\tx$ on $S$ subspace grows exponentially and exceeds the threshold distance within $\tdo(\frac{1}{\eta\gamma})$ steps. We use the following expansion,
\begin{align*}
x_t-\tx = (I-\eta\hessian)(x_{t-1}-\tx)-\eta\Delta_{t-1}(x_{t-1}-\tx)-\eta\xi_{t-1},
\end{align*}
where $\Delta_{t-1}=\int_0^1(\nabla^2 \hat{f}(\tx+\theta(x_{t-1}-\tx))-\hessian)d\theta.$
Intuitively, if we only have the first term, it's clear that $\n{\proj{S}(x_t-\tx)}\geq (1+\frac{\eta\gamma}{\log(d)})\n{\proj{S}(x_{t-1}-\tx)}.$ We show that the Hessian changing term and the variance term are negligible in the sense that $\n{\eta\Delta_{t-1}(x_{t-1}-\tx)-\eta\xi_{t-1}}\leq \frac{\eta\gamma}{2\log(d)}\n{\proj{S}(x_{t-1}-\tx)}$. The Hessian changing term can be easily bounded because the threshold distance $\mathL=\tdo(\frac{\gamma}{\rho}).$ We will bound the variance by showing that $\n{x_t-x_{t-1}}\leq \tdo(1/t) \n{x_{t-1}-\tx}$. We also need $x_{t-1}-\tx$ to be roughly aligned with the negative directions in order to bound $\n{x_{t-1}-\tx}$ by $\tdo(1)\n{\proj{S}(x_{t-1}-\tx)}.$ 

There are several key differences between Phase 1 and Phase 2 . First, we use Lemma~\ref{lem:coupled_gradient_estimator} to bound the variance (this is effective because the point does not move far in Phase 1), but we use Lemma~\ref{lem:varaincedistance} to bound variance in Phase 2 (this is effective because in Phase 2 the projection in the most negative eigenvalue is already large). Second, in Phase 1 we need to analyze the difference between two points, and the direction $e_1$ is dominating. In Phase 2 we can analyze the dynamics of a single point, and focus on the entire subspace with eigenvalues less than $-\gamma/\log d$ instead of a single $e_1$ direction.

\begin{lemma}\label{lem:phase2}
Let the threshold distance $\mathL := \frac{\gamma}{C_3\rho}$. Let $T$ be the length of the super epoch, which means $T:=\inf\{t|\ \n{x_t-\tx}\geq \mathL\}$. Assume for any $0\leq t\leq T-1$, $\n{\xi_t}\leq \frac{C_1 L}{\sqrt{b}}\n{x_t-x_{s(t)}}$, where $C_1$ comes from Lemma~\ref{lem:varaincedistance}. Assume Phase 1 is successful in the sense that
\begin{align*}
 \frac{1}{\eta\gamma}\leq T_1\leq \frac{\log(d)}{\eta\gamma},\qquad
&\forall\ 1\leq t\leq T_1,\ \n{x_t-x_{t-1}}\leq \frac{C_2}{t}\delta,\\
\n{\proj{S}(x_{T_1}-\tx)}\geq \frac{\delta}{10},\qquad
&\forall\ 0\leq t\leq T_1,\ \n{x_{t}-\tx}\leq C\frac{\delta}{10},
\end{align*} 
where $C_2$ comes from Lemma~\ref{lem:phase1} and $C$ comes from Lemma~\ref{lem:distanceproperty}.
There exists large enough absolute constant $c$ such that as long as 
\begin{align*}
&\eta\leq \frac{1}{L\cdot cCC_1\left(\log^2(n)+\log(n\frac{\log(d)\log(\frac{\gamma}{\rho\delta})}{\eta\gamma})\right)},\\
&C_3\geq c\left(C_2 + C\log(\frac{\log(d)\log(\frac{\gamma}{\rho\delta})}{\eta\gamma})\log(d)\log(\frac{\gamma}{\rho\delta})\right),\\
&b\geq n^{2/3}\cdot c\left(C\log(d)\log(\frac{\gamma}{\rho\delta})\left(C_2 + C\log(\frac{\log(d)\log(\frac{\gamma}{\rho\delta})}{\eta\gamma})\log(d)\log(\frac{\gamma}{\rho\delta}) \right) \right)^{2/3},
\end{align*}
we have 
$$T\leq T_1+ \frac{4\log(d)\log(\frac{10\gamma}{\rho\delta})}{\eta\gamma}\leq \frac{\log(d)+4\log(d)\log(\frac{10\gamma}{\rho\delta})}{\eta\gamma}.$$
\end{lemma}

\begin{proofof}{Lemma~\ref{lem:phase2}}
Let $T_{\max}=T_1+\frac{4\log(d)\log(\frac{10\gamma}{\rho\delta})}{\eta\gamma}$. If there exists $t\leq T_{\max}-1, \n{x_t-\tx}\geq \mathL$, we are done. Otherwise, we show $\n{x_t-\tx}$ increases exponentially and will become larger than $\mathL$ after $T_{\max}$ steps. 

Formally, we show the following four hypotheses hold for any $T_1\leq t\leq T_{\max}$ by induction, 
\begin{enumerate}
\item $$\n{\proj{S}(x_t-\tx)}\geq (1+\frac{\eta\gamma}{2\log(d)})^{t-T_1}\n{\proj{S}(x_{T_1}-\tx)};$$
\item $$\frac{\n{\proj{S^\perp}(x_t-\tx)}}{\n{\proj{S}(x_t-\tx)}}\leq C(1+\frac{\eta\gamma}{4\log(d)\log(\frac{10\gamma}{\rho\delta})})^{t-T_1},$$ where $S^\perp$ denotes the orthogonal subspace of $S$;
\item For any $0\leq \tau\leq t-1$, we have $$\n{x_t-\tx}\geq \n{\proj{S}(x_t-\tx)}\geq\frac{1}{eC+1}\n{x_\tau-\tx};$$
\item For any $1\leq \tau\leq t$, we have $$\n{x_\tau -x_{\tau-1}}\leq \frac{\mu}{\tau}\max(\n{x_{\tau-1}-\tx}, \frac{\delta}{10}),$$ where $\mu=\tdo(1).$
\end{enumerate}

Hypothesis 1 is our goal, which is showing the distance to the initial point increases exponentially in Phase 2. We use hypothesis 4 to bound the variance term. We also need Hypothesis 2 and 3 for some technical reason, which will only be clear in the later proof. Basically, hypothesis 2 guarantees that $x_t-\tx$ roughly aligns with the $S$ subspace. Hypothesis 3 guarantees that the distance to the initial point cannot shrink by too much. 
 
Let's first check the initial case first. If $t=T_1$, the first hypothesis clearly holds. For the second hypothesis, we have $$\frac{\n{\proj{S^\perp}(x_{T_1}-\tx)}}{\n{\proj{S}(x_{T_1}-\tx)}}\leq \frac{\n{x_{T_1}-\tx}}{\n{\proj{S}(x_{T_1}-\tx)}}\leq C.$$ The third hypothesis holds because $\n{x_{T_1}-\tx}\geq \n{\proj{S}(x_{T_1}-\tx)}\geq \delta/10$ and $\n{x_t-\tx}\leq C\delta/10$ for any $t\leq T_1.$ Since $\n{x_t-x_{t-1}}\leq \frac{C_2}{t}\delta$ for any $1\leq t\leq T_1,$ the fourth hypothesis holds as long as $\mu\geq 10C_2.$ 

Now, fix $T_1<t\leq T_{\max}$, assume all four hypotheses hold for every $T_1\leq t'\leq t-1$, we prove they still hold for $t$. 
\paragraph{Proving Hypothesis 4:} In order to prove Hypothesis 4, we only need to show $\n{x_t-x_{t-1}}\leq \frac{\mu}{t}\max(\n{x_{t-1}-\tx},\delta/10).$ Let $S^+$ be the subspace spanned by all the eigenvectors of $\hessian$ with positive eigenvalues. Let $S^-$ be the subspace spanned by all the eigenvectors of $\hessian$ with non-positive eigenvalues. We project $x_t-x_{t-1}$ into these two subspaces and bound them separately. 

\paragraph{Bounding $\n{\proj{S^-}(x_t-x_{t-1})}$:} Consider the following expansion of $x_t-x_{t-1}:$
\begin{align*}
x_t-x_{t-1} =& -\eta(\nabla \hat{f}(x_{t-1}) +\xi_{t-1})\\
=& -\eta\hessian(x_{t-1}-\tx)-\eta\Delta_{t-1}(x_{t-1}-\tx)-\eta\xi_{t-1},
\end{align*}
where $\Delta_{t-1}=\int_0^1(\nabla^2 \hat{f}(\tx+\theta(x_{t-1}-\tx))-\hessian)d\theta.$ We bound $\proj{S^-}(x_t-x_{t-1})$ by separately considering these three terms. 

The first term can be bounded because within subspace $S^-$, the largest singular value of $\hessian$ is just $\gamma$. Precisely, we have
\begin{align*}
\n{\proj{S^-}\eta\hessian(x_{t-1}-\tx)}\leq& \eta\gamma\n{x_{t-1}-\tx}\\
\leq& \frac{\left(\log(d) + 4\log(d)\log(\frac{10\gamma}{\rho\delta})\right)}{t}\n{x_{t-1}-\tx},
\end{align*}
where the second inequality holds because $t\leq T_{\max}\leq \frac{\left(\log(d) + 4\log(d)\log(\frac{10\gamma}{\rho\delta})\right)}{\eta\gamma}.$

Since $f$ is Hessian lipschitz and the total distance is upper bounded by $\frac{\gamma}{C_3 \rho},$ the second term can also be well bounded. We have,
\begin{align*}
\n{\proj{S^-}\eta\Delta_{t-1}(x_{t-1}-\tx)}\leq& \n{\eta\Delta_{t-1}(x_{t-1}-\tx)}\\
\leq& \eta\rho\n{x_{t-1}-\tx}\n{x_{t-1}-\tx}\\
\leq& \eta\rho\mathL \n{x_{t-1}-\tx}\\
\leq& \eta \frac{\gamma}{C_3}\n{x_{t-1}-\tx}\\
\leq& \frac{\log(d) + 4\log(d)\log(\frac{10\gamma}{\rho\delta})}{C_3 t}\n{x_{t-1}-\tx},
\end{align*}
where the second inequality holds due to the Hessian-lipshcitzness of $f$.

We can bound the variance term using Hypothesis 3 and 4. 
Precisely, we have
\begin{align*}
\n{\eta\xi_{t-1}} \leq& \eta \frac{m}{\sqrt{b}}\frac{C_1 L}{m}\sum_{\tau=s(t-1)+1}^{t-1} \n{x_\tau-x_{\tau-1}}\\
 \leq&\eta \frac{m}{\sqrt{b}}\frac{C_1 L}{m}\sum_{\tau=s(t-1)+1}^{t-1} \frac{\mu}{\tau}\max(\n{x_{\tau-1}-\tx}, \frac{\delta}{10})\\
\leq& \eta \frac{m}{\sqrt{b}}\frac{C_1 L}{m}\sum_{\tau=s(t-1)+1}^{t-1} \frac{\mu}{\tau}(eC+1)\n{x_{t-1}-\tx}, 
\end{align*}
where the last inequality holds requires $\n{x_{t-1}-\tx}\geq \frac{1}{eC+1}\max(\n{x_{\tau-1}-\tx}, \frac{\delta}{10})$ for any $\tau\leq t-1.$ According to induction hypothesis 3, we have $\n{x_{t-1}-\tx}\geq \frac{1}{eC+1}\n{x_{\tau-1}-\tx}$ for any $\tau\leq t-1.$ By induction hypothesis 1, we have $\n{x_{t-1}-\tx} \geq \n{\proj{S}(x_{t-1}-\tx)}
\geq (1+\frac{\eta\gamma}{2})^{t-1-T_1}\n{\proj{S}(x_{T_1}-\tx)}
\geq\frac{\delta}{10}$. Using the same analysis in Lemma~\ref{lem:phase1}, we further have 
$$\n{\eta\xi_{t-1}}\leq \frac{m}{\sqrt{b}} 4(eC+1)C_1\log(2m)\frac{1}{t}\eta L\mu\n{x_{t-1}-\tx}.$$

\paragraph{Bounding $\n{\proj{S^+}(x_t-x_{t-1})}:$} For the projection onto $S^+$, we use the following expansion:
\begin{align*}
x_t-x_{t-1}=& -\eta(I-\eta\hessian)^{t-1} \nabla \hat{f}(x_0) +\eta^2\hessian \sum_{\tau=0}^{t-2}(I-\eta\hessian)^{t-2-\tau}\xi_\tau \\
&-\eta \sum_{\tau=0}^{t-2}(I-\eta\hessian)^{t-2-\tau}\Delta_\tau (x_{\tau+1}-x_\tau) -\eta\xi_{t-1},
\end{align*}
Similar as the analysis in Lemma~\ref{lem:phase1}, we can bound the first term as follows,
\begin{align*}
\n{\proj{S^+} \eta(I-\eta\hessian)^{t-1} \nabla \hat{f}(x_0)}\leq \frac{2\log(d)}{t}\delta\leq \frac{20\log(d)}{t}\n{x_{t-1}-\tx},
\end{align*}
where the second inequality holds because $\n{x_{t-1}-\tx}\geq \delta/10.$

Using a similar analysis as in Lemma~\ref{lem:phase1}, we have the following bound for the second term,
\begin{align*}
&\left\| \proj{S^+} \eta^2\hessian \sum_{\tau=0}^{t-2}(I-\eta\hessian)^{t-2-\tau}\xi_\tau\right\|\\
\leq& \frac{m}{\sqrt{b}}(eC+1)\max(8C_1\log^2(2m), 8C_1\log(mT_{\max}))\frac{1}{t}\eta L \mu\n{x_{t-1}-\tx}.
\end{align*}

For the hessian changing term, we have 
\begin{align*}
&\left\| \proj{S^+} \eta \sum_{\tau=0}^{t-2}(I-\eta\hessian)^{t-2-\tau}\Delta_\tau (x_{\tau+1}-x_{\tau})\right\|\\
\leq& \eta\sum_{\tau=0}^{t-2}\n{\Delta_\tau}\n{x_{\tau+1}-x_{\tau}}\\
\leq& \eta\sum_{\tau=0}^{t-2}\frac{\gamma}{C_3} (eC+1)\frac{\mu}{\tau+1}\n{x_{t-1}-\tx}\\
\leq& 2\log(T_{\max})\eta\gamma (eC+1)\frac{\mu}{C_3}\n{x_{t-1}-\tx}\\
\leq& 2(eC+1)\log(T_{\max})\frac{\mu}{C_3}\frac{\log(d) + 4\log(d)\log(\frac{10\gamma}{\rho\delta})}{t}\n{x_{t-1}-\tx}\\
\leq& 2(eC+1)\log(T_{\max})\frac{\log(d) + 4\log(d)\log(\frac{10\gamma}{\rho\delta})}{t}\n{x_{t-1}-\tx},
\end{align*}
where the last inequality holds as long as $C_3\geq \mu.$

Overall, we can upper bound $\n{x_t-x_{t-1}}$ as follows,
\begin{align*}
&\n{x_t-x_{t-1}}\\
\leq& \Big(20\log(d)+(\frac{1}{C_3}+1+2(eC+1)\log(T_{\max}))\big(\log(d) + 4\log(d)\log(\frac{10\gamma}{\rho\delta})\big)\Big)\frac{1}{t}\n{x_{t-1}-\tx}\\
& + \left(4(eC+1)C_1\log(2m)+(eC+1)\max(8C_1\log^2(2m), 8C_1\log(mT_{\max}))\right)\frac{1}{t}\eta L\mu\n{x_{t-1}-\tx}\\
\leq& \Big(20\log(d)+(2+2(eC+1)\log(T_{\max}))\big(\log(d) + 4\log(d)\log(\frac{10\gamma}{\rho\delta})\big)\Big)\frac{1}{t}\n{x_{t-1}-\tx}\\
& + \left(4(eC+1)C_1\log(2n)+(eC+1)\max(8C_1\log^2(2n), 8C_1\log(nT_{\max}))\right)\frac{1}{t}\eta L\mu\n{x_{t-1}-\tx},
\end{align*}
assuming $C_3\geq 1.$
As long as 
$$\eta\leq \frac{1}{2L\cdot \left(4(eC+1)C_1\log(2n)+(eC+1)\max(8C_1\log^2(2n), 8C_1\log(nT_{\max}))\right)}$$
and 
$$\mu\geq 2\left(20\log(d)+(2+2(eC+1)\log(T_{\max}))\left(\log(d) + 4\log(d)\log(\frac{10\gamma}{\rho\delta})\right)\right),$$
we have $\n{x_t-x_{t-1}}\leq \frac{\mu}{t}\n{x_{t-1}-\tx}.$

\paragraph{Proving Hypothesis 2:} In order to prove condition 2 holds for time $t$, we only need to show
$$\frac{\n{\proj{S^\perp}(x_t-\tx)}}{\n{\proj{S}(x_t-\tx)}}\leq (1+\frac{\eta\gamma}{4\log(d)\log(\frac{10\gamma}{\rho\delta})})P_{t-1},$$
where $P_{t-1}:= C(1+\frac{\eta\gamma}{4\log(d)\log(\frac{10\gamma}{\rho\delta})})^{t-1-T_1}.$

We can express $x_t-\tx$ as follows,
\begin{align*}
x_t-\tx = (I-\eta\hessian)(x_{t-1}-\tx)-\eta\Delta_{t-1}(x_{t-1}-\tx)-\eta\xi_{t-1}.
\end{align*}
Assuming $\n{\eta\Delta_{t-1}(x_{t-1}-\tx)}+\n{\eta\xi_{t-1}}\leq \tdc \eta\gamma\n{x_{t-1}-\tx}, \tdc=\tdo(1)$, we have 
$$\n{\proj{S^\perp}(x_t-\tx)} \leq (1+\frac{\eta\gamma}{\log(d)})\n{\proj{S^\perp}(x_{t-1}-\tx)} + \tdc \eta\gamma\n{x_{t-1}-\tx}$$ 
and 
$$\n{\proj{S}(x_t-\tx)} \geq (1+\frac{\eta\gamma}{\log(d)})\n{\proj{S}(x_{t-1}-\tx)} - \tdc \eta\gamma\n{x_{t-1}-\tx}.$$ 
Then, we have 
\begin{align*}
\frac{\n{\proj{S^\perp}(x_t-\tx)}}{\n{\proj{S}(x_t-\tx)}}\leq& \frac{P_{t-1}(1+\frac{\eta\gamma}{\log(d)}) + (P_{t-1}+1) \tdc \eta\gamma}{1+\frac{\eta\gamma}{\log(d)} - (P_{t-1}+1) \tdc \eta\gamma}\\
=& P_{t-1}\left( \frac{1+\frac{\eta\gamma}{\log(d)}+(1+\frac{1}{P_{t-1}})\tdc \eta\gamma}{1+\frac{\eta\gamma}{\log(d)} - (P_{t-1}+1) \tdc \eta\gamma}\right)\\
=& P_{t-1}\left( 1+ \frac{(1+\frac{1}{P_{t-1}})\tdc \eta\gamma+ (P_{t-1}+1) \tdc \eta\gamma}{1+\frac{\eta\gamma}{\log(d)} - (P_{t-1}+1) \tdc \eta\gamma}\right)\\
\leq& P_{t-1}\left( 1+ (1+\frac{1}{P_{t-1}}+P_{t-1}+1) \tdc \eta\gamma\right)\\
\leq& P_{t-1}\left( 1+ (3+eC) \tdc \eta\gamma\right),
\end{align*}
where the second last inequality holds as long as $(P_{t-1}+1)\tdc\leq (eC+1)\tdc\leq 1/\log(d)$ and the last inequality holds because $1\leq P_{t-1}\leq eC.$ Now, as long as $\tdc\leq \frac{1}{(3+eC)4\log(d)\log(\frac{10\gamma}{\rho\delta})}$, we have 
\begin{align*}
\frac{\n{\proj{S^\perp}(x_t-\tx)}}{\n{\proj{S}(x_t-\tx)}}\leq& (1+\frac{\eta\gamma}{4\log(d)\log(\frac{10\gamma}{\rho\delta})})P_{t-1}\\
\leq & C(1+\frac{\eta\gamma}{4\log(d)\log(\frac{10\gamma}{\rho\delta})})^{t-T_1}.
\end{align*}

For the hessian changing term, we have $\n{\eta\Delta_{t-1}(x_{t-1}-\tx)}\leq \frac{1}{C_3}\eta\gamma\n{x_{t-1}-\tx}$. For the variance term, according to the previous analysis and the choosing of $\eta,$ we have 
\begin{align*}
\n{\eta\xi_{t-1}} \leq \frac{m}{2\sqrt{b}}\mu\frac{1}{t}\n{x_{t-1}-\tx}\leq \frac{m}{2\sqrt{b}}\mu\eta\gamma\n{x_{t-1}-\tx}
\end{align*}
where the second inequality holds because $t\geq T_1\geq \frac{1}{\eta\gamma}.$
As long as $C_3\geq \frac{2}{\tdc}$ and $b\geq (\frac{\mu}{\tdc})^{2/3}n^{2/3}$, we have $\n{\eta\Delta_{t-1}(x_{t-1}-\tx)}+\n{\eta\xi_{t-1}}\leq \tdc\eta\gamma\n{x_{t-1}-\tx}.$ 

\paragraph{Proving Hypothesis 1.}
In order to prove hypothesis 1, we show $\n{\proj{S}(x_t-\tx)}\geq (1+\frac{\eta\gamma}{2\log(d)})\n{\proj{S}(x_{t-1}-\tx)}.$ We know,
\begin{align*}
\n{\proj{S}(x_t-\tx)}\geq& (1+\frac{\eta\gamma}{\log(d)})\n{\proj{S}(x_{t-1}-\tx)} -\tdc \eta\gamma\n{x_{t-1}-\tx}\\
\geq& (1+\frac{\eta\gamma}{\log(d)})\n{\proj{S}(x_{t-1}-\tx)}- (eC+1)\tdc \eta\gamma\n{\proj{S}(x_{t-1}-\tx)}\\
\geq& (1+\frac{\eta\gamma}{2\log(d)})\n{\proj{S}(x_{t-1}-\tx)},
\end{align*}
where the last inequality holds as long as $\tdc\leq \frac{1}{2(eC+1)\log(d)}.$

\paragraph{Proving Hypothesis 3.}
For $\tau\leq t-2,$ we have 
\begin{align*}
\n{x_t-\tx}\geq& \n{\proj{S}(x_t-\tx)}\\
\geq& \n{\proj{S}(x_{t-1}-\tx)}\\
\geq& \frac{1}{eC+1}\n{x_\tau-\tx}, 
\end{align*}
where the second inequality holds because $\n{\proj{S}(x_t-\tx)}\geq(1+\frac{\eta\gamma}{2\log(d)})\n{\proj{S}(x_{t-1}-\tx)}$ and the last inequality holds due to the induction hypothesis 3.

Since $\n{\proj{S}(x_{t-1}-\tx)}\geq \frac{1}{eC+1}\n{x_{t-1}-\tx}$, we also have 
\begin{align*}
\n{x_t-\tx}\geq& \n{\proj{S}(x_t-\tx)}\\
\geq& \n{\proj{S}(x_{t-1}-\tx)}\\
\geq& \frac{1}{eC+1}\n{x_{t-1}-\tx}. 
\end{align*} 

Thus, there exists large enough absolute constant $c$ such that the induction holds as long as 
\begin{align*}
&\eta\leq \frac{1}{L\cdot cCC_1\left(\log^2(n)+\log(n\frac{\log(d)\log(\frac{\gamma}{\rho\delta})}{\eta\gamma})\right)},\\
&C_3\geq c\left(C_2 + C\log(\frac{\log(d)\log(\frac{\gamma}{\rho\delta})}{\eta\gamma})\log(d)\log(\frac{\gamma}{\rho\delta})\right)\\
&b\geq cn^{2/3}\left(C\log(d)\log(\frac{\gamma}{\rho\delta})\left(C_2 + C\log(\frac{\log(d)\log(\frac{\gamma}{\rho\delta})}{\eta\gamma})\log(d)\log(\frac{\gamma}{\rho\delta}) \right) \right)^{2/3}.
\end{align*}

Finally, we have 
\begin{align*}
\n{x_{T_{\max}} -\tx}\geq& \n{\proj{S}(x_{T_{\max}}-\tx)}\\
\geq& (1+\frac{\eta\gamma}{2\log(d)})^{T_{\max}-T_1}\n{\proj{S}(x_{T_1}-\tx)}\\
\geq& (1+\frac{\eta\gamma}{2\log(d)})^{\frac{4\log(d)\log(\frac{10\gamma}{\rho\delta})}{\eta\gamma}}\frac{\delta}{10}\\
\geq& \frac{\gamma}{\rho}
\geq \frac{\gamma}{C_3\rho}:=\mathL.
\end{align*}

\end{proofof}
\subsection{Proof of Lemma~\ref{lem:lemma2}}
Finally, we combine the analysis for Phase 1 and Phase 2 to show that starting from a randomly perturbed point, with at least constant probability the function value decreases significantly after a super epoch. 
\begin{lemma}\label{lem:lemma2}
Let $\tx$ be the initial point with gradient $\n{\nabla f(\tx)}\leq \mathG$ and $\lambda_{\min}(\hessian)=-\gamma<0$. 
Define stabilized function $\hat{f}$ such that $\hat{f}(x):=f(x)-\inner{\nabla f(\tx)}{x-\tx}$. Let $\{x_t\}$ be the iterates of SVRG running on $\hat{f}$ starting from $x_0$, which is the perturbed point of $\tx$. Let $T$ be the length of the current super epoch. There exists $\eta=\tdo(1/L), b=\tdo(n^{2/3}), m=n/b, \delta=\tdo(\min(\frac{\gamma}{\rho}, \frac{m\gamma}{\rho'})), \mathG=\tdo(\frac{\gamma^2}{\rho}), \mathL=\tdo(\frac{\gamma}{\rho}), T_{\max}=\tdo(\frac{1}{\eta\gamma})$ such that with probability at least $1/8,$
$$f(x_T)-f(\tx)\leq -C_5\cdot \frac{\gamma^3}{\rho^2};$$
and with high probability, 
$$f(x_T)-f(\tx)\leq \frac{C_5}{20}\cdot \frac{\gamma^3}{\rho^2};$$
where $C_5=\tdtheta(1)$ and $T\leq T_{\max}.$
\end{lemma}
\begin{proofof}{Lemma~\ref{lem:lemma2}}
Combining Lemma~\ref{lem:distanceproperty} and the coupling probabilistic argument in Lemma~\ref{lem:lemma2_psvrg}, we know from a randomly perturbed point $x_0$, sequence $\{x_t\}$ succeeds in Phase 1 with probability at least $1/6.$ By Lemma~\ref{lem:varaincedistance}, we know with high probability, there exists $C_1=\tdo(1),$ such that $\n{\xi_t}\leq \frac{C_1 L}{\sqrt{b}}\n{x_t-\snap}$ for any $0\leq t\leq T-1$, where $T$ is the super epoch length. Then, combing with Lemma~\ref{lem:phase2} and Lemma~\ref{lem:dis_value_cross_epoch}, with probability at least $1/8$ we know there exists $\eta=\frac{1}{C_6 L}, b= \tdo(n^{2/3}), \delta=\tdo(\min( \frac{\gamma}{\rho}, \frac{m\gamma}{\rho'}))$, $T\leq T_{\max}:=\frac{C_7}{\eta\gamma}$ such that,
$$\n{x_T-\tx}\geq \mathL :=\frac{\gamma}{C_3\rho},\quad \ns{x_T-x_0}\leq \frac{T}{C_4 L}(\hat{f}(x_0)-\hat{f}(x_T))$$
where $C_3, C_4, C_6, C_7=\tdo(1)$.

Since $\ns{x_T-x_0}\leq \frac{T}{C_4 L}(\hat{f}(x_0)-\hat{f}(x_T))$, we have
\begin{align*}
\hat{f}(x_0)-\hat{f}(x_T) \geq& \frac{C_4 L}{T}\ns{x_T-x_0}\\
\geq& \frac{C_4 L}{T}\left(\n{x_T-\tx}-\n{x_0-\tx}\right)^2\\
\geq& \frac{C_4 L}{T}\left(\frac{\gamma}{C_3 \rho}-\delta\right)^2\\
\geq& \frac{C_4 L}{T}\frac{\gamma^2}{4C_3^2 \rho^2}\\
=& \frac{C_4}{4C_7 C_3^2 C_6}\frac{\gamma^3}{\rho^2},
\end{align*}
where the last inequality holds as long as $\delta \leq \frac{\gamma}{2C_3\rho}.$

Since $\hat{f}$ is $L$-smooth and $\nabla \hat{f}(\tx)=0$, we have 
\begin{align*}
\hat{f}(x_0)-\hat{f}(\tx)\leq \frac{L}{2}\ns{\tx-x_0}\leq \frac{L}{2}\delta^2.
\end{align*}
Let the threshold gradient $\mathG:=\frac{\gamma^2}{C_8\rho}$. For the function value difference between two sequence, we have 
\begin{align*}
f(x_T)-\hat{f}(x_T)\leq& \n{\nabla f(\tx)}\cdot \n{x_T -\tx}
\end{align*}
Since $T$ is the length of the current super epoch, we know $\n{x_{T-1} -\tx} < \mathL.$ According to the analysis in Lemma~\ref{lem:phase2}, we also know $\n{x_T -\tx} \leq 3 \n{x_{T-1} -\tx} \leq  3\mathL$. Thus, we have 
\begin{align*}
f(x_T)-\hat{f}(x_T)\leq& \mathG\cdot 3\mathL\\
\leq& \frac{\gamma^2}{C_8\rho}\frac{3\gamma}{C_3\rho}\\
=& \frac{3}{C_8C_3}\frac{\gamma^3}{\rho^2}.
\end{align*}

Thus, with probability at least $1/8$, we know
\begin{align*}
f(x_T)-f(\tx) 
=& f(x_T)-\hat{f}(\tx)\\
=& \hat{f}(x_T) - \hat{f}(x_0) + \hat{f}(x_0) - \hat{f}(\tx) + f(x_T) - \hat{f}(x_T)\\
\leq& -\frac{C_4}{4C_7 C_3^2 C_6}\frac{\gamma^3}{\rho^2}+ \frac{L}{2}\delta^2 + \frac{3}{C_8 C_3}\frac{\gamma^3}{\rho^2}.
\end{align*}

If Phase 1 is not successful, the function value may not decrease. On the other hand, we know $\hat{f}(x_T) - \hat{f}(x_0)\leq 0$ with high probability. Thus, with high probability, we know
\begin{align*}
f(x_T)-f(\tx) 
\leq \frac{L}{2}\delta^2 + \frac{3}{C_8 C_3}\frac{\gamma^3}{\rho^2}.
\end{align*}

Assuming $\delta\leq \sqrt{\frac{C_4}{84C_7 C_3^2  C_6}\frac{\gamma^3}{\rho^2}}$ and $C_8\geq \frac{504C_7 C_3  C_6}{C_4},$ we know with probability at least $1/8$,
$$f(x_T)-f(\tx)\leq  -\frac{20}{21}\cdot \frac{C_4}{4C_7 C_3^2 C_6}\frac{\gamma^3}{\rho^2};$$
and with high probability, 
$$f(x_T)-f(\tx)\leq  \frac{1}{21}\cdot \frac{C_4}{4C_7 C_3^2 C_6}\frac{\gamma^3}{\rho^2}.$$

We finish the proof by choosing $C_5:=\frac{20}{21}\frac{C_4}{4C_7 C_3^2 C_6}$.
\end{proofof}

\section{Proof of Theorem~\ref{thm:maintheorem}}\label{sec:maintheoremproof}
In the previous analysis, we already showed that Algorithm~\ref{alg:ssvrg} can decrease the function value either when the current point has a large gradient or has a large negative curvature. In this section, we combine these two cases to show Stabilized SVRG will at least once get to an $\epsilon$-second-order stationary point within 
$\tdo(\frac{n^{2/3}L\Delta f}{\epsilon^2}+\frac{n\sqrt{\rho}\Delta f}{\epsilon^{1.5}})$ time. We omit the proof for Theorem~\ref{thm:maintheorem_psvrg} since it's almost the same as the proof for Theorem~\ref{thm:maintheorem} except for using different guarantees for negative curvature exploitation super-epoch. 

Recall Theorem~\ref{thm:maintheorem} as follows. 
\begingroup
\def\thetheorem{\ref{thm:maintheorem}}
\begin{theorem}
Assume the function $f(x)$ is $\rho$-Hessian Lipschitz, and each individual function $f_i(x)$ is $L$-smooth and $\rho'$-Hessian Lipschitz. Let $\Delta f:= f(x_0)-f^*$, where $x_0$ is the initial point and $f^*$ is the optimal value of $f$. There exists mini-batch size $b=\tdo(n^{2/3})$, epoch length  $m=n/b$, step size $\eta=\tdo(1/L)$, perturbation radius $\delta=\tdo(\min(\frac{\sqrt{\epsilon}}{\sqrt{\rho}}, \frac{m\sqrt{\rho\epsilon}}{\rho'}))$, super epoch length $T_{\max}=\tdo(\frac{L}{\sqrt{\rho\epsilon}})$, threshold gradient $\mathG=\tdo(\epsilon)$, threshold distance $\mathL=\tdo(\frac{\sqrt{\epsilon}}{\sqrt{\rho}}),$ such that Stabilized SVRG (Algorithm~\ref{alg:ssvrg}) will at least once get to an $\epsilon$-second-order stationary point with high probability using 
$$\tdo(\frac{n^{2/3}L\Delta f}{\epsilon^2}+\frac{n\sqrt{\rho}\Delta f}{\epsilon^{1.5}})$$
stochastic gradients.
\end{theorem}
\addtocounter{theorem}{-1}
\endgroup
\begin{proofof}{Theorem~\ref{thm:maintheorem}}
Recall that we call the steps between the beginning of perturbation and the end of perturbation a super epoch. Outside of the super epoch, we use random stopping, which is equivalent to finish the epoch first and then uniformly sample a point from this epoch. In light of Lemma~\ref{lem:large_gradients}, we divide epochs~\footnote{Here, we only mean the epochs outside of super epochs.} into two types: if at least half of points from $\{x_\tau\}_{\tau=t+1}^{t+m}$ have gradient norm at least $\mathG,$ we call it a {\em useful epoch}; otherwise, we call it a {\em wasted epoch}. For simplicity of analysis, we further define {\em extended epoch}, which constitutes of a useful epoch or a super epoch and all its preceding wasted epochs. With this definition, we can view the iterates of Algorithm~\ref{alg:ssvrg} as a concatenation of extended epochs.

First, we show that within each extended epoch, the number of wasted epochs before a useful epoch or a super epoch is well bounded with high probability. Suppose $\{x_\tau\}_{\tau=t+1}^{t+m}$ is a wasted epoch, we know at least half of points from $\{x_\tau\}_{\tau=t+1}^{t+m}$ have gradient norm at most $\mathG.$ Thus, uniformly sampled from $\{x_\tau\}_{\tau=t+1}^{t+m}$, point $x_{t'}$ has gradient norm $\n{\nabla f(x_{t'})}\leq \mathG$ with probability at least half. Note for different wasted epochs, returned points are independently sampled. Thus, with high probability, the number of wasted epochs in an extended epoch is $\tdo(1)$. As long as the number of ``extended'' epochs is polynomially many through the algorithm, by union bound the number of ``wasted'' epochs for every ``extended'' epoch is $\tdo(1)$ with high probability.  

We divide the extended epochs into the following three types.
\begin{itemize}
\item Type-1: the extended epoch ends with a useful epoch. 
\item Type-2: the extended epoch ends with a super epoch whose starting point has Hessian with minimum eigenvalue less that $-\sqrt{\rho\epsilon}$. 
\item Type-3: the extended epoch ends with a super epoch whose starting point is an $\epsilon$-second-order stationary point.
\end{itemize}

For the type-1 extended epoch, according to Lemma~\ref{lem:large_gradients}, we know with probability at least $1/5,$ the function value decrease by at least $\tdomega(n^{1/3}\epsilon^2/L)$; and with high probability, the function value does not increase. By standard concentration bound, we know after logarithmic number of type-$1$ extended epochs, with high probability, at least $1/6$ fraction of them decrease the function value by $\tdo(n^{1/3}\epsilon^2/L)$.

For the type-2 extended epoch, according to Lemma~\ref{lem:lemma2}, we know with probability at least $1/8,$ the function value decreases by at least $C_5 \epsilon^{1.5}/\sqrt{\rho}$; and with high probability, the function value cannot increase by more than $\frac{C_5}{20}\epsilon^{1.5}/\sqrt{\rho}$, where $C_5=\tdtheta(1).$ Again, by standard concentration bound, we know after logarithmic number of type-2 extended epochs, with high probability, at least $1/10$ fraction of them decreases the function value by at least $C_5\epsilon^{1.5}/\sqrt{\rho}$. Let the total number of type-$2$ extended epochs be $N_2$, we know with high probability the overall function value decrease within these type-2 extended epochs is at least $\frac{N_2 C_5}{20}\epsilon^{1.5}/\sqrt{\rho}$. 

Thus, after $\tdo(\frac{L \Delta f}{n^{1/3}\epsilon^2})$ number of type-1 extended epochs or $\tdo(\frac{\sqrt{\rho}\Delta f}{\epsilon^{1.5}})$ number of type-2 extended epochs, with high probability the function value decrease will be more than $\Delta f$. We also know that the time consumed within a type-1 extended epoch is $\tdo(n)$ with high probability; and that for a type-2 extended epoch is $\tdo(n+ n^{2/3}L/\sqrt{\rho\epsilon})$. Therefore, after 
$$\tdo\left(\frac{L \Delta f}{n^{1/3}\epsilon^2}\cdot n+\frac{\sqrt{\rho}\Delta f}{\epsilon^{1.5}}(n+ \frac{n^{2/3}L}{\sqrt{\rho\epsilon}})\right)$$
stochastic gradients, we will at least once get to an $\epsilon$-second-order stationary point with high probability.  
\end{proofof}
\section{Hessian Lipschitz Parameters for Matrix Sensing}\label{sec:matrix}

In this section we consider a simple example for non-convex optimization and show that in natural conditions the Hessian Lipschitz parameter for the average function $f$ can be much smaller than the Hessian Lipschitz parameter for the individual functions.

The problem we consider is the symmetric matrix sensing problem. In this problem, there is an unknown low rank matrix $M^*\in \mathbb{R}^{d\times d} = U^*(U^*)^\top $ where $U^* \in \R^{d\times r}$. In order to find $M^*$, one can make observations $b_i=\inner{A_i}{M^*}$, where $A_i$'s are random matrices with i.i.d. standard Gaussian entries. A typical non-convex formulation of this problem is as follows:
\begin{equation}
\min_{U\in \mathbb{R}^{d\times r}} f(U)=\frac{1}{2n}\sum_{i=1}^{n}(\inner{A_i}{M}-b_i)^2,\label{eq:sensingobj}
\end{equation}
where $M:=UU^\top,\ U\in\R^{d\times r}.$ It was shown in \citep{bhojanapalli2016global,ge2017no} that all local minima of this objective satisfies $UU^\top  = M^*$ when $n = Cd$ for a large enough constant $C$. We can easily view this objective as a finite sum objective by defining $f_i(U) = \frac{1}{2} (\inner{A_i}{M}-b_i)^2$.

Without loss of generality, we will assume $\|U^*\| = 1$ (otherwise everything just scales with $\|U^*\|$). A slight complication for the objective \eqref{eq:sensingobj} is that the function is not Hessian Lipschitz in the entire $\R^{d\times r}$. However, it is easy to check that if the initial $U_0$ satisfies $\|U_0\| \le 4$ then all the iterates $U_t$ for gradient descent (and SVRG) will satisfy $\|U_t\|\le 4$ (with high probability for SVRG). So we will constrain our interest in the set of matrices $\mathcal{B} = \{U\in \R^{d\times r}: \|U\| \le 4\}$.

\begin{theorem}\label{thm:hessianlipschitzsensing}
Assume sensing matrices $A_i$'s are random matrices with i.i.d. standard Gaussian entries.. When $n \ge Cdr$ for some large enough universal constant $C$, for any $U, V$ in $\mathcal{B} = \{U\in \R^{d\times r}: \|U\| \le 4\}$, for objective $f$ in Equation \eqref{eq:sensingobj}, with high probability
$$
\|\nabla^2 f(U) - \nabla^2 f(V)\| \le O(1)\|U-V\|_F.
$$
On the other hand, for the individual function  $f_i(U) = \frac{1}{2} (\inner{A_i}{M}-b_i)^2$ with high probability, there exists $U, V$ in $\mathcal{B}$ such that
$$
\|\nabla^2 f_i(U) - \nabla^2 f_i(V)\| = \Omega(d)\|U-V\|_F.
$$
\end{theorem}

Before we prove the theorem, let us first see what this implies. In a natural case when $r$ is a constant, $n = Crd$ for large enough $C$, for the matrix sensing we have $\rho = O(1)$, but $\rho' = \Omega(d) = \Omega(n)$. Therefore, the guarantee for Perturbed SVRG (Theorem~\ref{thm:maintheorem_psvrg}) is going to be much worse compared to the guarantee of Stabilized SVRG (Theorem~\ref{thm:maintheorem}).

Let us first adapt the notation from \cite{ge2017no} and write out the Hessian of the objective.

\begin{definition}
For matrices $B, B'$, let $B:\mathcal{H}:B' \triangleq \frac{1}{n}\sum_{i=1}^{n}\inner{A_i}{B}\inner{A_i}{B'}$.
\end{definition}

\begin{lemma}[\cite{ge2017no}]\label{lem:hessiancomputation}
The Hessian of the objective $f(U)$ in direction $Z \in \R^{d\times r}$ can be computed as
\begin{equation*}
\nabla^2f(U)(Z,Z) = (UZ^\top+ZU^\top):\mathcal{H}:(UZ^\top+ZU^\top)
    +2(UU^\top-M^*):\mathcal{H}:ZZ^\top.
\end{equation*}
Similarly, the Hessian of an individual function $f_i(U)$ satisfies
\begin{equation*}
\nabla^2f_i(U)(Z,Z) = \inner{UZ^\top}{A_i+A_i^\top}^2
    +1/2\inner{UU^\top-M^*}{A_i+A_i^\top}\inner{ZZ^\top}{A_i+A_i^\top}.
\end{equation*}
\end{lemma}

Another key property we will need is the Restrict Isometry Property (RIP)~\citep{recht2010guaranteed}.

\begin{definition}[Matrix RIP] The set of sensing matrix is $(r,\delta)$-RIP if for any matrix $B$ of rank at most $r$ we always have
$$
(1-\delta)\|B\|_F^2 \le B:\mathcal{H}:B \le (1+\delta) \|B\|_F^2.
$$
\end{definition}

\cite{candes2011tight} showed that random Gaussian sensing matrices satisfy RIP with high probability as long as $n$ is sufficiently large

\begin{theorem}[\cite{candes2011tight}]\label{thm:rip} Suppose $n \ge Cdr/\delta^2$, then random Gaussian sensing matrices satisfy the $(r,\delta)$-RIP with high probability.
\end{theorem}


Now we are ready to prove Theorem~\ref{thm:hessianlipschitzsensing}.

\begin{proofof}{Theorem~\ref{thm:hessianlipschitzsensing}} We will first prove the upperbound for the average function.

For the upperbound, assume that the sensing matrices are $(2r, \delta)$-RIP for $\delta = 1/10$. By Theorem~\ref{thm:rip} we know this happens with high probability when $n \ge 200Crd$ where $C$ was the constant in Theorem~\ref{thm:rip}.

For any $\|U\|, \|V\| \le 4$ and $Z\in\mathbb{R}^{d\times r}$, we use Lemma~\ref{lem:hessiancomputation} to compute the Hessian and take the difference in the direction of $Z$
\begin{align*}
&|\nabla^2f(U)(Z,Z)-\nabla^2f(V)(Z,Z)| \\
&= (UZ^\top +ZU^\top ):\mathcal{H}:(UZ^\top +ZU^\top )
    -(VZ^\top +ZV^\top ):\mathcal{H}:(VZ^\top +ZV^\top )\\
    &\qquad
      +2(UU^\top -M^*):\mathcal{H}:ZZ^\top  -2(VV^\top -M^*):\mathcal{H}:VV^\top  \\
&=(UZ^\top +ZU^\top ):\mathcal{H}:\big((U-V)Z^\top +Z(U-V)^\top \big)\\
    &\qquad
    +\big((U-V)Z^\top +Z(U-V)^\top \big):\mathcal{H}:(VZ^\top +ZV^\top )\\
    &\qquad
      +2(UU^\top -VV^\top ):\mathcal{H}:ZZ^\top  \\
&\leq (1+\delta) \|UZ^\top +ZU^\top \|_F\|(U-V)Z^\top +Z(U-V)^\top \|_F\\
    &\qquad +(1+\delta)\|(U-V)Z^\top +Z(U-V)^\top \|_F\|VZ^\top +ZV^\top \|_F \\
    &\qquad +2(1+\delta) \|UU^\top -VV^\top \|_F\|ZZ^\top \|_F \\
&\leq 32(1+\delta)\|Z\|_F^2\|U-V\|_F + 16(1+\delta)\|U-V\|_F\|Z\|_F^2\\
&=48(1+\delta)\|U-V\|_F\|Z\|_F^2,
\end{align*}
where the first inequality uses the definition of RIP and Cauchy-Schwartz inequality, and the second inequality uses $\|U\|, \|V\|\leq 4$ and the fact that $\|AB\|_F \le \|A\| \|B\|_F$.
Thus, for any $U,V\in\mathcal{B}$, and any direction $Z$, we have 
\begin{align*}
\frac{|\nabla^2f(U)(Z,Z)-\nabla^2f(V)(Z,Z)|}{\|Z\|_F^2}
\leq 48(1+\delta)\|U-V\|_F.
\end{align*}
This implies that $\rho \le 48(1+\delta)=\frac{264}{5}$.

Next we prove the lowerbound for individual functions. We will consider $V = U+\epsilon \Delta$ and let $\epsilon$ go to 0. This allows us to ignore some higher order terms in $\epsilon$. Following Lemma~\ref{lem:hessiancomputation}, let $A = A_i+A_i^\top$, we have
\begin{align*}
\nabla^2 f_i(V)(Z, Z) - \nabla^2 f_i(U)(Z,Z) & = 2\epsilon \inner{\Delta Z^\top}{A}\inner{UZ^\top}{A} + \epsilon \inner{\Delta U^\top}{A} \inner{ZZ^\top}{A} + O(\epsilon^2).
\end{align*}

It is easy to check that the matrix $A/\sqrt{2}$ has the same distribution as the Gaussian Orthogonal Ensemble. By standard results in random matrix theory \citep{bai1988necessary,tao2012topics} we know with high probability
$\lambda_{max}(A) \ge \sqrt{d}$. Let $\lambda = \lambda_{max}(A)$ and $v$ be a corresponding eigenvector. We will take $U = \Delta=Z = ve_1^\top$ where $e_1$ is the first basis vector. In this case, we have
\begin{align*}
\nabla^2 f_i(V)(Z, Z) - \nabla^2 f_i(U)(Z,Z) & = 2\epsilon \inner{\Delta Z^\top}{A}\inner{UZ^\top}{A} + \epsilon \inner{\Delta U^\top}{A} \inner{ZZ^\top}{A} + O(\epsilon^2)\\
& = 2\epsilon\inner{vv^\top}{A}^2 + \epsilon \inner{vv^\top}{A}^2 + O(\epsilon^2) \\
& = 3\epsilon \lambda^2 + O(\epsilon^2)\\
& = 3\lambda^2 \|U-V\|_F + o(\|U-V\|_F).
\end{align*}

Note that $Z$ satisfies $\|Z\|_F = 1$, so the calculation above implies $\rho' \ge 3\lambda^2 \ge 3d$.

\end{proofof}
\section{Tools}

Matrix concentration bounds tell us that with enough number of independent samples, the empirical mean of a random matrix can converge to the mean of this matrix.

\begin{lemma}[Matrix Bernstein; Theorem 1.6 in \cite{tropp2012user}]\label{lm:bernstein_original}
Consider a finite sequence $\{Z_k\}$ of independent, random matrices with dimension $d_1\times d_2$. Assume that each random matrix satisfies
$$\E[Z_k]=0\ and \ \n{Z_k}\leq R \ almost \ surely. $$
Define 
$$\sigma^2:= \max\Big\{\big\lVert\sum_k \E[Z_k Z_k^*]\big\rVert, \big\lVert \sum_k \E[Z_k^* Z_k]\big\rVert \Big\}.$$
Then, for all $t\geq 0$, 
$$\Pr\Big\{\big\lVert\sum_k Z_k\big\rVert\geq t \Big\}\leq (d_1+d_2) \exp\Big(\frac{-t^2/2}{\sigma^2+Rt/3}\Big).$$
\end{lemma} 

As a corollary, we have:
\begin{lemma}[Bernstein Inequality: Vector Case]\label{vectorBernstein}
Consider a finite sequence $\{v_k\}$ of independent, random vectors with dimension $d$. Assume that each random vector satisfies
$$\n{v_k-\E[v_k]}\leq R\ almost\ surely.$$
Define
$$\sigma^2 := \sum_{k}\E\big[\ns{v_k-\E[v_k]}\big].$$
Then, for all $t\geq 0$,
$$\Pr\Big\{\n{\sum_{k}(v_k-\E[v_k])}\geq t\Big\}\leq (d+1)\cdot \exp\Big(\frac{-t^2/2}{\sigma^2+Rt/3}\Big).$$
\end{lemma}


\end{document}